\documentclass{article}

\usepackage{arxiv}

\usepackage[utf8]{inputenc} 
\usepackage[T1]{fontenc}    
\usepackage{hyperref}       
\usepackage{url}            
\usepackage{booktabs}       
\usepackage{amsfonts}       
\usepackage{nicefrac}       
\usepackage{microtype}      
\usepackage{lipsum}		
\usepackage{graphicx}
\usepackage{amsmath}
\usepackage[numbers]{natbib}
\usepackage{doi}
\usepackage{subcaption}
\usepackage{tikz}
\usepackage{float}
\usepackage{rotating}
\usetikzlibrary{positioning, arrows.meta, backgrounds, fit, calc}

\newcommand{\til}{$\sim$}

\title{From Syntax to Semantics: Unveiling the Emergence of Chirality in SMILES Translation Models}

\author{
Zehao Li$^{1}$, Yasuhiro Yoshikai$^{1}$, Shumpei Nemoto$^{1}$,\\ Hiroyuki Kusuhara$^{1}$, Tadahaya Mizuno$^{1,2,3,\dagger}$\\[1.0ex]
\begin{minipage}{0.65\textwidth}
\small
\raggedright
$^{1}$Laboratory of Molecular Pharmacokinetics, Graduate School of Pharmaceutical Sciences, The University of Tokyo, 7-3-1 Hongo, Bunkyo, Tokyo, Japan\\
$^{2}$The Institute of Statistical Mathematics (ISM), Research Organization of Information and Systems, 190-8562 Tachikawa, Tokyo, Japan\\
$^{3}$tadahaya@gmail.com\\
$^{\dagger}$Author to whom correspondence should be addressed
\end{minipage}
}

\date{}

\renewcommand{\headeright}{}

\hypersetup{
pdftitle={From Syntax to Semantics: Unveiling the Emergence of Chirality in SMILES Translation Models},
pdfsubject={cs.LG, q-bio.QM},
pdfauthor={Zehao Li, Yasuhiro Yoshikai, Shumpei Nemoto, Hiroyuki Kusuhara, Tadahaya Mizuno},
pdfkeywords={Chemical language model, Mechanistic interpretability, Phase transition, Training dynamics, Molecular representation learning, Drug discovery},
}

\begin{document}
\maketitle

\begin{abstract}
	Understanding how chemical language models (CLMs) learn chemical meaning from molecular string representations, rather than only surface-level string patterns, is an important question in chemical representation learning and machine learning for chemistry. Chirality provides a demanding test case: enantiomers can differ greatly in pharmacological activity and toxicity, yet CLMs often struggle to distinguish chiral configurations reliably. Here we present Pan-CORE (Pan-Chemical Omniscale Representation Engine), a family of autoregressive Transformer-based encoder-decoder models for SMILES translation, and use high-temporal-resolution checkpoint analysis to investigate how chiral information is learned during training. Across all tested Pan-CORE variants, we observe a reproducible jump-up in which chiral-token accuracy rises abruptly after a long plateau, suggesting that chiral learning stagnation is not explained by model capacity alone and instead reflects the complexity of chiral constraints. Analyses of attention dynamics, residual-stream trajectories, and latent-space geometry support an encoder-centered mechanism in which chiral-token representations undergo transient destabilization and reconstruction, seen as a V-shaped drop and recovery in vector norm and directional stability, together with a clear reorganization of chiral molecular representations in the latent space. Encoder-decoder cross-evaluation further supports the encoder-centered nature of the transition, and targeted attention-head ablation identifies a small set of chiral-sensitive heads whose removal selectively reduces chiral-token accuracy even in the fully trained model. These findings show that SMILES translation can serve as a useful experimental system for mechanistic analysis of semantic emergence in CLMs, with implications for interpretable chemical representation learning.
\end{abstract}


\section{Introduction}
\label{sec:introduction}
The rise of large language models (LLMs) has revolutionized the field of natural language processing (NLP), and their applications are rapidly expanding into vision, speech, and the natural sciences~\cite{Naveed2025-ce}. In particular, the fields of chemistry and biology have witnessed remarkable advances in chemical language models (CLMs)~\cite{Flores-Hernandez2024-cd}, protein language models (PLMs)\cite{Xiao2025-zz}, and genomic language models (GLMs)\cite{Shu2026-lr}, which bypass natural language to operate directly on molecular, amino acid, and nucleotide sequences. Representation learning from one-dimensional strings such as the Simplified Molecular Input Line Entry System (SMILES)~\cite{Weininger1988-lb} was greatly accelerated by variational autoencoder (VAE)-based approaches that explore structures within continuous chemical latent spaces~\cite{Gomez-Bombarelli2018-zy}. Extensive subsequent research addressed the effective learning of structural information from such strings and the nature of the resulting molecular representations~\cite{Fender2025-xt, Ganeeva2025-he, Krenn2019-ej}. Today, capitalizing on the powerful Transformer~\cite{Vaswani2017-fv} architectures whose design principles have been shaped by advances in LLM research, large-scale CLMs pretrained on hundreds of millions of molecular data points, such as ChemBERTa~\cite{Chithrananda2020-ia}, MoLFormer~\cite{Ross2022-mw} and SMI-TED~\cite{Soares2025-pf}, have been established. Multimodal models capable of integrating not only natural and chemical languages but also fundamentally different molecular representations are further beginning to emerge~\cite{Luo2023-ii, Singh2025-mb, Soares2025-ia}.

As these AI models are increasingly deployed in real-world scientific endeavors such as drug discovery and materials development, securing interpretability—understanding why a model arrives at a given prediction and what it has learned internally—has emerged as a critical challenge. In NLP, mechanistic interpretability research that reverse-engineers models' internal representations and training dynamics is actively progressing~\cite{Clark2019-ge, Rogers2020-in, Nelson-Elhage-Neel-Nanda-Catherine-Olsson-Tom-Henighan-Nicholas-Joseph-Ben-Mann-Amanda-Askell-Yuntao-Bai-Anna-Chen-Tom-Conerly-Nova-DasSarma-Dawn-Drain-Deep-Ganguli-Zac-Hatfield-Dodds-Danny-Hernandez-Andy-Jones-Jackson-Kernion-Liane-Lovitt-Kamal-Ndousse-Dario-Amodei-Tom-Brown-Jack-Clark-Jared-Kaplan-Sam-McCandlish-Chris-Olah2021-jy, Im2025-el, Ge2026-kj}. Natural language, however, is highly context-dependent and contains many ambiguous words and fuzzy boundaries, making it difficult to track when a model moves beyond surface grammar during training. Chemical languages such as SMILES, by contrast, are composed of symbols representing atoms, bonds, and ring openings and closures, each carrying a precise and unambiguous physicochemical meaning. This property makes SMILES a well-suited experimental system for tracking, at high resolution, how a language model moves from surface string processing toward learning chemical semantic constraints. Nevertheless, compared with standard metric-based analyses of CLMs~\cite{Nemoto2023-vo}, studies of their attention circuits, features, and training dynamics remain limited~\cite{Varadi2025-kw}.

Among the chemical semantic constraints that models must learn, chirality represents a particularly important and challenging case. Enantiomers can differ dramatically in pharmacological activity and in vivo toxicity; as illustrated by the thalidomide tragedy~\cite{Lenz1988-lw, Blaschke1979-av}, chiral distinctions require rigorous and precise treatment. Yet, many existing CLMs are intentionally trained on SMILES stripped of stereochemical information~\cite{Ross2022-mw, Soares2025-pf, Mahdizadeh2025-id}. Notably, Yoshikai et al. have reported that Transformers face a specific difficulty in recognizing chirality-encoding tokens (@ and @@), and that the resulting confusion of enantiomers leads to prolonged stagnation in training~\cite{Yoshikai2024-ea}.

Whether this learning difficulty can be solved simply by increasing model capacity, or instead reflects a reproducible training dynamic linked to complex stereochemical constraints, remains unclear. In this work, we construct Pan-CORE (Pan-Chemical Omniscale Representation Engine), a family of autoregressive Transformer-based encoder-decoder models for SMILES translation incorporating a fixed-length bottleneck latent layer, and systematically track the dynamics of chiral semantic acquisition through high-temporal-resolution checkpoint analysis. We identify a phase transition in which both training and evaluation accuracy rise simultaneously and abruptly following a clear loss plateau, and elucidate its mechanistic basis through multi-level analysis. This work shows that SMILES translation is a useful experimental system for studying, at high resolution, how CLMs learn complex chemical semantic constraints, and lays the groundwork for studying semantic emergence in CLMs. Our main contributions are as follows:
\begin{itemize}
    \item We show, using the Pan-CORE framework, that chiral learning stagnation and the subsequent "jump-up" are reproducible across tested model variants and are unlikely to be explained by model capacity alone.
    \item We identify the jump-up as the abrupt learning of SMILES-level chiral constraints, characterized by a sudden rise in chiral-token accuracy and a sharp improvement in model confidence over R/S configurations.
    \item We provide a mechanistic account of the jump-up as an encoder-centered semantic emergence event, supported by analyses of attention dynamics, residual-stream trajectories, latent-space geometry, encoder-decoder cross-evaluation, and attention-head ablation.
\end{itemize}

\section{Materials \& Methods}
\label{sec:materialsandmethods}

\subsection{Dataset}
\label{sec:dataset}
\subsubsection*{Data Sources}
The chemical compounds used for training and evaluation were collected in SMILES format from the following databases:
\begin{itemize}
    \item ZINC20~\cite{Irwin2020-za} (accessed on March 15, 2025) \\
    A database of commercially available small organic molecules for virtual screening, which compiles three-dimensional structures and other properties. Approximately 1 billion unique compounds were obtained.
    \item PubChem~\cite{Kim2025-cx} (accessed on March 7, 2025) \\
    A comprehensive compound database focusing on small to medium-sized molecules. It compiles the structures, physical properties, and bioactivities of a wide variety of chemical species, including organic compounds, inorganic compounds, salts, and mixtures. Approximately 119 million unique compounds were obtained.
\end{itemize}

\subsubsection*{Tokenization}
To enable the model to process SMILES accurately, we employed a tokenization method that decomposes sequences into symbols, numbers, and elemental symbols. Methods commonly used in natural language processing (NLP), such as BPE\cite{Sennrich2015-yh}, WordPiece\cite{Wu2016-za}, and SentencePiece\cite{Kudo2018-ku}, rely on frequency-based splitting and are effective for learning natural language distributions. However, when processing SMILES—a one-dimensional string representation of compound structures—a method based on chemical grammar rather than statistical string frequency is preferable for preserving chemically meaningful units. In our approach, the smallest semantic units comprising SMILES—atoms, bonds, ring closure digits, and structural symbols (brackets and stereochemical notations)—are strictly separated and extracted.

Our tokenization method, which isolates individual symbols rather than treating bracketed contents (e.g., [C@@H]) as single collective tokens, has two main characteristics. The first is representational robustness against unknown atomic states. By combining basic tokens such as "Structural \& Bonds" and "Chemical Elements" (listed in Table ~\ref{tokentable}), this approach comprehensively covers theoretically possible atomic states with a minimal vocabulary size. This alleviates the need to register every combination of isotope, charge, hydrogen count, and stereoconfiguration as separate vocabulary entries, allowing the model to infer properties from constituent elements even for molecules containing rare charges or special isotopes unseen in the training data.
The second advantage is the promotion of abstracting chemical grammar and generalizing learning. By isolating symbols, the model directly learns universal chemical regularities regardless of atomic species, such as '@' and '@@' governing stereoconfiguration, and '+' or '-' indicating charge. This encourages a more fundamental understanding of chemical structures independent of the frequency distribution of specific compound groups.

\subsubsection*{Preprocessing and Filtering}
Considering computational resource constraints and the balance of data distribution, undersampling was applied to ZINC20 via random sampling, reducing it to 10\% of its total (approximately 100 million compounds).

During data preprocessing, we initially used RDKit~\cite{UnknownUnknown-kl} (ver. 2024.3.6) to extract valid compound structures. Sanitization (structure standardization and validation) was intentionally disabled to preserve broad structural diversity, including salts, mixtures, and hypervalent compounds. Aromaticity addition was applied later as needed, and unsuitable structures containing wildcard or quadruple bonds were selectively removed. Duplicates were eliminated by first converting all compounds to canonical SMILES and preferentially deleting overlapping entries from the PubChem dataset. Furthermore, to improve the geometric invariance and robustness of the model~\cite{Bjerrum2017-jz, Arus-Pous2019-sr}, at least one randomized SMILES with a randomly specified starting atom was generated for each canonical SMILES.

Balancing target molecular size with computational cost, we adopted the sequence length of SMILES as a filtering criterion. To encompass molecules up to a medium size (molecular weight around 2000), the maximum sequence length was limited to 100 or less for ZINC20, and 400 or less for PubChem. The correlation between the sequence length after canonical SMILES tokenization and molecular weight is shown in Table ~\ref{mwseqlen}, confirming that this criterion adequately covers the intended molecular weight range.

Additionally, to optimize learning efficiency and minimize padding within minibatches, we performed bucketing based on sequence length. The data were classified into five distinct buckets: ZINC20 (sequence length \til 100) and PubChem (\til 100, 101\til 200, 201\til 300, 301\til 400).

\subsubsection*{Training and Evaluation Split}
To appropriately evaluate generalization performance on the bucketed dataset, we split the data into training and evaluation (Validation / Test) sets. Specifically, 2,000 SMILES were randomly extracted from each of the five aforementioned sequence length buckets to construct an evaluation dataset totaling 10,000 Random-Canonical pairs. The remaining data were entirely allocated to the training dataset.

As shown in Table ~\ref{numdataraw}, the amount of PubChem data in buckets exceeding a sequence length of 100 was significantly smaller than in buckets under 100. We anticipated that this scarcity of long-sequence SMILES would hinder the acquisition of long-range dependencies. To mitigate this imbalance and ensure learning efficiency for long-chain molecules, we implemented randomized augmentation for the longer sequences. The number of randomized SMILES generated per canonical SMILES was scaled according to sequence length: 10 for the PubChem 101\til 200 bucket, 50 for 201\til 300, and 100 for 301\til 400. This targeted augmentation effectively corrected the extreme imbalance in Random-Canonical pairs while maintaining geometric invariance, yielding a data distribution highly suitable for training (Table ~\ref{numdataaug}).

\subsection{Model Architecture}
\label{sec:modelarchitecture}
\begin{figure}[H]
    \centering
    
    \begin{subfigure}{0.75\textwidth}
        \centering
        \includegraphics[width=\linewidth]{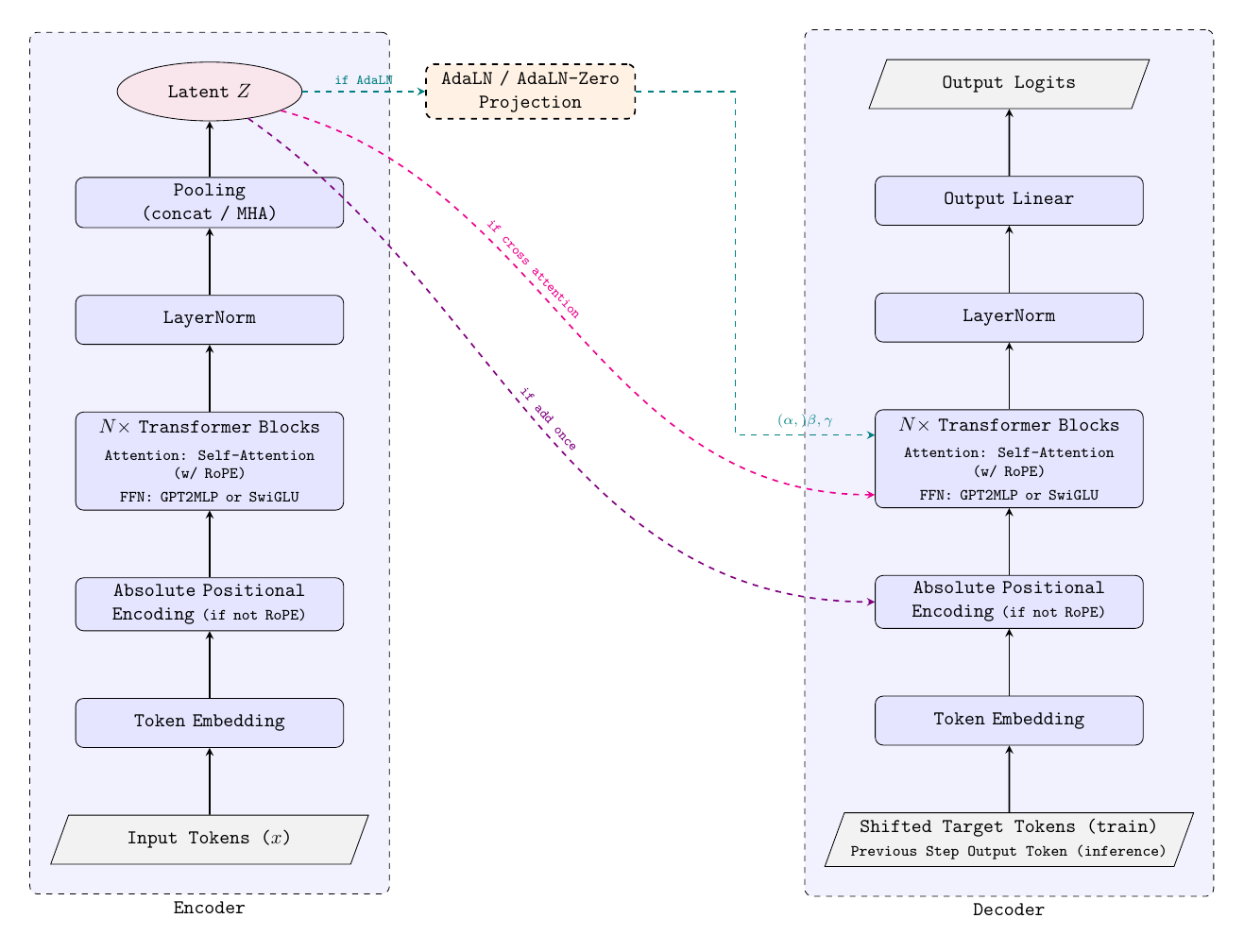}
        \caption{Overall Model}
        \label{overallmodel}
    \end{subfigure}
    
    \vspace{1em}
    
    \begin{subfigure}{0.48\textwidth}
        \centering
        \includegraphics[width=\linewidth]{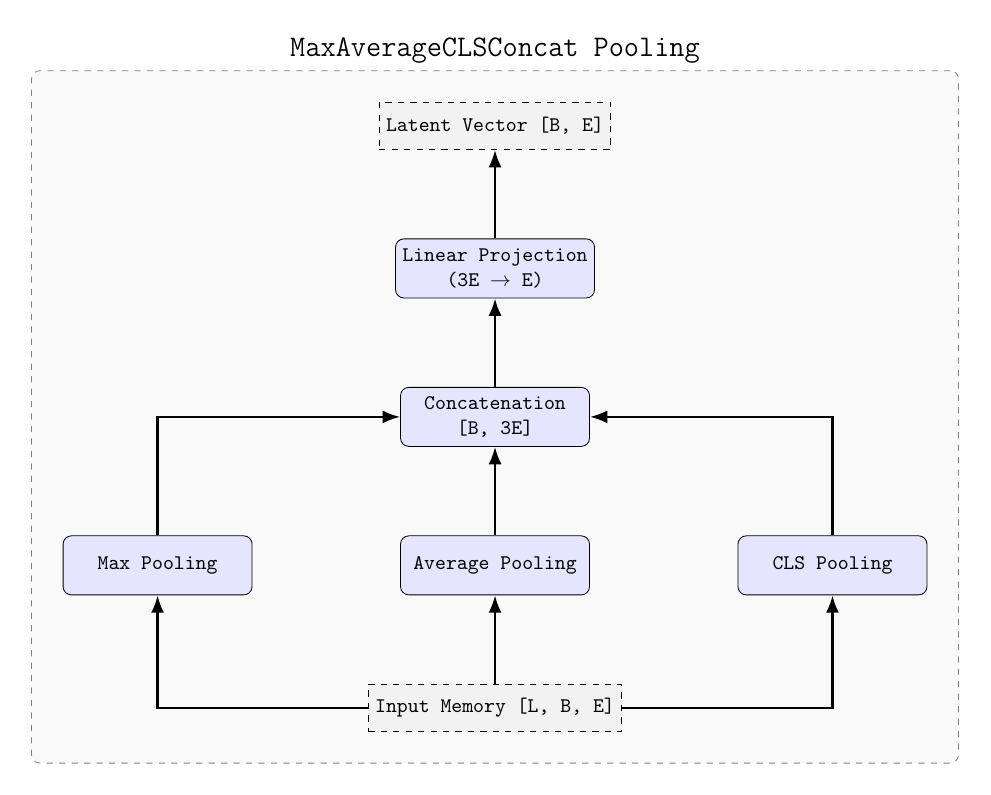}
        \caption{Concat Pooling}
        \label{concatpooling}
    \end{subfigure}
    \hfill
    \begin{subfigure}{0.48\textwidth}
        \centering
        \includegraphics[width=\linewidth]{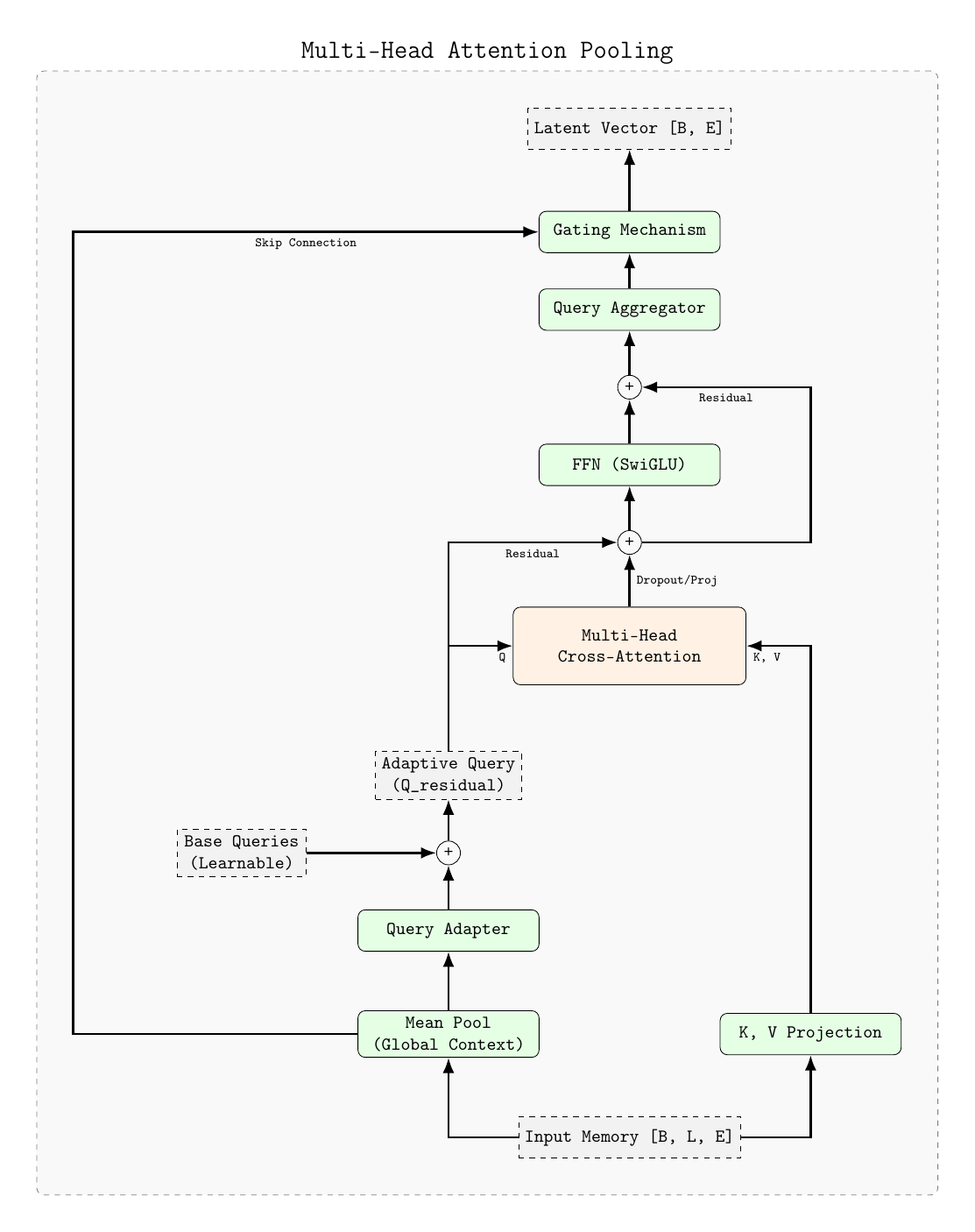}
        \caption{Attention Pooling}
        \label{mhapooling}
    \end{subfigure}

    \caption{Overview of the Pan-CORE architecture. (\subref{overallmodel}) Overall encoder-decoder structure. The encoder processes input tokens through N Transformer blocks and compresses the resulting sequence into a fixed-length latent vector $z$ via a pooling layer. The decoder receives z as conditioning and autoregressively generates the target token sequence. Latent conditioning is applied via one of three mechanisms: Add-Once, Cross-Attention (XAttn), or AdaLN-Zero. (\subref{concatpooling}) Concat Pooling: extracts max, mean, and CLS representations and projects their concatenation to the latent dimension. (\subref{mhapooling}) Multi-Head Attention (MHA) Pooling: adaptively aggregates token representations via learned queries and multi-head cross-attention.}
    \label{model}
\end{figure}

To simultaneously achieve the robust featurization of SMILES and the high-fidelity generation of molecular sequences—while tracking the dynamics of semantic constraint acquisition—we constructed Pan-CORE, an autoregressive Transformer-based encoder-decoder model. The overall architecture inherited Transformer-VAE\cite{Yoshikai2024-zb} with the variational components removed to yield a standard autoencoder architecture.(Fig.~\ref{overallmodel}) The encoder featurizes the input token sequence via token embeddings and positional encodings, processing it through $N$ Transformer blocks containing Self-Attention and Feed-Forward Networks (FFN). An aggregation layer then compresses this variable-length sequence into a single fixed-length latent vector $z$. The decoder receives $z$ as conditioning and autoregressively generates the target sequence through $N$ decoder Transformer blocks, ultimately outputting the generation probability for each token via a linear layer. To optimize this architecture, we systematically evaluated multiple approaches across four key components: positional encoding, feed-forward network (FFN) design, sequence aggregation, and latent conditioning.

\textbf{Positional Encoding:} To inject sequence order information into the model, we established a baseline using classical absolute positional encoding. To enhance the model's ability to capture relative distances between tokens—a crucial factor for understanding topological relationships within long SMILES strings—we also evaluated Rotary Position Embedding (RoPE)~\cite{Su2021-pi}. Widely adopted in modern large language models (LLMs)~\cite{Grattafiori2024-fg, OpenAI2025-yo, Yang2025-ng, Gemma-Team2024-it}, RoPE dynamically incorporates relative positional information directly into the attention mechanism, improving generalization to varying sequence lengths.

\textbf{Feed-Forward Network (FFN) Design:} For the point-wise transformations within the Transformer blocks, our baseline employed the standard FFN block from the original Transformer~\cite{Vaswani2017-fv} (hereafter called GPT2MLP~\cite{Radford-A-Wu-J-Child-R-Luan-D-Amodei-D-and-Sutskever-I2019-qa}). To further boost the model's expressiveness, we tested an advanced configuration utilizing SwiGLU~\cite{Shazeer2020-gi}, another cornerstone architecture of state-of-the-art LLMs~\cite{Grattafiori2024-fg, OpenAI2025-yo, Yang2025-ng}. By combining the Swish activation function with a gating mechanism, SwiGLU provides more sophisticated non-linear mappings, which have been empirically shown to significantly enhance representational capacity.

\textbf{Aggregation Mechanism:} We compared two contrasting approaches based on their feature extraction dynamics. As a static and deterministic strategy, Concat Pooling extracts the maximum value, mean value, and the initial \texttt{<s>} token (analogous to the \texttt{[CLS]} token in BERT~\cite{Devlin2018-ck}) representation along the sequence direction, concatenating them before projecting to a fixed length (Fig.~\ref{concatpooling}). In contrast, Multi-Head Attention (MHA) Pooling utilizes a dynamic and context-aware approach. It generates a query capturing the global context and performs adaptive feature extraction via a Multi-Head Cross-Attention mechanism, allowing the model to dynamically weigh the importance of each token based on the specific input sequence (Fig.~\ref{mhapooling}).

\textbf{Latent Conditioning:} Finally, to integrate the latent vector into the decoder, we evaluated three methods varying in computational cost and conditioning strength: (1) a lightweight Add-Once method that adds the latent vector only at the decoder's initial input layer; (2) XAttn, which computes Cross-Attention with the latent vector after Self-Attention in all decoder blocks; and (3) AdaLN-ZeRO, an adaptive layer normalization method originally proposed for the Diffusion Transformer (DiT) architecture~\cite{Peebles2022-up}. This technique dynamically generates scale and shift parameters for both residual connections and LayerNorms directly from the latent vector, enabling highly stable and expressive conditioning.

In this study, the model and latent vector dimensions ($H$) were set to 512. The number of Transformer blocks was 8 for both the encoder and decoder, with 8 Attention heads. Following their respective original papers~\cite{Vaswani2017-fv, Shazeer2020-gi}, the intermediate dimension for GPT2MLP was set to 2048 ($4H$), and 1365 ($\frac{2}{3}4H$) for SwiGLU. The model utilizing absolute positional encoding, GPT2MLP, Concat Pooling, and Add-Once is denoted as \textit{pancore-baseline}. The model utilizing RoPE, SwiGLU, MHA Pooling, and Add-Once is denoted as \textit{pancore-addonce}. Finally, the model incorporating RoPE, SwiGLU, MHA Pooling, XAttn, and AdaLN-ZeRO is denoted as \textit{pancore-xattn-adalnzero}. For specific model sizes, refer to Table ~\ref{modelsize}.

\subsection{Training Strategy}
\label{sec:trainingstrategy}
To encourage the model to learn SMILES as structural entities rather than mere strings, we framed the objective as a translation task (compliant with CDDD frameworks~\cite{Winter2019-sm}). Specifically, the encoder receives a randomized SMILES, and the decoder is trained to output the corresponding canonical SMILES. This strategy promotes the learning of structure-related features beyond surface string syntax.

Given the variance in sequence lengths, we devised a curriculum learning-style bucket sampling strategy to stabilize and accelerate training~\cite{Bengio2009-ep}. The selection probability of each bucket was dynamically adjusted throughout the training process. In the early stages, shorter sequences (ZINC20 (\til 100), PubChem (\til 100)) were preferentially sampled. As training progressed, the probability of sampling longer sequences gradually increased, culminating in equal sampling probabilities for all buckets by the final stage. Furthermore, the sampling probability for PubChem (\til 100) was decreased more slowly than ZINC20 (\til 100) to maintain structural diversity. Detailed algorithms are provided in Supplementary Section.~\ref{sampling}.

Training was conducted across four NVIDIA H100 GPUs with a global batch size of 2048 SMILES per step. We utilized the RAdamScheduleFree optimizer~\cite{Defazio2024-ja} (lr=3.0e-4, r=0.5, weight\_decay=0.0) combined with ZClip~\cite{Kumar2025-lx} for adaptive gradient clipping. We defined one epoch as 10,000 steps, updating bucket selection probabilities per epoch over a total of 700,000 steps (70 epochs). At the initial epoch, probabilities were set to 40\% for both ZINC20 (\til 100) and PubChem (\til 100), 10\% for PubChem (101\til 200), and 5\% each for PubChem (201\til 300) and (301\til 400). All buckets converged to equal probabilities at epoch 69. To manage VRAM for long sequences, local batch sizes were reduced and gradient accumulation was employed for long-sequence buckets, ensuring sampling from the identical bucket during accumulation steps. Dropout was set to 0.2. Checkpoints were saved every 5,000 steps for standard training, and every 500 steps for the high-resolution interval used for dynamic analysis.

\subsection{Analytical Metrics}
\label{sec:analyticalmetrics}

\subsubsection*{Autoregressive Inference}
To evaluate translation performance, we calculated the exact match rate (accuracy) between the autoregressive output and the canonical SMILES target for each bucket. To investigate the contribution of chirality, we calculated several sub-metrics: accuracy for SMILES containing chiral marks (chi-SMILES accuracy), accuracy after masking chiral marks in mistranslated chi-SMILES (Mistranslated chi-SMILES accuracy w/o '@'), and accuracy for SMILES without chiral marks (non-chi-SMILES accuracy). Token concordance was also calculated to assess vocabulary-level accuracy.

\subsubsection*{Logits-level Analysis}
To evaluate confidence in predicting chiral tokens (@, @@), we analyzed the decoder's output logits. Average Perplexity, representing the uncertainty of the vocabulary distribution, was calculated at chiral token positions within the target sequence.
\begin{equation}
\mathrm{Perplexity}
  = \frac{1}{|\mathcal{C}|}
    \sum_{(i,t)\in\mathcal{C}}
    \exp\!\left(
      H\!\left(\hat{\mathbf{p}}_{i,t}\right)
    \right),
\qquad
H(\hat{\mathbf{p}}_{i,t}) = -\sum_{v} \hat{p}_{i,t} \log(\hat{p}_{i,t} + \epsilon),
\end{equation}
let $\mathcal{C}$ be the set of (sample, position) pairs ($i,t$) where the token $t$ is a chiral token (@, @@) in target sequences, $z_{i,t}$ be the output logit vector at position t, $\hat{p}_{i,t}$ be softmax($z_{i,t}$).

Additionally, to detect when the model hesitated in determining R/S configurations, we calculated the Chiral Dilemma Rate (CDR): the frequency at which both the top and second-choice predictions at a target chiral position were chiral tokens.
\begin{equation}
\mathrm{CDR}
  = \frac{1}{|\mathcal{C}|}
    \sum_{(i,t)\in\mathcal{C}}
    \mathbf{1}\!\left[
      \left\{\tau_1(i,t),\, \tau_2(i,t)\right\}
      = \left\{\texttt{@},\, \texttt{@@}\right\}
    \right],
\end{equation}
where $\tau_1$, $\tau_2$ are the top-1 and top-2 tokens by logit value.

\subsubsection*{Attention Dynamics}
To quantify how strongly the Transformer attends to chiral tokens, we computed the Attention Mass, defined as the average sum of attention weights directed toward chiral tokens across each head and layer.
\begin{equation}
\mathrm{AttnMass}^{(l,h)}
  = \frac{1}{|\mathcal{C}|}
    \sum_{(i,t)\in\mathcal{C}}
    \frac{1}{\ell_i}
    \sum_{q=1}^{\ell_i}
    A^{(l,h)}_{i,q,t},
\end{equation}
where $A^{(l,h)}_{i,q,k}$ represents attention weight from query $q$ to key $k$ at layer $l$, head $h$, sample $i$ (post-softmax).

We also calculated the weighted average of graph distances for elements to which chiral tokens attend. We also evaluated attention entropy, which measures the dispersion of attention from chiral tokens to other tokens. Details are shown in Supplementary Section.~\ref{otherattnmetrics}.

\subsubsection*{Residual Stream and Latent Vector Dynamics}
To track internal representations, we evaluated the residual stream: the token-wise hidden state vector additively updated by each Transformer layer via residual connections. We computed the average L2 norm of chiral and non-chiral token vectors, alongside the average cosine similarity between consecutive checkpoints to track directional stability. Similarly, for the latent space (the encoder's final output), we evaluated the L2 norm and cosine similarity for chi-SMILES and non-chi-SMILES, utilizing Centered Kernel Alignment (CKA)~\cite{Kornblith2019-el} to assess global structural changes in the latent space. For computational efficiency, we employ the linear approximation of CKA.
\begin{equation}
\mathrm{CKA}(s \to s{+}1)
  = \frac{
      \left\|\tilde{X}^\top \tilde{Y}\right\|_F^2
    }{
      \left\|\tilde{X}^\top \tilde{X}\right\|_F
      \left\|\tilde{Y}^\top \tilde{Y}\right\|_F
    },
\quad
\tilde{X} = X - \mathbf{1}\bar{\mathbf{x}}^\top,
\end{equation}
where $X$ and $Y$ denote the latent vector matrices for a fixed set of SMILES samples at consecutive training steps $s$ and $s+1$, respectively. $\tilde{X}$ and $\tilde{Y}$ represent their column-centered matrices, and $\|\cdot\|_F$ is the Frobenius norm.

\section{Results \& Discussion}
\label{sec:resultsanddiscussion}

\subsection{Training Progress and Translation Accuracy}
\label{sec:trainingprogress}
Training the \textit{pancore-baseline}, \textit{pancore-addonce}, and \textit{pancore-xattn-adalnzero} models required 86, 98, and 111 hours respectively, reflecting the computational cost of their increased expressivity. Evaluation across sequence length buckets (Table ~\ref{genaccuracy}) shows that models incorporating modern architectures (RoPE, SwiGLU, MHA Pooling) outperformed the classic configuration across all domains. This gap was most pronounced in the longest bucket, PubChem (301\til 400), where \textit{pancore-baseline} achieved an accuracy of 0.6960, compared to 0.8005 for \textit{pancore-addonce}. Training progress (Fig.~\ref{progress}) shows that \textit{pancore-addonce} and \textit{xattn-adalnzero} reduce loss faster and show earlier accuracy gains on long sequences than \textit{pancore-baseline}, suggesting that architectural modernization is vital for long-range dependency modeling.

However, a clear trade-off emerged between expressive power and generalization. \textit{pancore-xattn-adalnzero}, possessing the strongest latent conditioning, achieved near-perfect fitness (>0.98 accuracy) on the training data across all buckets. Yet, on unseen evaluation data, it underperforms \textit{pancore-addonce} on sequences longer than 100 tokens. These results suggest that the highly expressive Cross-Attention and AdaLN-ZeRO mechanisms may have led to overfitting on the augmented long-sequence data. Instead of learning universal molecular representations, the model exhausted its capacity by memorizing specific augmented randomization patterns, ultimately allowing \textit{pancore-addonce} to demonstrate the best overall generalization performance.

\subsection{Discontinuity in Learning and Stereochemistry}
\label{sec:discontinuity}
\begin{figure}[H]
    \centering
    \begin{subfigure}{0.48\textwidth}
        \centering
        \includegraphics[width=\linewidth]{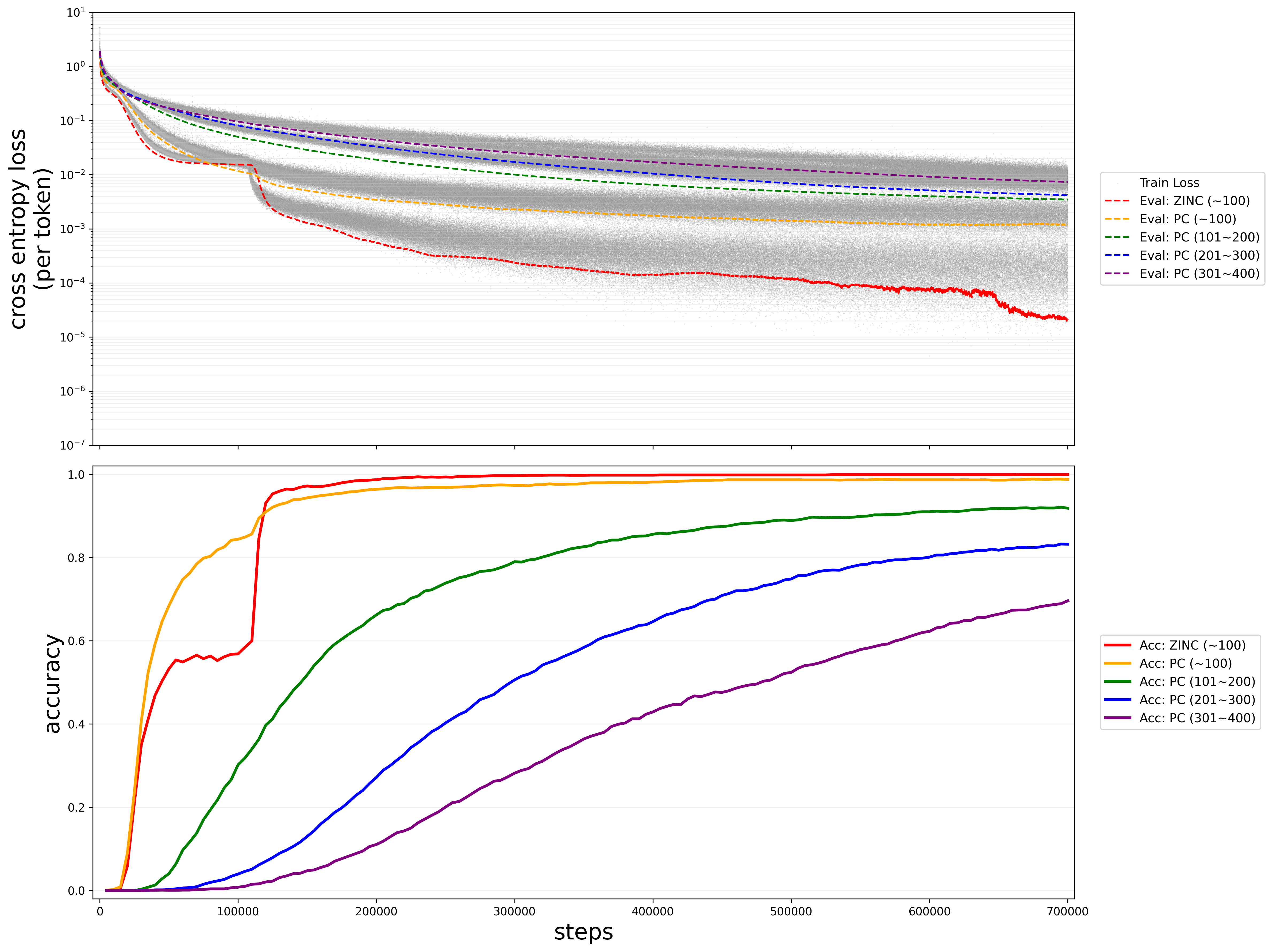}
        \caption{\textit{pancore-baseline}}
        \label{baselineprogress}
    \end{subfigure}
    \hfill
    \begin{subfigure}{0.48\textwidth}
        \centering
        \includegraphics[width=\linewidth]{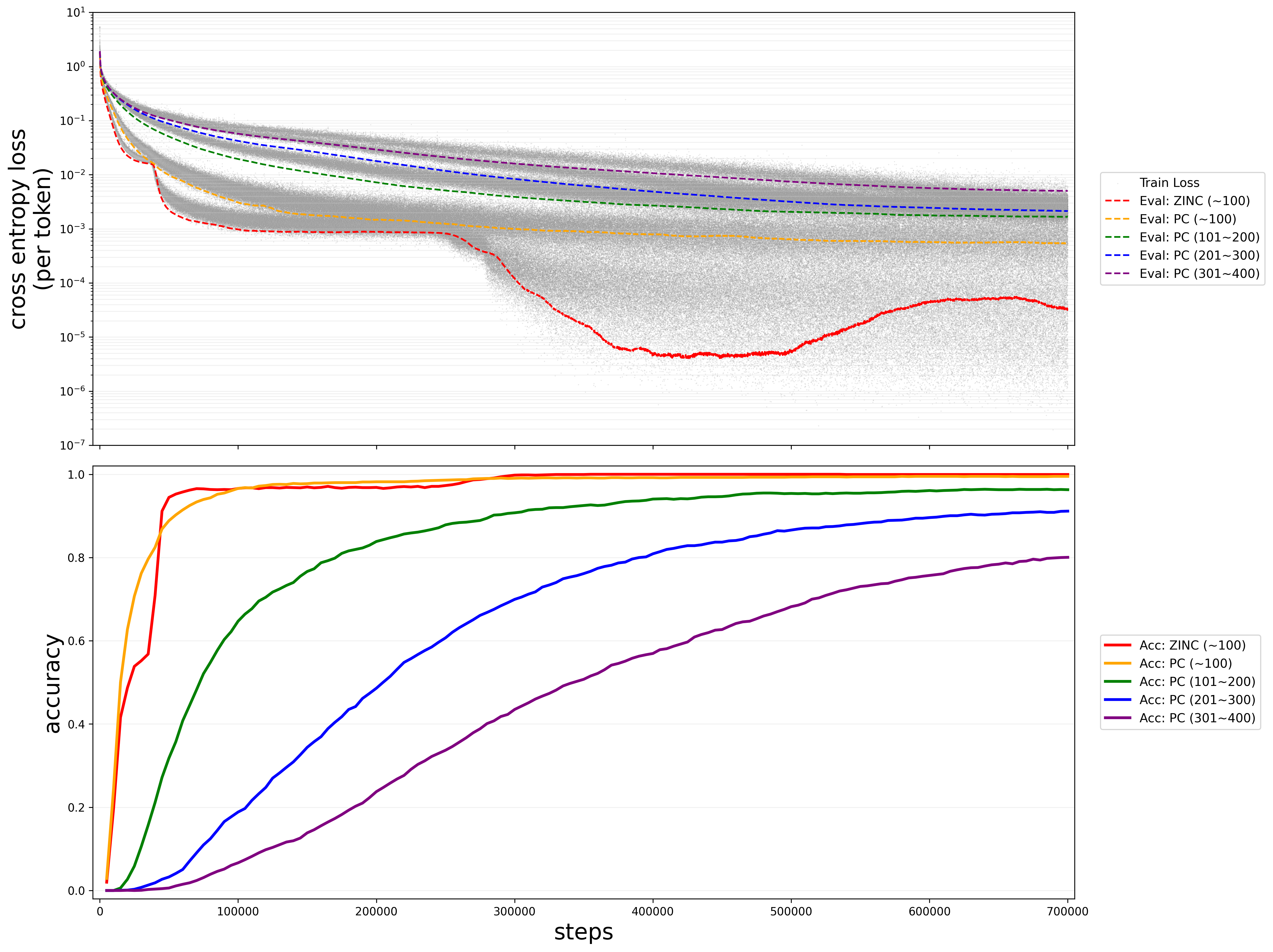}
        \caption{\textit{pancore-addonce}}
        \label{addonceprogress}
    \end{subfigure}
    
    \vspace{1em}
    
    \begin{subfigure}{0.48\textwidth}
        \centering
        \includegraphics[width=\linewidth]{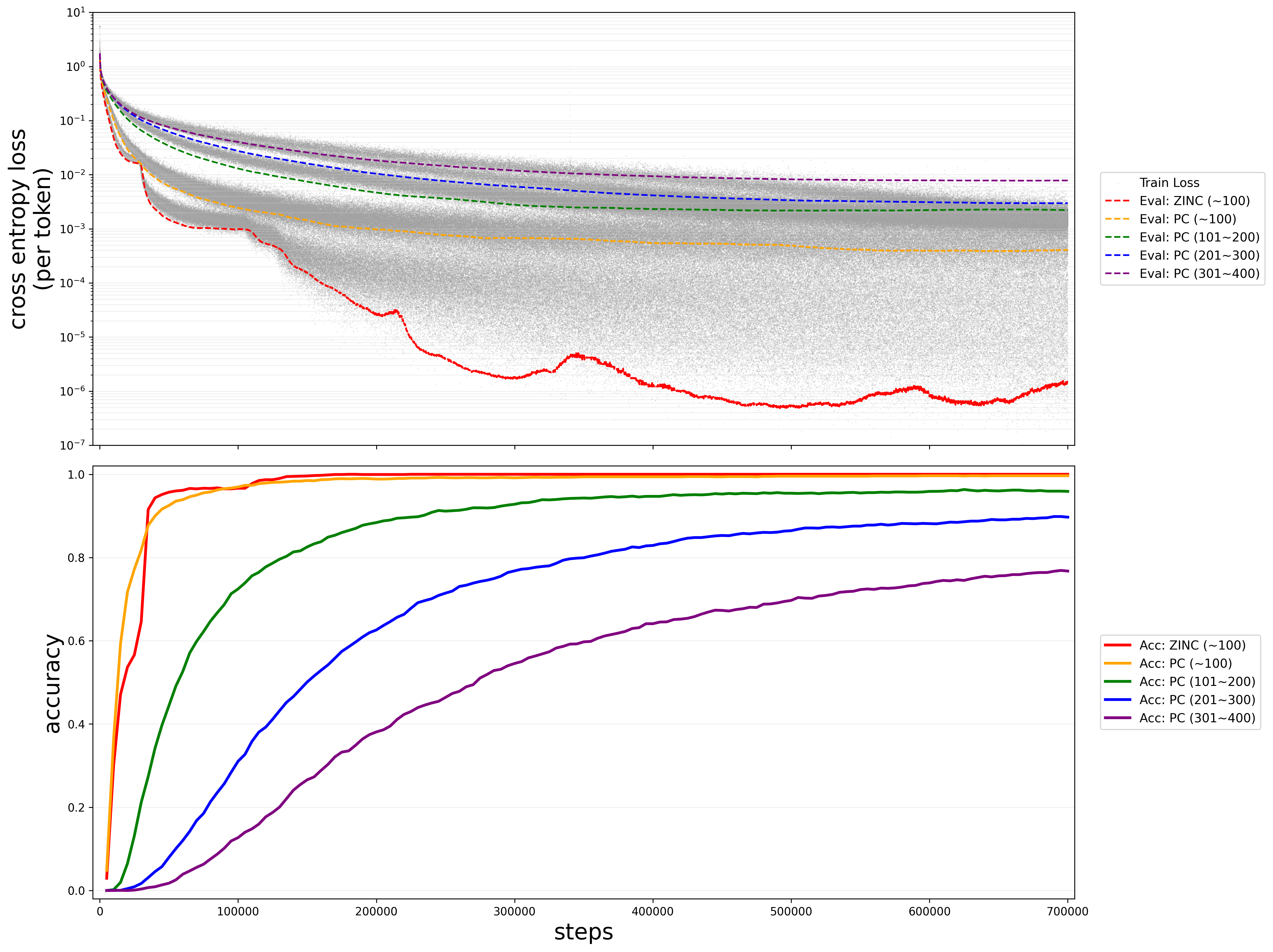}
        \caption{\textit{pancore-xattn-adalnzero}}
        \label{xattnadalnzeroprogress}
    \end{subfigure}
    \caption{Training progress (\subref{baselineprogress}) \textit{pancore-baseline}, (\subref{addonceprogress}) \textit{pancore-addonce}, (\subref{xattnadalnzeroprogress}) \textit{pancore-xattn-adalnzero}. The top row shows the per-token cross-entropy loss for both training and evaluation data across each sequence-length bucket, while the bottom row shows the accuracy for each bucket.}
    \label{progress}
\end{figure}

Across all models, we observed a sudden performance surge — hereafter referred to as the "jump-up" — characterized by an abrupt, discontinuous increase in translation accuracy alongside sharp drops in training and evaluation losses. This phase transition-like behavior is most dramatic in the ZINC20 (\til 100) bucket, where accuracy stagnated near 0.5 before leaping to 0.9. While loss progress suggests a secondary transition occurred later for advanced models (Fig.~\ref{progress}\subref{addonceprogress},\subref{xattnadalnzeroprogress}), our subsequent analysis focuses on the primary jump-up. We utilized \textit{pancore-addonce} and a high-resolution 500-step checkpoint interval for these analyses.

\begin{figure}[H]
    \centering
    \begin{subfigure}{0.48\textwidth}
        \centering
        \includegraphics[width=\linewidth]{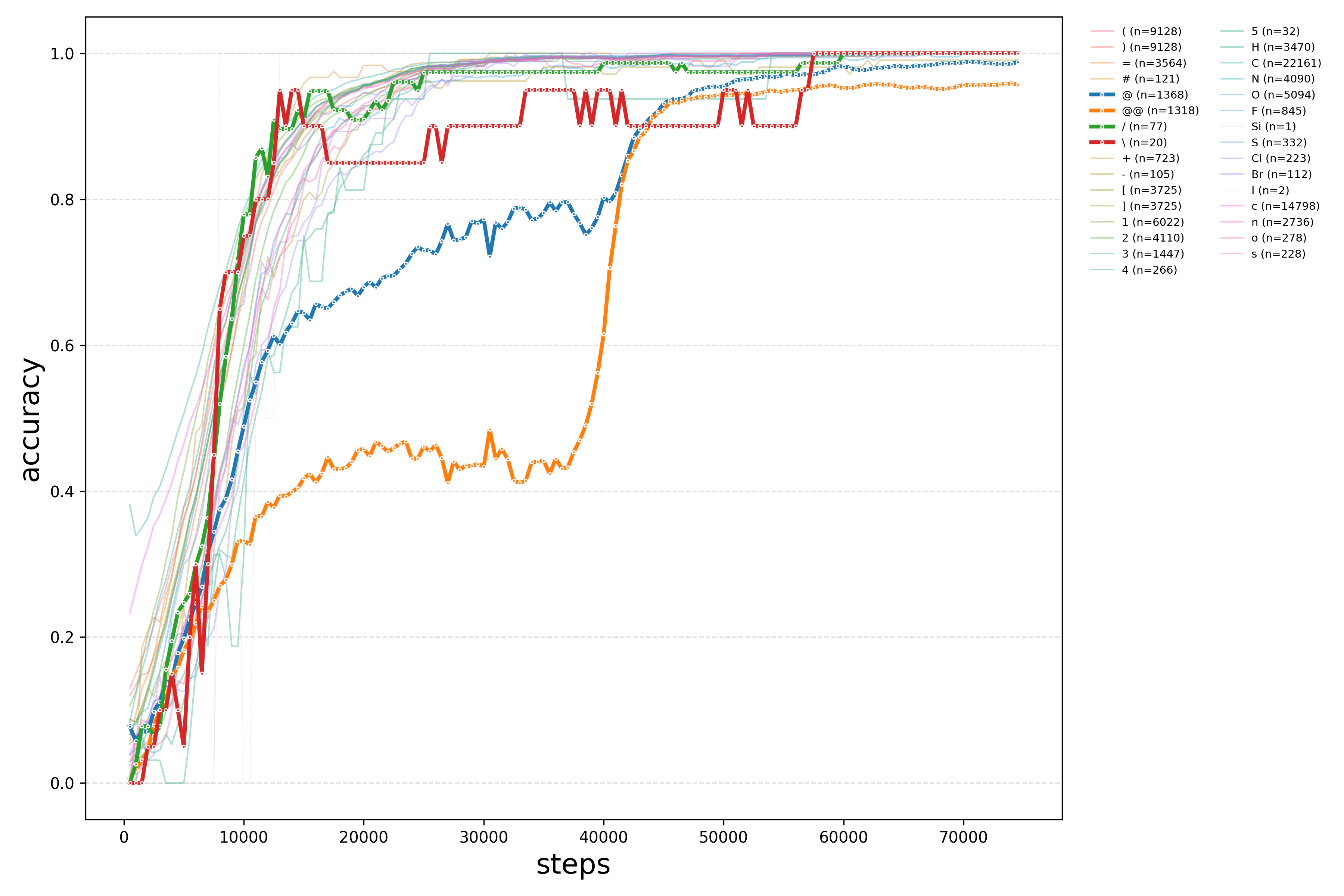}
        \caption{ZINC20 (\til 100)}
        \label{zinctokenacc}
    \end{subfigure}
    \hfill
    \begin{subfigure}{0.48\textwidth}
        \centering
        \includegraphics[width=\linewidth]{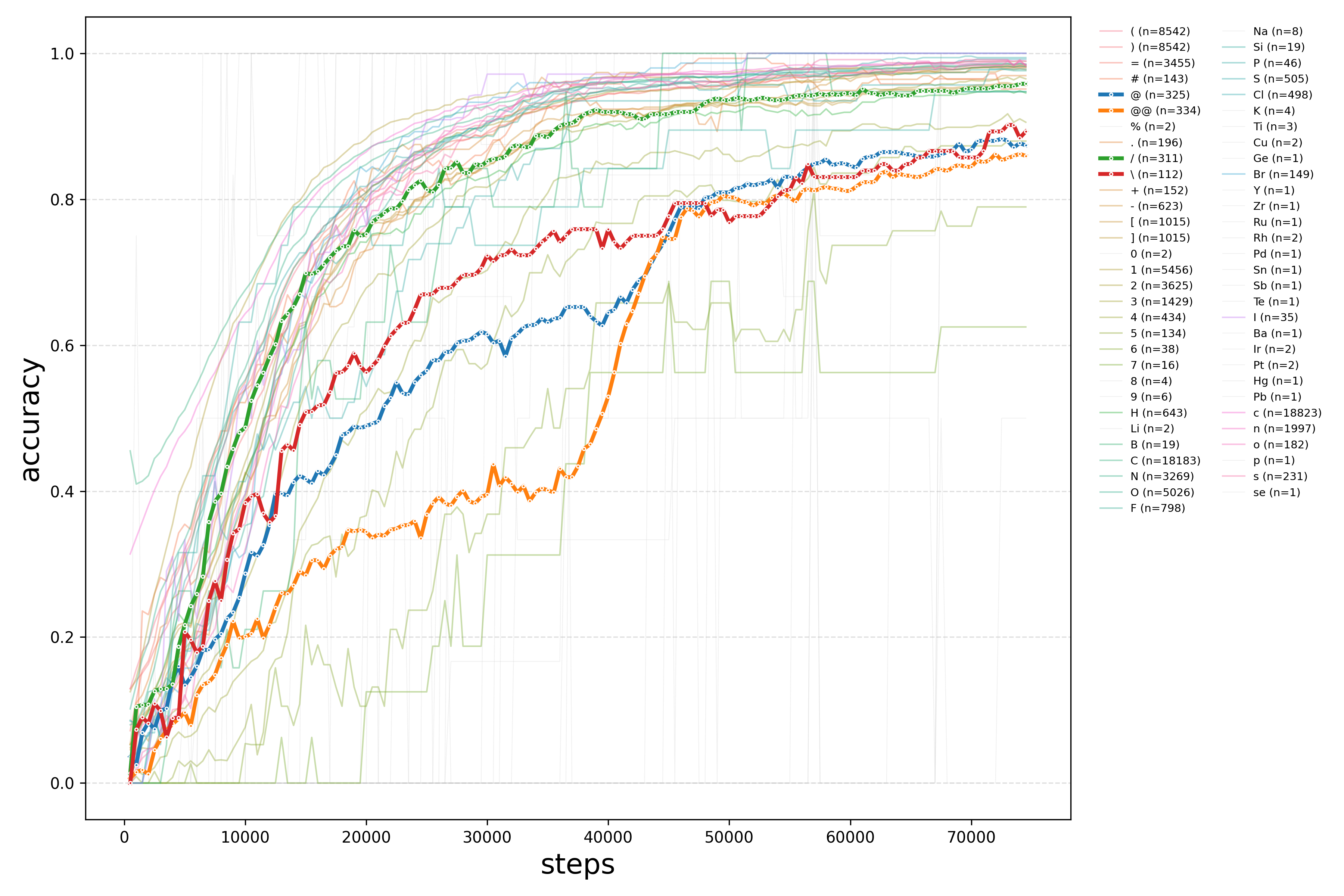}
        \caption{PubChem (\til 100)}
        \label{pubchemtokenacc}
    \end{subfigure}
    \caption{Token accuracy progress of (\subref{zinctokenacc}) ZINC20 (\til 100) and (\subref{pubchemtokenacc}) PubChem (\til 100) for \textit{pancore-addonce}. Highlighted lines indicate stereochemistry tokens; chirality tokens (@ and @@), geometric isomerism tokens (/ and \textbackslash).}
    \label{tokenacc}
\end{figure}

Previous studies indicate that chemical language models often stagnate when learning enantiomeric representations~\cite{Yoshikai2024-ea}. Our cross-checkpoint token accuracy analysis supports this view: chiral token (@, @@) accuracy surged precisely during the jump-up in both ZINC20 and PubChem datasets (Fig.~\ref{tokenacc}). While standard tokens quickly reached high accuracy, chiral tokens plateaued near 0.4–0.5 for over 30,000 steps before spiking. The less dramatic appearance of this surge in the PubChem (\til 100) SMILES overall accuracy is due to the lower absolute frequency of chiral SMILES in that bucket; filtering strictly for chi-SMILES reveals a similar abrupt transition (Fig.~\ref{chiralacc}).

In contrast, related to stereochemistry, accuracy for tokens denoting geometric isomerism (/, \textbackslash) rose smoothly and continuously from the start. This disparity highlights a fundamental difference in grammatical complexity. Geometric isomerism involves a localized, binary syntactical rule mapping adjacent single bonds across a double bond (e.g., C/C=C\textbackslash C). Stereoisomerism, however, dictates the three-dimensional permutation of up to four positional constraints (substituents) distributed throughout the entire molecular graph, based on CIP priority rules. The jump-up transition likely marks the point at which the model becomes able to use enough graph-level context to resolve these chiral-token dependencies.

\begin{figure}[H]
    \centering
    \begin{subfigure}{0.48\textwidth}
        \centering
        \includegraphics[width=\linewidth]{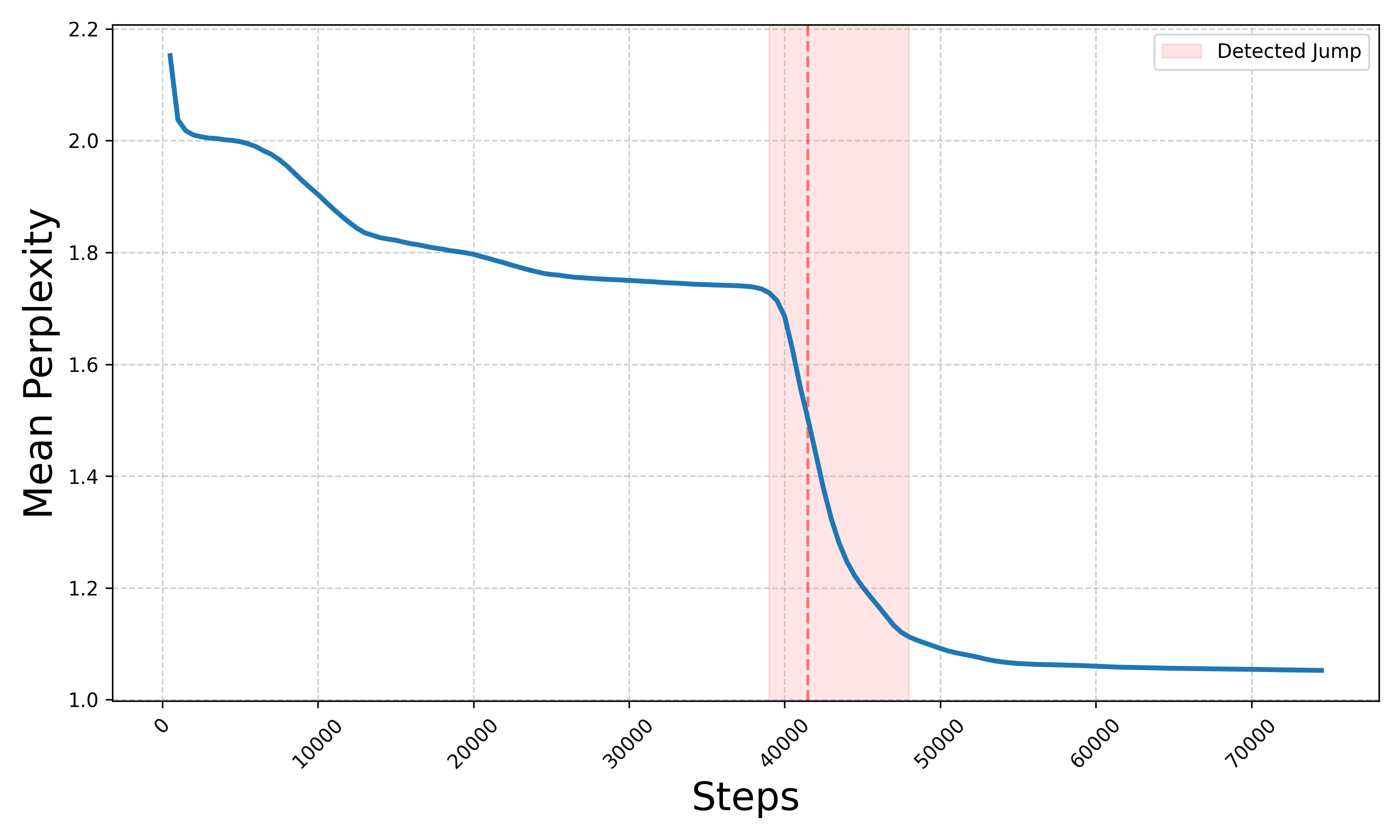}
        \caption{Perplexity}
        \label{perplexity}
    \end{subfigure}
    \hfill
    \begin{subfigure}{0.48\textwidth}
        \centering
        \includegraphics[width=\linewidth]{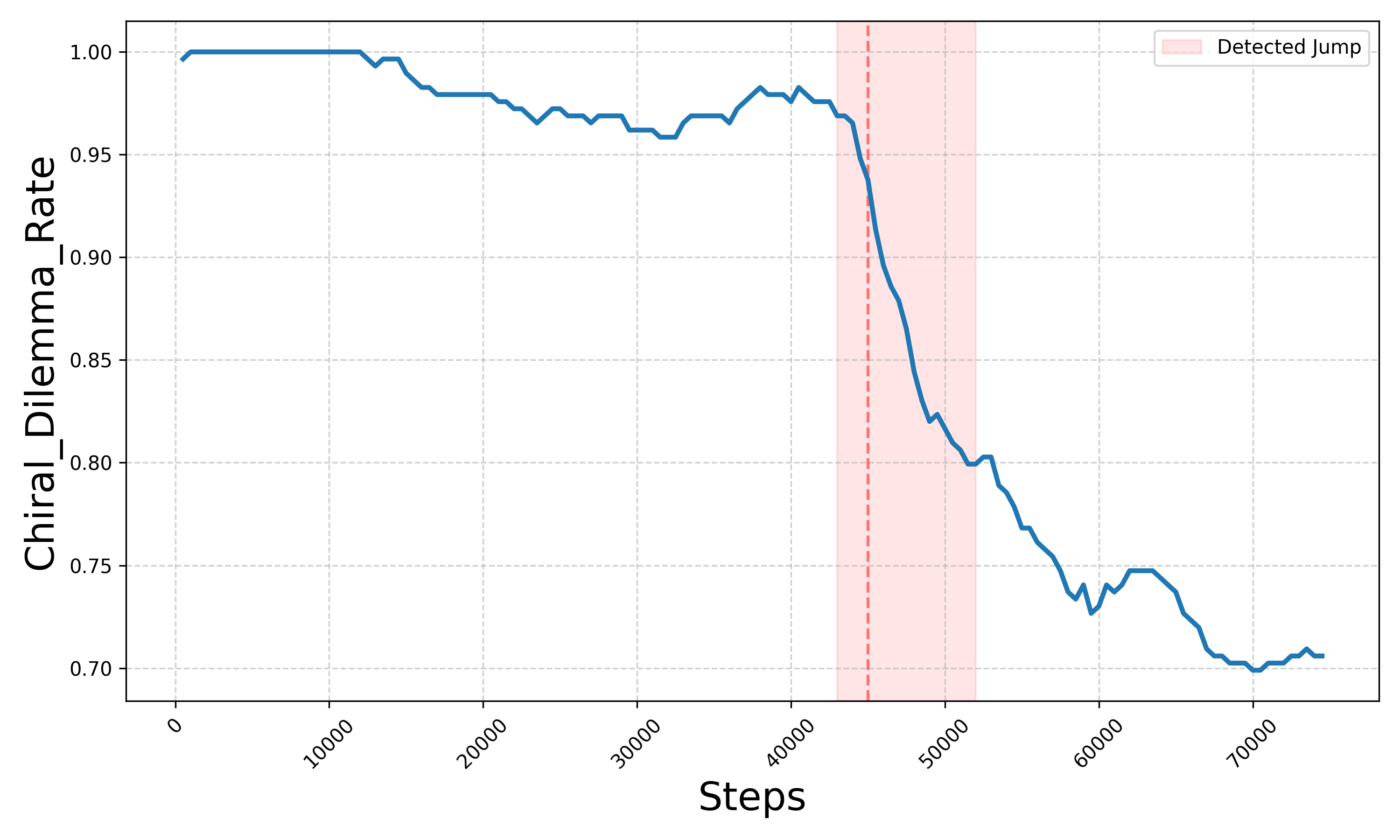}
        \caption{Chiral Dilemma Rate}
        \label{cdr}
    \end{subfigure}
    \caption{Trajectories of logits-based metrics for \textit{pancore-addonce} on the ZINC20 (\til 100) evaluation subset: (\subref{perplexity}) Mean Perplexity at target chiral positions and (\subref{cdr}) Chiral Dilemma Rate (CDR). The vertical red dashed line indicates the training step with the highest rate of perplexity change, and the light red shaded region represents the jump-up interval, defined as the ±15\% window around that step.}
\end{figure}

Logit-level analysis corroborates this phase transition. Within the ZINC20 (\til 100) evaluation subset (200 SMILES randomly sampled), average perplexity at target chiral positions dropped sharply from roughly 1.7 to 1.1 during the jump-up (Fig.~\ref{perplexity}). The Chiral Dilemma Rate (Fig.~\ref{cdr}) also decreased, though it remained relatively high since a sudden surge in primary prediction probability does not necessarily demote the secondary chiral token candidate. These metrics confirm a decisive internal transition enabling the confident generation of chiral information.

\subsection{Reorganization of the Representation Space and Concentration of Attention}
\label{sec:reorganization}
To examine the internal basis of this phase transition, we analyze the model's internal representations during the jump-up interval (defined as the $\pm 15\%$ zone of the maximum perplexity change rate), focusing on the ZINC20 (\til 100) evaluation subset.

\subsubsection*{Attention Mechanism Sharpening}
\begin{figure}[H]
    \centering
    \begin{subfigure}{0.48\textwidth}
        \centering
        \includegraphics[width=\linewidth]{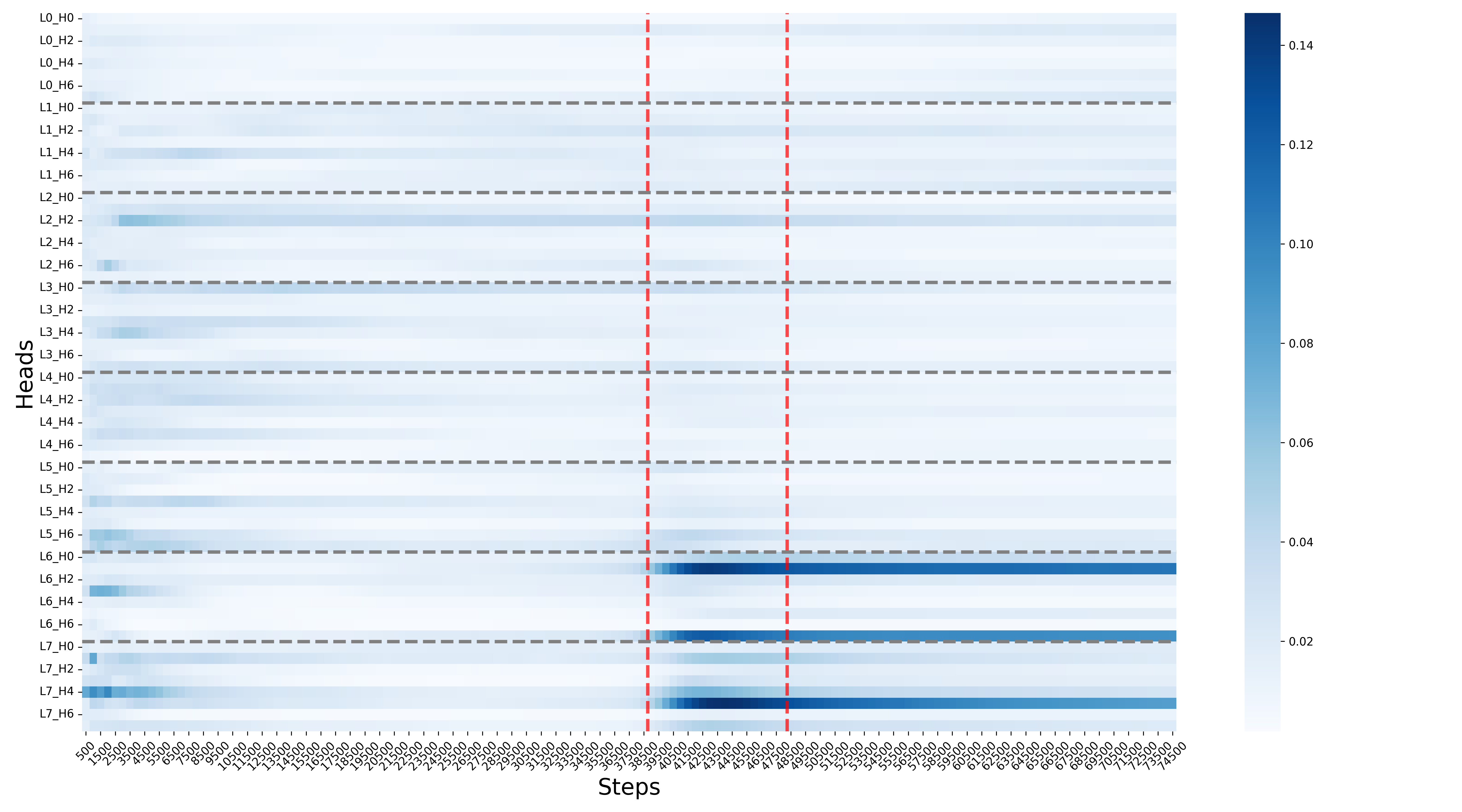}
        \caption{Encoder - all heads}
        \label{attnmassencoderheatmap}
    \end{subfigure}
    \hfill
    \begin{subfigure}{0.48\textwidth}
        \centering
        \includegraphics[width=\linewidth]{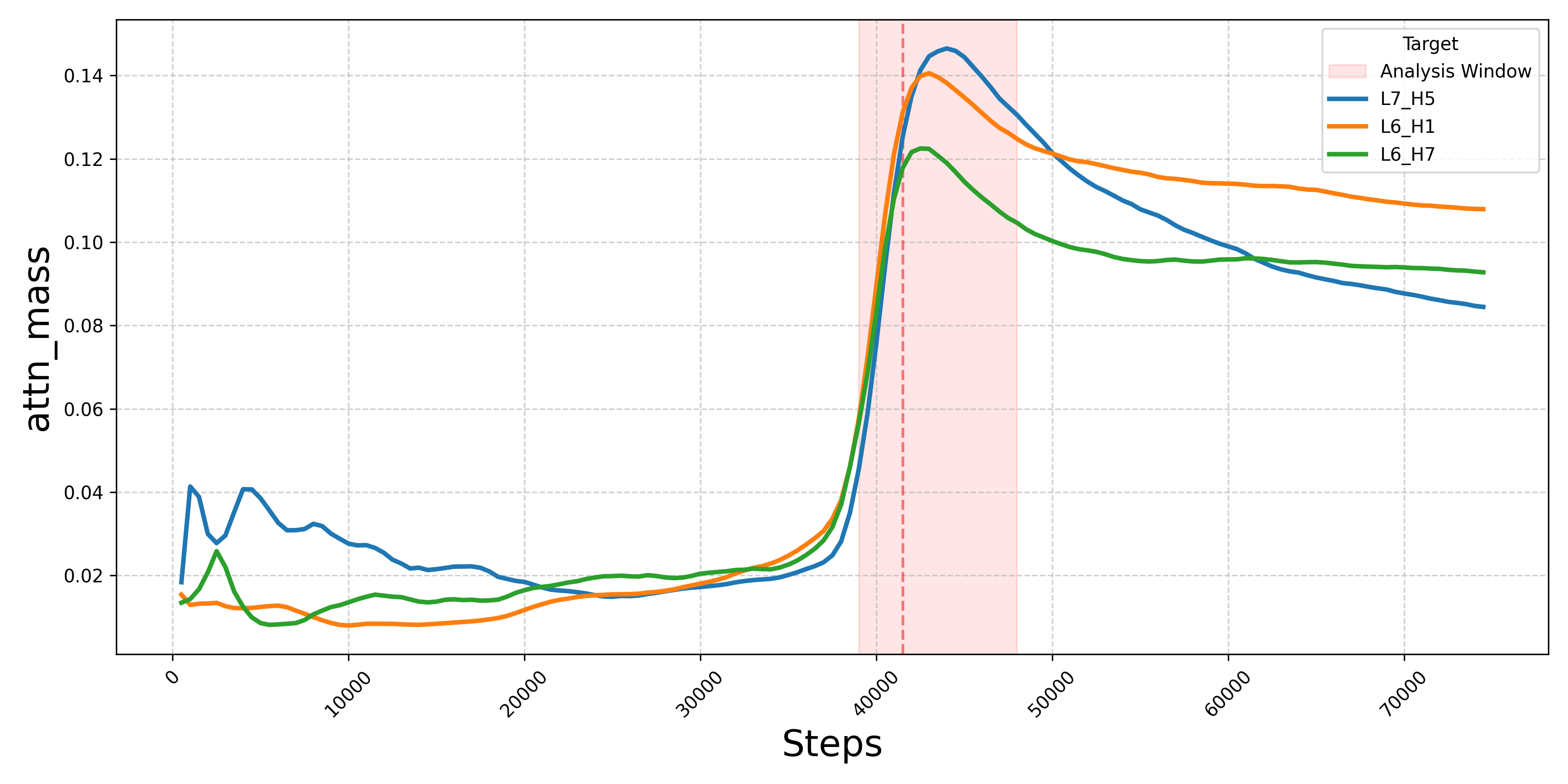}
        \caption{Encoder - Top3 heads}
        \label{attnmassencodertop3}
    \end{subfigure}

    \vspace{1em}

    \begin{subfigure}{0.48\textwidth}
        \centering
        \includegraphics[width=\linewidth]{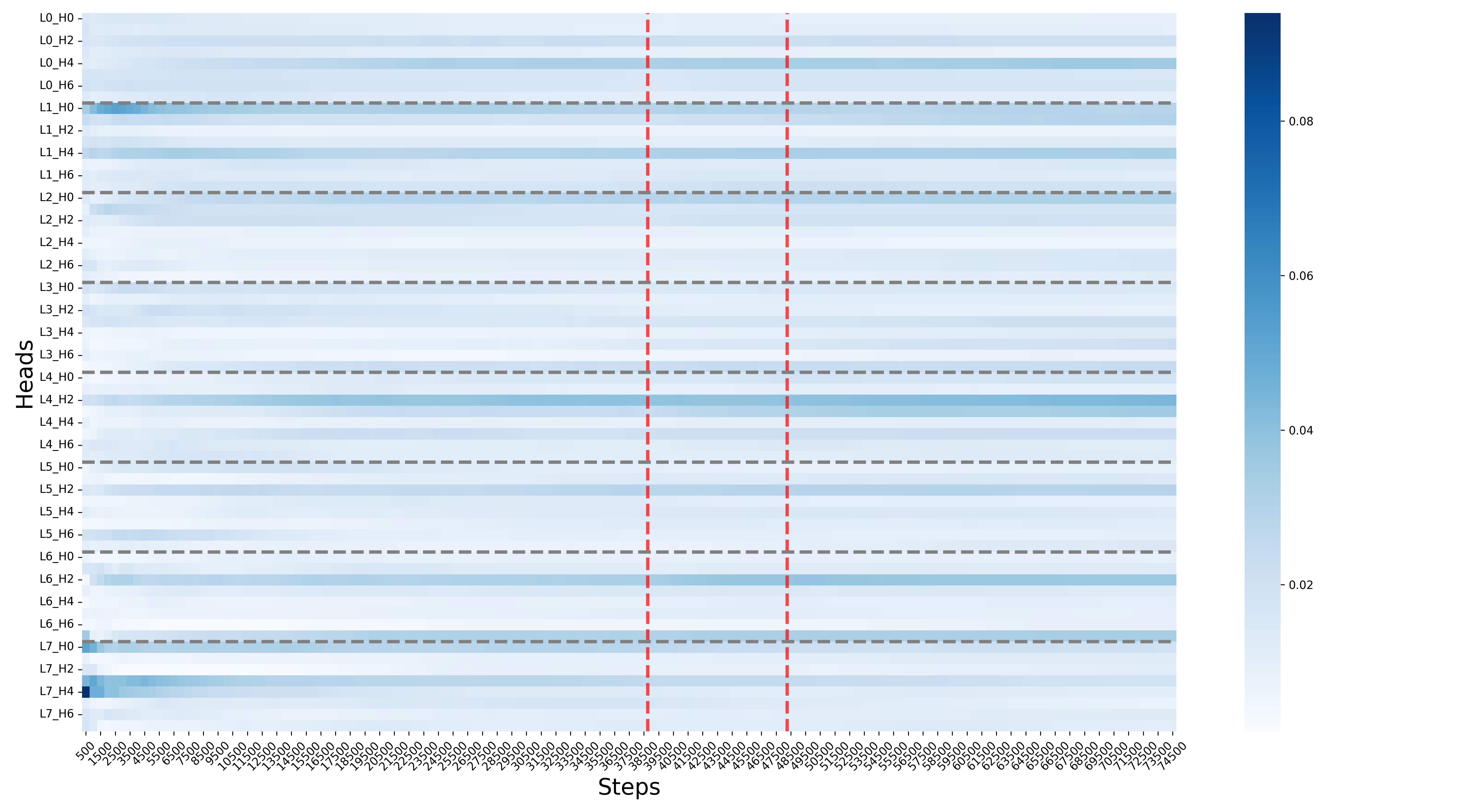}
        \caption{Decoder - all heads}
        \label{attnmassdecoderheatmap}
    \end{subfigure}
    \hfill
    \begin{subfigure}{0.48\textwidth}
        \centering
        \includegraphics[width=\linewidth]{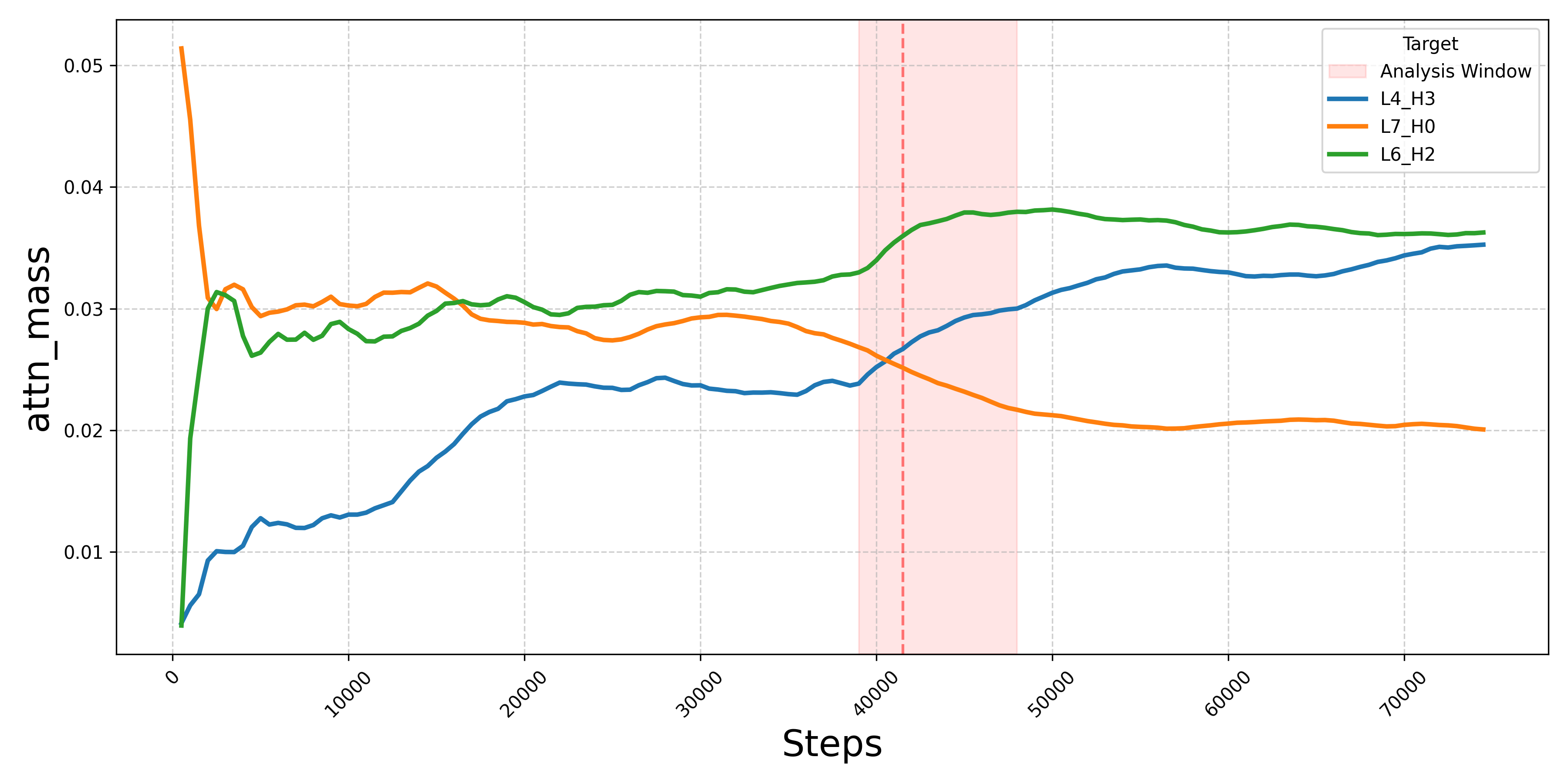}
        \caption{Decoder - Top3 heads}
        \label{attnmassdecodertop3}
    \end{subfigure}
    
    \caption{Trajectories of the chiral tokens' attention weight mass across heads from L0H0 (layer 0, head 0) to L7H7, for \textit{pancore-addonce} on ZINC20 (\til 100). (\subref{attnmassencoderheatmap}, \subref{attnmassencodertop3}) Encoder; (\subref{attnmassdecoderheatmap}, \subref{attnmassdecodertop3}) Decoder. (\subref{attnmassencoderheatmap}, \subref{attnmassdecoderheatmap}) Heatmaps across all heads; The region bounded by the two vertical dashed lines indicates the jump-up interval defined by perplexity. (\subref{attnmassencodertop3}, \subref{attnmassdecodertop3}) Changes of the three heads with the largest variation within the jump-up interval. The red line and the light red shaded region indicate the perplexity-derived jump-up interval.}
    \label{attnmass}
\end{figure}

To investigate the internal mechanisms underlying this sudden performance surge, we first analyzed the dynamics of the attention mechanism. As shown in Fig.~\ref{attnmass}, attention mass toward chiral tokens increases sharply during the jump-up interval in specific deep-layer heads of the encoder (Layers 6, 7). Conversely, decoder attention remained largely static. Furthermore, attention entropy (Fig.~\ref{attnentropy}) in the encoder decreased, indicating that specific heads began focusing intensely on a narrow set of tokens. Graph distance analysis (Fig.~\ref{graphdist}) shows that this concentrated attention is biased toward topologically nearby substructures. Interestingly, this entropy drop occurred in intermediate layers rather than the deep layers where mass increased. 

These observations are consistent with an encoder-centered transition, in which chiral tokens become important targets within the encoder’s attention mechanism. The complementary patterns — entropy reduction concentrated in intermediate layers and attention mass increase in deep layers — suggest a functional differentiation in which intermediate heads may preferentially encode the localized chemical environment surrounding the chiral center, while deep-layer heads appear to consolidate this contextual information into the chiral token representation. We note, however, that these interpretations are inferred from aggregate attention statistics; directly establishing the computational roles of individual heads would require more targeted interventional analyses beyond the scope of the present study.

\subsubsection*{Dynamic Restructuring of Representations in the Residual Stream}
\begin{figure}[H]
    \centering
    \begin{subfigure}{0.48\textwidth}
        \centering
        \includegraphics[width=\linewidth]{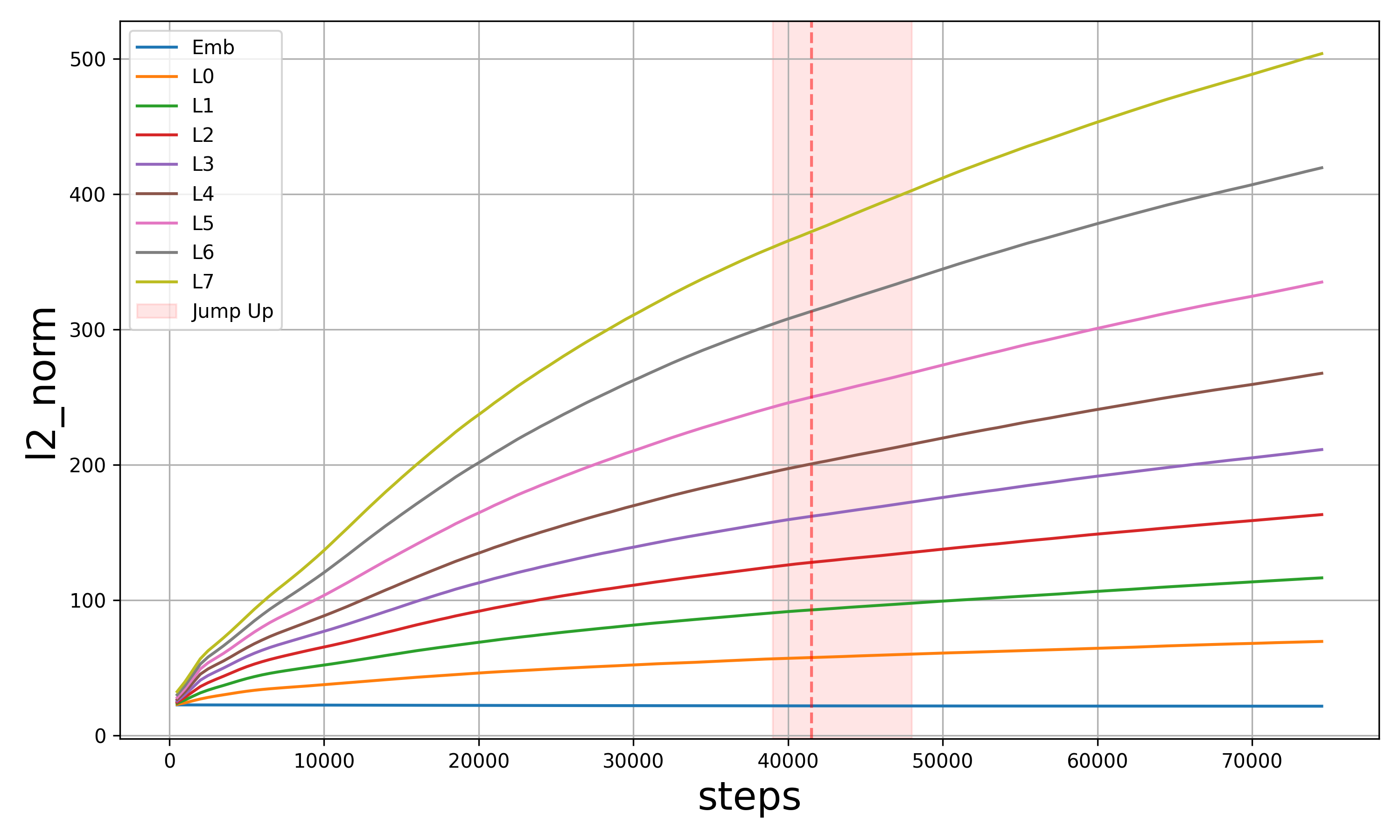}
        \caption{Background - L2 Norm}
        \label{backgroundl2norm}
    \end{subfigure}
    \hfill
    \begin{subfigure}{0.48\textwidth}
        \centering
        \includegraphics[width=\linewidth]{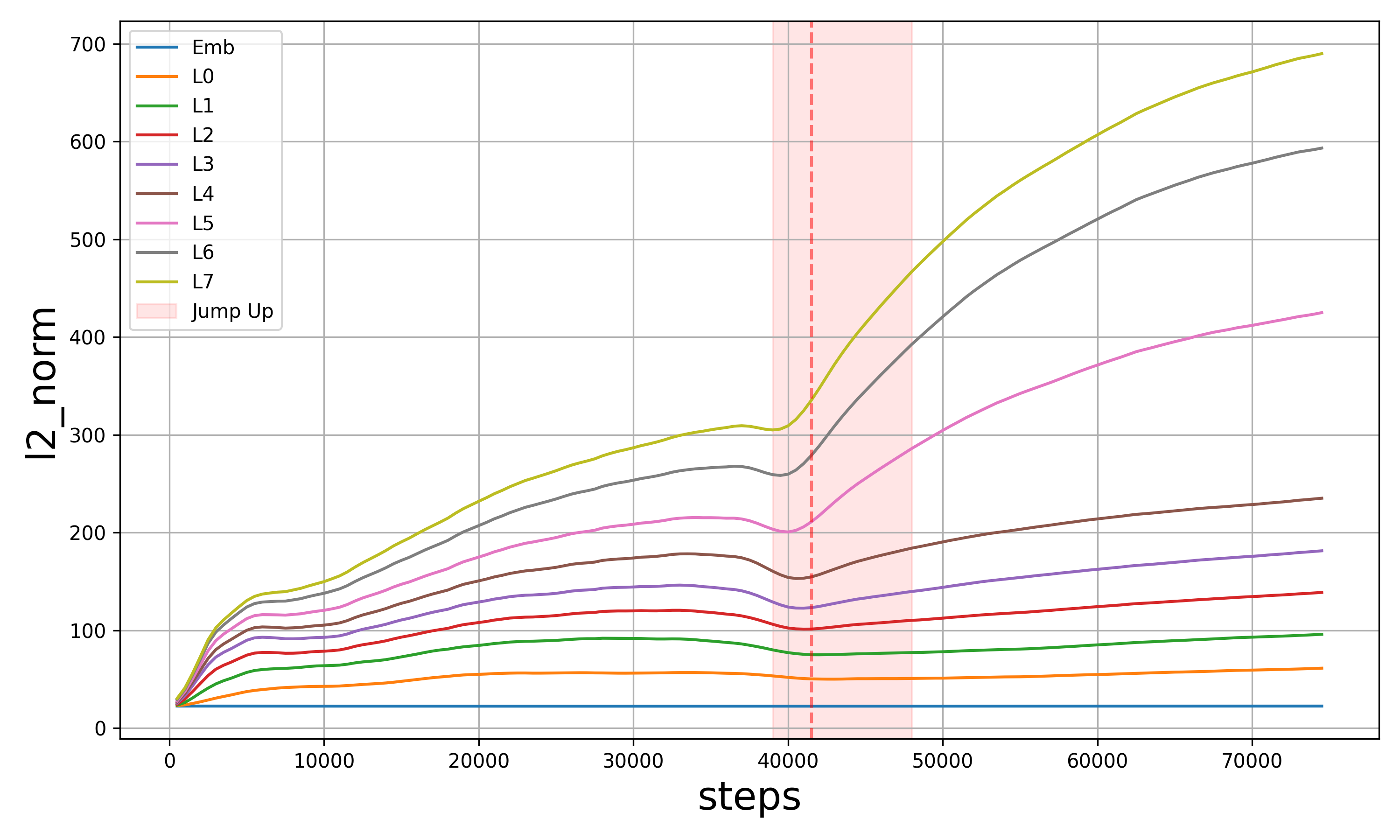}
        \caption{Chiral - L2 Norm}
        \label{chirall2norm}
    \end{subfigure}

    \vspace{1em}

    \begin{subfigure}{0.48\textwidth}
        \centering
        \includegraphics[width=\linewidth]{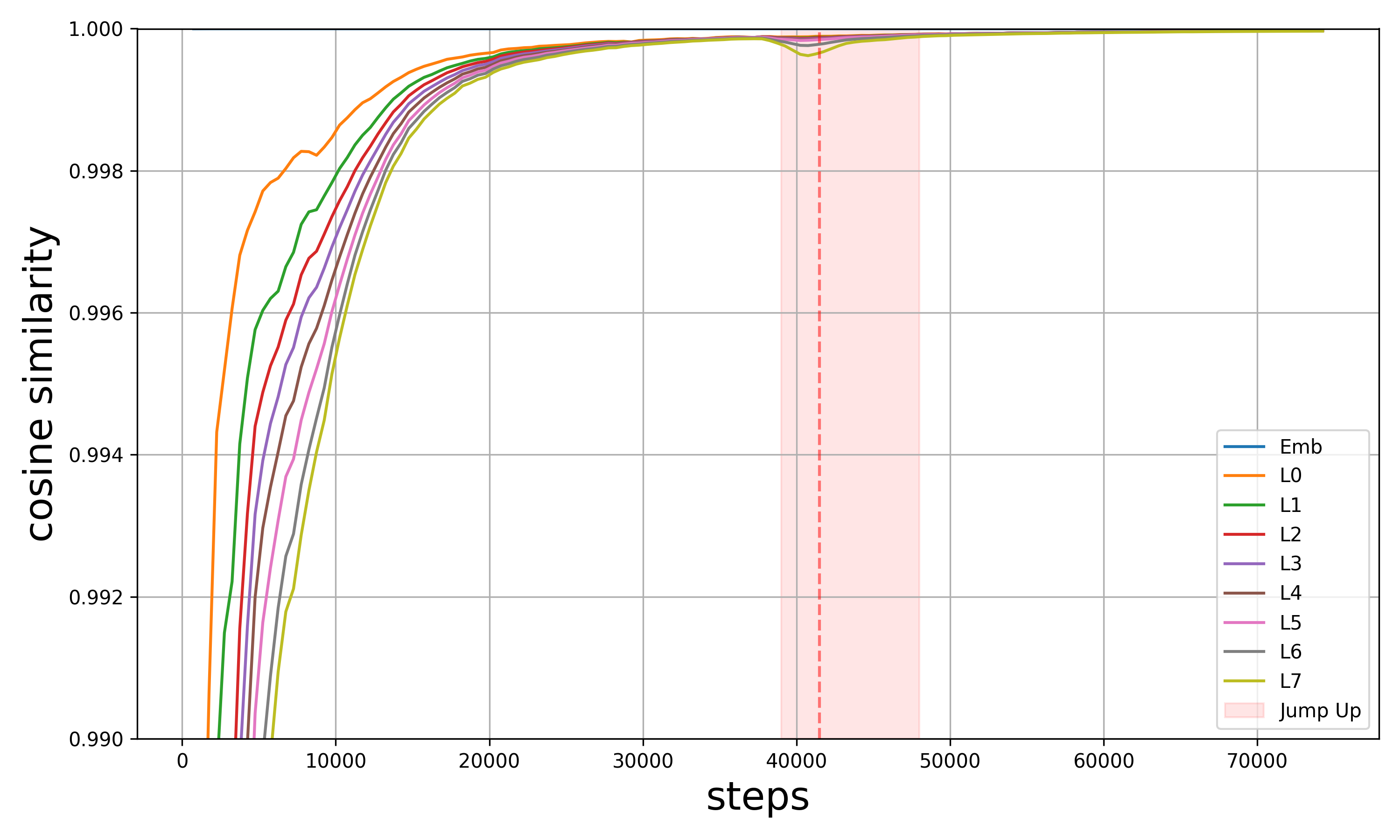}
        \caption{Background - cosine similarity}
        \label{backgroundcos}
    \end{subfigure}
    \hfill
    \begin{subfigure}{0.48\textwidth}
        \centering
        \includegraphics[width=\linewidth]{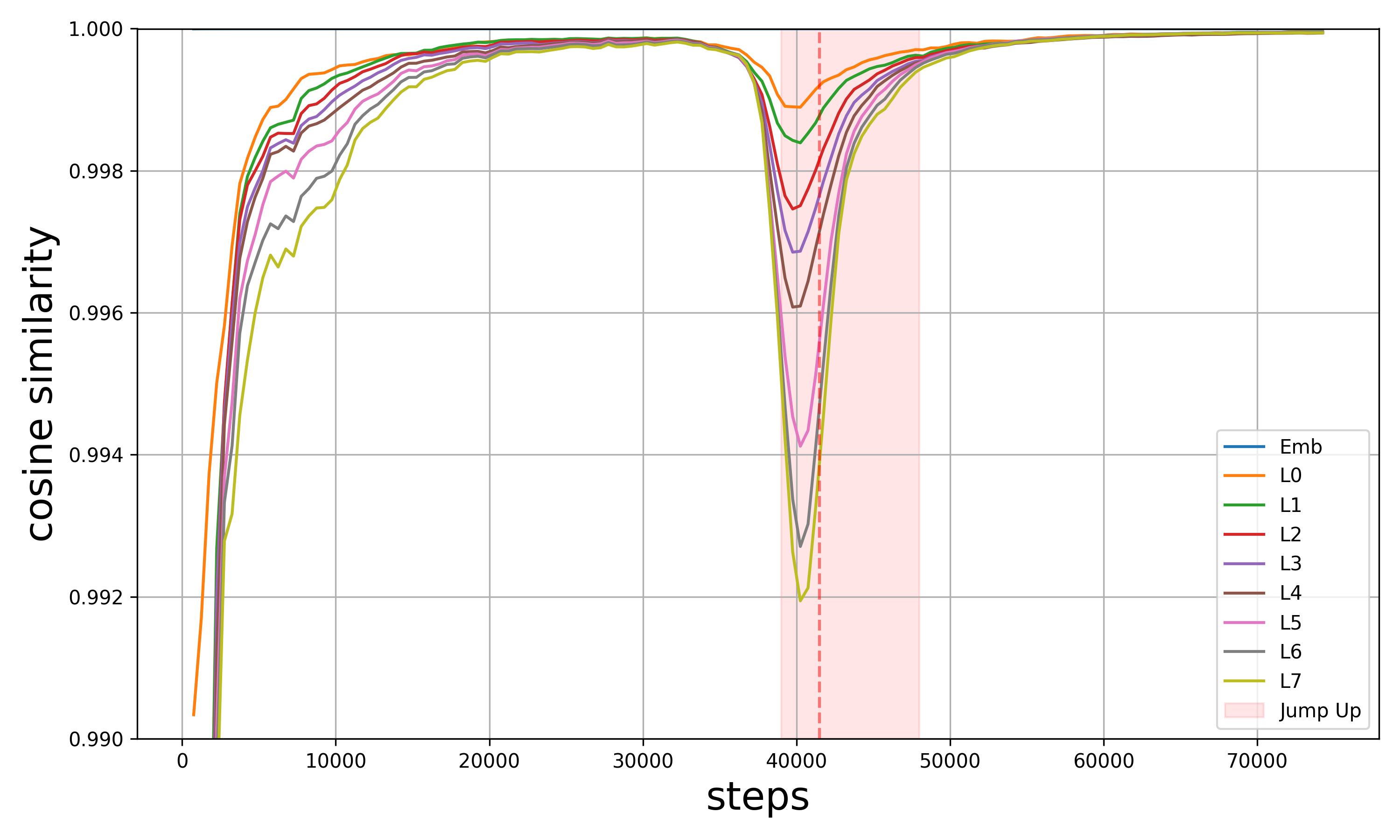}
        \caption{Chiral - cosine similarity}
        \label{chiralcos}
    \end{subfigure}
    
    \caption{Trajectories of residual stream–related metrics from L0 (layer 0) to L7, for \textit{pancore-addonce} encoder on ZINC20 (\til 100). (\subref{backgroundl2norm}, \subref{chirall2norm}) L2 Norm; (\subref{backgroundcos}, \subref{chiralcos}) cosine similarity between consecutive checkpoints. (\subref{backgroundl2norm}, \subref{backgroundcos}) Background tokens excluding chiral tokens. (\subref{chirall2norm}, \subref{chiralcos}) Chiral tokens. The red vertical dashed line and the light red shaded region indicate the perplexity-derived jump-up interval.}
    \label{residualmetrics}
\end{figure}

We next analyzed the residual stream via L2 norm and cosine similarity. Background tokens shows stable, monotonic increases in L2 norm and maintained high cosine similarity (\til 1.0) throughout the jump-up interval (Fig.~\ref{residualmetrics}\subref{backgroundl2norm},\subref{backgroundcos}), implying steady learning of baseline chemical grammar. 

In contrast, chiral tokens display highly distinctive, dynamic trajectories (Fig.~\ref{residualmetrics}\subref{chirall2norm},\subref{chiralcos}). Just prior to the jump-up, the L2 norm of chiral tokens dropped, followed by a sharp V-shaped recovery and surge. The cosine similarity also showed a temporally aligned transient dip. Although absolute values of cosine similarity remained close to unity—an expected consequence of the narrow 500-step checkpoint interval, over which the per-step directional change is inherently small—this drop remarkably exceeded that of background tokens (Fig.\ref{backgroundcos}) and had been more pronounced in absolute terms when evaluated at the coarser 5,000-step resolution (Fig.~\ref{residualcosdelta5000}). While these fluctuations alone do not explicitly prove the internal mechanics, such trajectories are highly analogous to the representational restructuring and phase transitions recently reported in mechanistic interpretability studies of deep neural networks~\cite{Gopalani2025-cb, Billa2026-lt, Chen2023-yt}. Drawing on these parallels, these results suggest a transient, chiral-token-specific reconfiguration of the residual-stream representation, rather than a global restructuring of all token representations. One possible interpretation is the model temporarily destabilizes its previously incomplete representations of chiral tokens — which lacked sufficient chiral context — to reconstruct them into semantically accurate embeddings.

Conversely, the decoder exhibited a qualitatively distinct pattern (Fig.~\ref{decoderresidualmetrics}). The V-shaped L2 norm curve was strictly localized to the final output layer and extended beyond chiral tokens to encompass the entire background token population, consistent with a structural reorganization concentrated immediately before token prediction. The cosine similarity, however, showed a different picture: mild spike-like drops were detectable across all decoder layers for both chiral and background tokens during the jump-up interval. Critically, the magnitude of these cosine similarity fluctuations was substantially smaller than those observed in the encoder — as reflected by the shared y-axis scale between Fig.~\ref{residualmetrics} and Fig.~\ref{decoderresidualmetrics} — suggesting a mild but distributed directional realignment rather than a dramatic representational restructuring. These observations suggest that while the decoder undergoes a distributed, layer-wide reorientation of its internal representations during the phase transition, pronounced structural reorganization in terms of vector magnitude is confined to the output layer.

\subsubsection*{Reorganization of the Latent Space}
\begin{figure}[H]
    \centering
    \begin{subfigure}{0.48\textwidth}
        \centering
        \includegraphics[width=\linewidth]{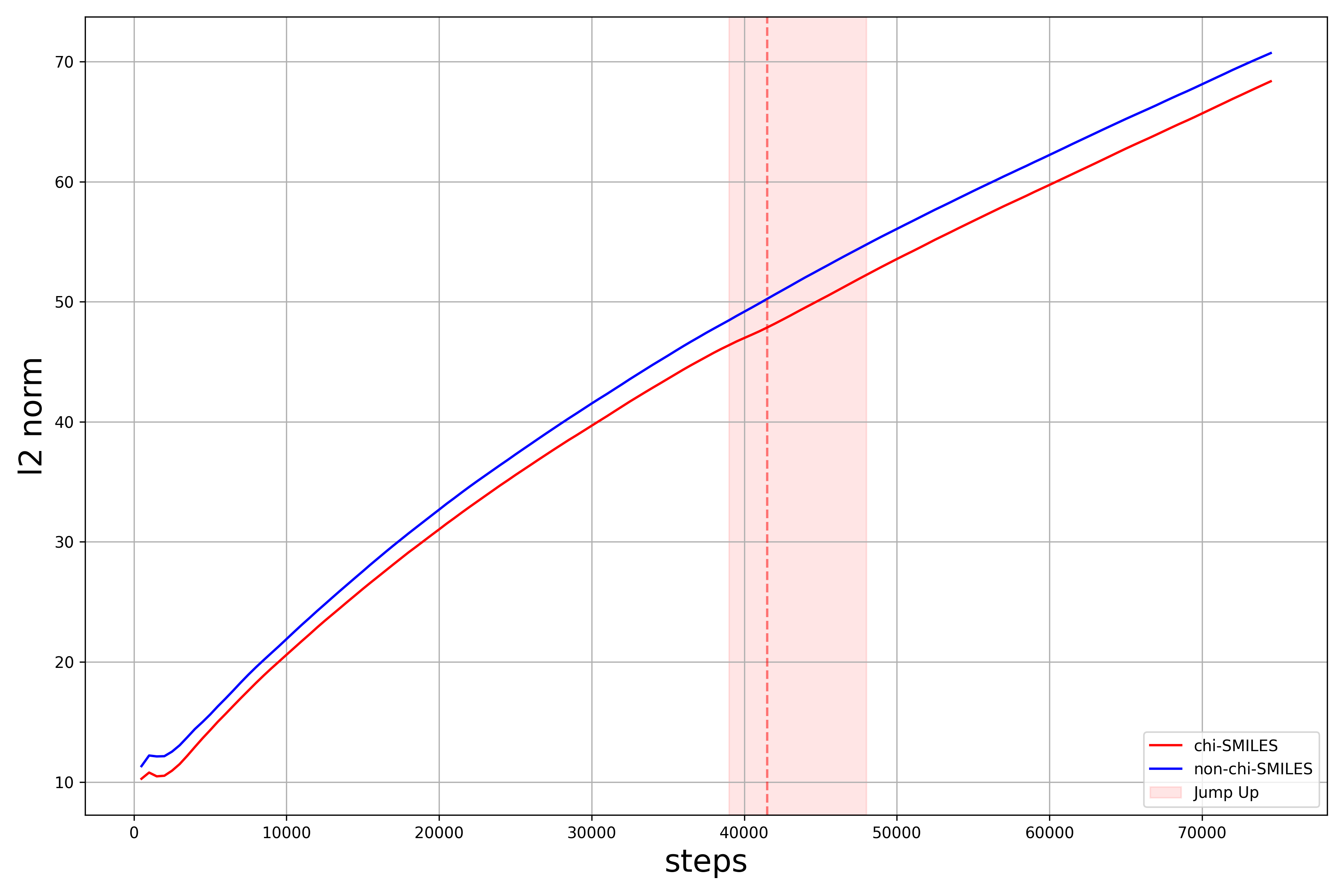}
        \caption{L2 Norm}
        \label{latentl2norm}
    \end{subfigure}
    \hfill
    \begin{subfigure}{0.48\textwidth}
        \centering
        \includegraphics[width=\linewidth]{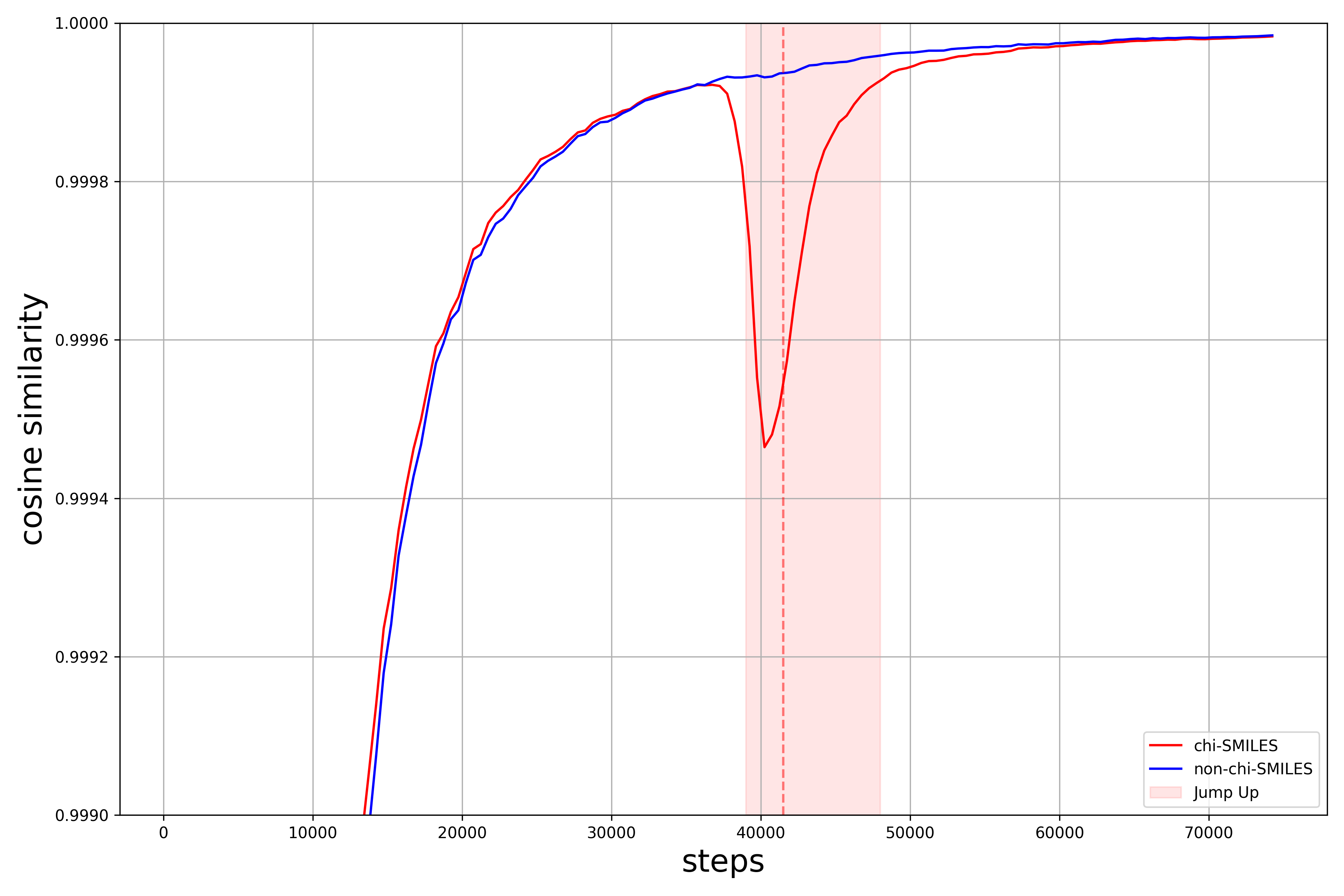}
        \caption{cosine similarity}
        \label{latentcos}
    \end{subfigure}
    
    \vspace{1em}
    
    \begin{subfigure}{0.48\textwidth}
        \centering
        \includegraphics[width=\linewidth]{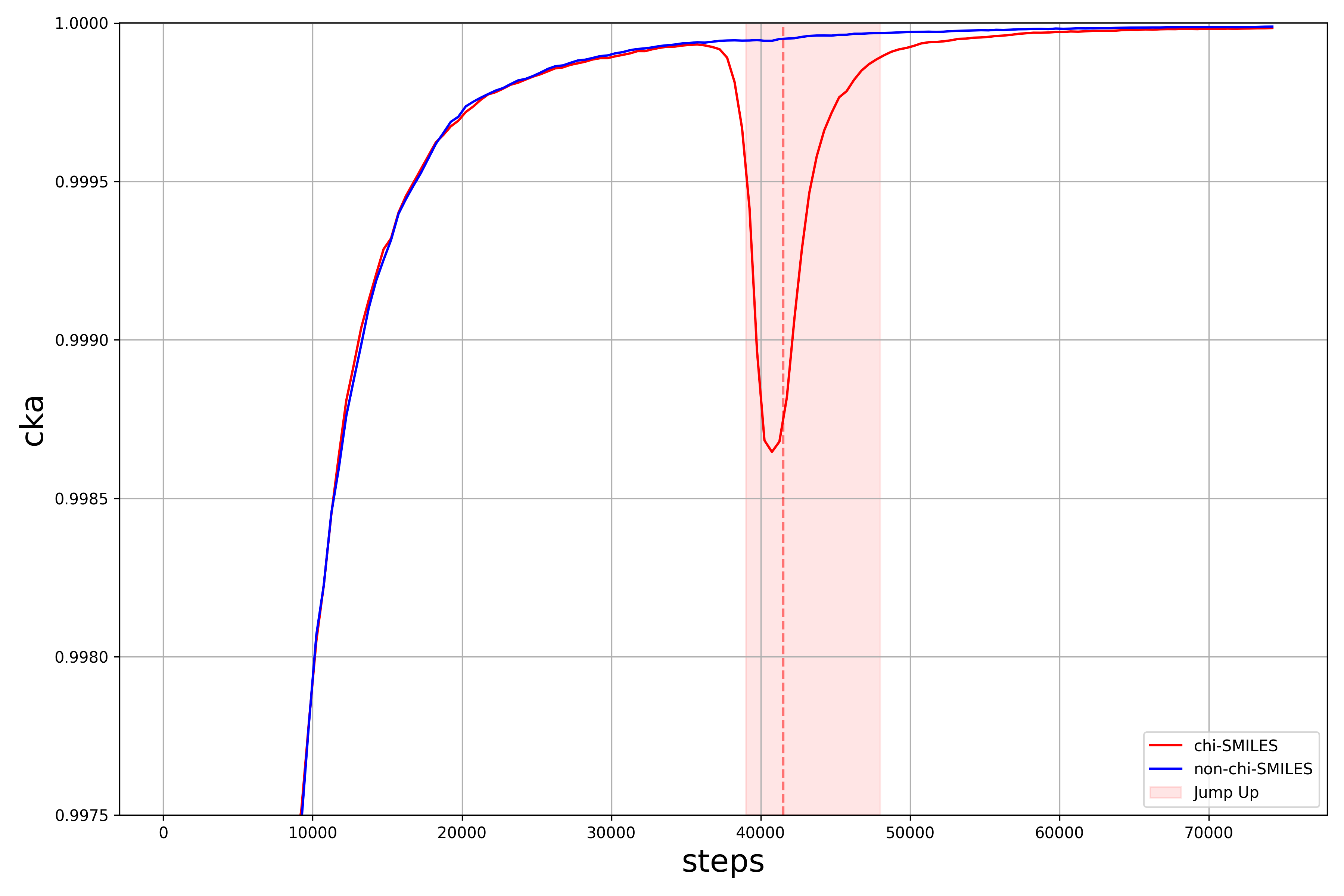}
        \caption{Linear CKA}
        \label{latentcka}
    \end{subfigure}
    \caption{Trajectories of latent vector–related metrics for \textit{pancore-addonce} on ZINC20 (\til 100): (\subref{latentl2norm}) L2 Norm, (\subref{latentcos}) cosine similarity and (\subref{latentcka}) Linear CKA between consecutive checkpoints. Red lines correspond to SMILES containing chiral tokens, while blue lines correspond to SMILES without chiral tokens. The vertical red dashed line and the light red shaded region indicate the perplexity-derived jump-up interval.}
    \label{latentmetrics}
\end{figure}

Finally, we analyzed the latent space bottleneck (Fig.~\ref{latentmetrics}). The L2 norm of latent vectors changed steadily for both chiral and non-chiral SMILES, indicating stable informational capacity. However, while overall directional stability (cosine similarity) and macroscopic spatial structure (Linear CKA) remained high, microscopic yet distinct spike-like fluctuations were observed in cosine similarity and CKA exclusively for chiral SMILES during the jump-up interval—fluctuations that become more pronounced in absolute terms at 5,000-step resolution (Fig.\ref{latentdelta5000}).

Because CKA measures the structural similarity of relative vector arrangements, these exclusive fluctuations imply a fundamental reorganization of topological arrangement principles within the latent space for chiral molecules. We infer that the essence of this reorganization lies in a fundamental shift of the representational basis. Prior to the phase transition, the latent space successfully captured baseline chemical grammar and local topological features, but lacked the capacity to explicitly resolve complex chiral dependencies. However, during the phase transition, the model integrates global chiral constraints—extracted via the attention mechanism and subsequent FFN—into this vector space. Thus, the latent space evolves to describe more holistic chemical structural entities. The continuous stability of the L2 norm corroborates that the latent space maintains its informational capacity, fundamentally changing only the quality of the representations it holds.

This reorganization of the latent space substantially alters the latent vectors transmitted to the decoder. As described above, the decoder displays mild cosine similarity fluctuations distributed across all its layers alongside a more pronounced L2 norm change confined to the final output layer. This pattern indicates that the decoder's response to the newly enriched latent representations involves two concurrent processes: a distributed, layer-wide reorientation of representational direction throughout the network, and a more localized restructuring of vector magnitude at the output stage. The V-shaped L2 norm fluctuations observed across all token types specifically in the final decoder layer are consistent with a global recalibration of the translational context that harmonizes the newly emerged chiral semantics with established chemical grammar immediately before token prediction. Notably, the relatively modest magnitude of these decoder-side changes compared to the dramatic restructuring in the encoder is consistent with the interpretation that the decoder plays a subordinate, adaptive role during the phase transition, rather than being its initiating source.

\subsection{Early Decoder Maturation and Encoder-Side Emergence of Chiral Semantics}
\label{sec:earlydecodermaturation}
To test whether this transition is mainly encoder-centered, we perform a cross-evaluation, swapping the encoder and decoder units from before the transition (39,000 steps) and after the transition (48,000 steps).

Table ~\ref{encdeccrossaccuracy} supports this interpretation. The Pre-to-Post configuration (immature encoder, mature decoder) yields a chi-SMILES Accuracy of 0.4852, showing virtually no improvement over the baseline Pre-to-Pre setup (0.4549). Conversely, the Post-to-Pre configuration (post-transition encoder, pre-transition decoder) increases accuracy to 0.6669. Strikingly, the pre-transition decoder—incapable of resolving chirality on its own—succeeded simply because it was fed a semantically enriched latent vector by the mature encoder. Furthermore, while similar results were consistently confirmed across other models for the ZINC20 (\til 100) dataset, this clear trend was not universally observed in the PubChem datasets (Table ~\ref{encdeccrossaccuracyfull}). We speculate that this discrepancy arises from the significantly higher structural diversity and longer sequence lengths inherent to PubChem, which likely prevent the latent vector space from sufficiently stabilizing during this specific early phase of training.

These results support the view that the primary transition is mainly encoder-centered. The discontinuous leap in performance occurs when the encoder learns to extract semantic constraints and project them into the latent space. The decoder appears to acquire much of its syntactic generation ability early in training, while accurate chiral-token generation depends on the encoder providing latent representations that contain sufficient chiral information.

\subsection{Robustness Evaluation of chiral Recognition via Head Ablation}
\label{sec:headablation}
Finally, we investigate whether this specialized chiral sensitivity remains localized or disperses in the fully trained model (700,000 steps). We performed an attention ablation experiment, forcibly masking the outputs of the top three chiral heads identified in Section~\ref{sec:reorganization} (L6H1, L6H7, L7H5). As a control, we ablated three random non-chiral heads from the same layers.

Table ~\ref{headablation} shows the results. Ablating the three chiral-sensitive heads reduces chi-SMILES accuracy from from 0.9997 to 0.8155 in ZINC20, and from 0.9975 to 0.7826 in PubChem. In contrast, ablation of three random heads causes negligible drops. Crucially, the "Mistranslated chi-SMILES Acc. w/o '@'" metric remained robustly high (>0.8) even when chiral heads were removed. While the magnitude of the ablation impact varied, similar results were consistently confirmed across the other models (Table ~\ref{headablationfull}).

These results are consistent with the targeted heads being selectively important for chiral-token processing, rather than regulators of general chemical grammar, at least for high confidence short-sequence SMILES. The persistence of this functional specialization in the fully trained model suggests that the chiral competence of Pan-CORE is not a transient artifact of the training trajectory, but an acquired capability consolidated into identifiable network components. These findings raise the possibility that emergence phenomena in chemical language models may more broadly promote the formation of functionally specialized computational units; whether this constitutes a general principle of chemical representation learning, however, remains to be examined across diverse architectures and molecular representation tasks.

\section{Conclusion}
\label{sec:conclusion}
In this work, we presented Pan-CORE, an autoregressive Transformer-based encoder-decoder with a bottleneck layer for SMILES translation, and provided a mechanistic account of how SMILES-level chiral information is learned during training. By tracking internal representations at high temporal resolution, we identified a discontinuous phase transition — the "jump-up" — in which accurate chiral token generation arises abruptly following a prolonged plateau. Analyses of attention dynamics, residual-stream trajectories, latent-space geometry, encoder-decoder cross-evaluation, and targeted head ablation support the view that this transition is mainly encoder-centered, with a small set of chiral-sensitive attention heads that contribute selectively to chiral-token resolution.

Several limitations of the present study should be noted. First, the training loss curves of the more expressive model variants suggest a possible secondary jump-up associated with further gains in chiral accuracy (Fig.~\ref{chiralaccdelta5000}); however, the limited number of evaluation samples that shift from incorrect to correct translation during this later stage prevented a systematic mechanistic analysis, leaving its nature open. Second, our analytical framework relies on the abrupt nature of chiral learning: the jump-up provides a clear temporal marker for mechanistic analysis, whereas properties learned more gradually, such as geometric isomerism, require different approaches. Third, our analysis is based on SMILES with a grammar-based tokenization scheme. Thus, our conclusions should be understood as evidence for SMILES-level chiral-token resolution, not as proof that the model explicitly applies human-defined stereochemical rules. Whether similar events occur with other molecular representations or tokenization strategies remains to be tested. 

Despite these limitations, our findings provide a mechanistic view of how CLMs can learn chemical semantic constraints. The identification of a discrete representational change together with chiral-sensitive attention heads suggests that semantic learning in CLMs can be linked to specific changes in internal representations and network components. Because chirality is a key determinant of biological activity, clarifying how AI models represent and resolve chiral constraints provides a step toward more transparent and reliable CLMs for drug discovery and materials development.

\section*{Availability}
The source code will be made publicly available upon publication.

\section*{Acknowledgement}
We thank all those who contributed to the construction of the following data sets employed in the present study such as ZINC20 and PubChem. This work used computational resources of supercomputer Miyabi provided by Joint Center for Advanced High Performance Computing (JCAHPC), University of Tsukuba and The University of Tokyo.

During the preparation of this work, the authors used ChatGPT, Gemini and Claude in order to improve the English grammar and phrasing. After using these services, the authors reviewed and edited the content as needed and take full responsibility for the content of the publication.

\section*{Funding}
This work was supported by AMED under Grant Number JP22mk0101250h, 23ak0101199h0001, and 25ak0101199h0003, and JST BOOST under Grant Number JPMJBY24H2.

\section*{Author Contributions}
\begin{table}[H]
    \begin{tabular}{ll}
        Conceptualization & Tadahaya Mizuno \\
        Methodology & Zehao Li, Yasuhiro Yoshikai, Shumpei Nemoto \\
        Software & Zehao Li, Yasuhiro Yoshikai, Shumpei Nemoto \\
        Investigation & Zehao Li \\
        Resources & Tadahaya Mizuno \\
        Visualization & Zehao Li \\
        Supervision & Hiroyuki Kusuhara, Tadahaya Mizuno \\
        Project administration & Tadahaya Mizuno \\
        Funding acquisition & Tadahaya Mizuno \\
        Writing – original draft & Zehao Li, Tadahaya Mizuno \\
        Writing – review \& editing & Hiroyuki Kusuhara, Tadahaya Mizuno
    \end{tabular}
\end{table}

\section*{Competing Interests}
The authors declare that they have no conflicts of interest. 

\bibliographystyle{unsrtnat}
\bibliography{paperpile}  






\clearpage
\section*{Tables and Figures}

\begin{table}[H]
    \centering
    \caption{Vocabulary and categorization of tokens used for SMILES tokenization.}
    \includegraphics[width=0.85\linewidth]{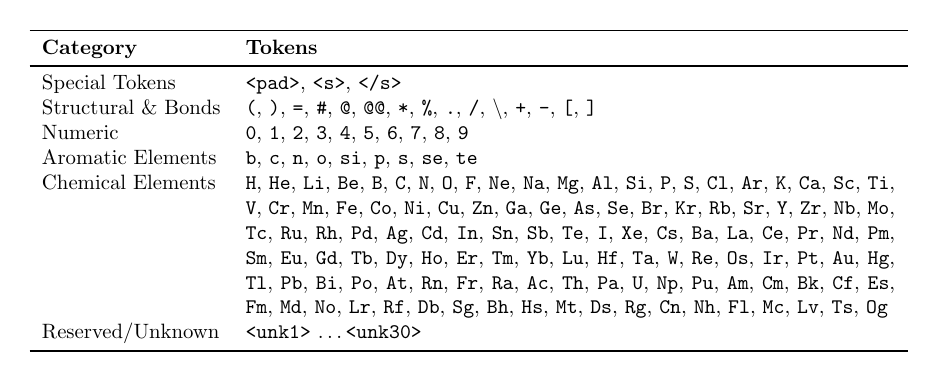}
    \label{tokentable}
\end{table}

\newpage
\begin{table}[H]
    \centering
    \caption{Translation accuracy of the evaluated models across different sequence length buckets for both evaluation and training datasets.}
    \includegraphics[width=0.85\linewidth]{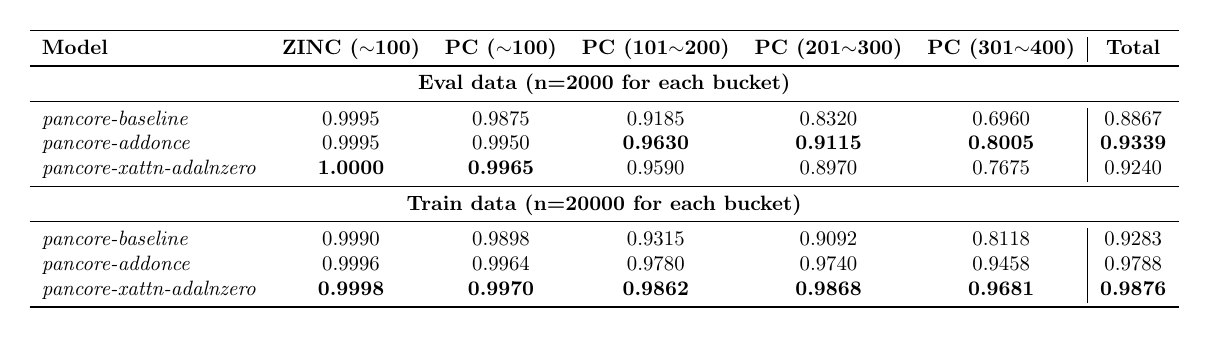}
    \label{genaccuracy}
\end{table}

\newpage
\begin{table}[H]
    \centering
    \caption{Cross-evaluation of translation accuracy for \textit{pancore-addonce} using pre-transition (step 39,000) and post-transition (step 48,000) encoder-decoder configurations. PC (\til 100) represents PubChem (\til 100). Results for all three model variants across all sequence-length buckets are provided in Table \ref{encdeccrossaccuracyfull}.}
    \includegraphics[width=0.85\linewidth]{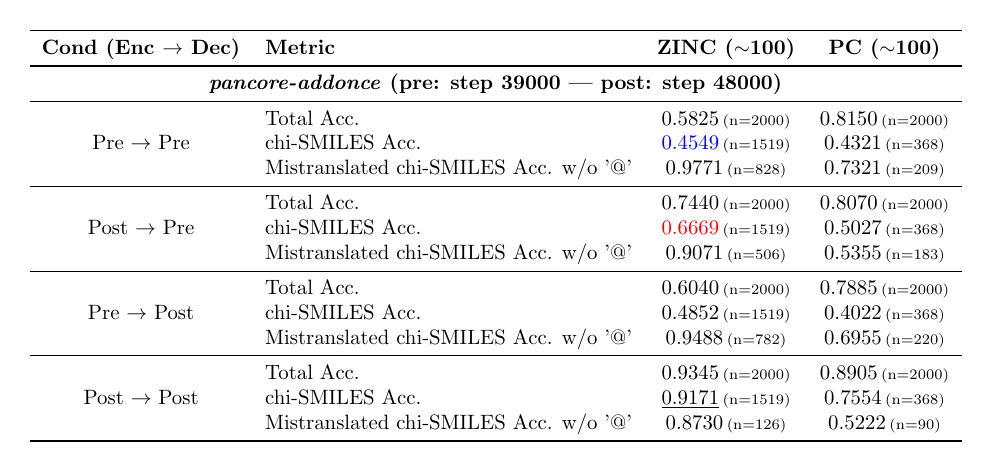}
    \label{encdeccrossaccuracy}
\end{table}

\newpage
\begin{table}[H]
    \centering
    \caption{Robustness evaluation of chiral recognition via attention head ablation pancore-addonce (pre-transition: step 39,000; post-transition: step 48,000; final step 700,000) across pre-transition, post-transition, and final training stages. PC (\til 100) represents PubChem (\til 100). Results for all model variants and all sequence-length buckets are provided in Table \ref{headablationfull}.}
    \includegraphics[width=0.85\linewidth]{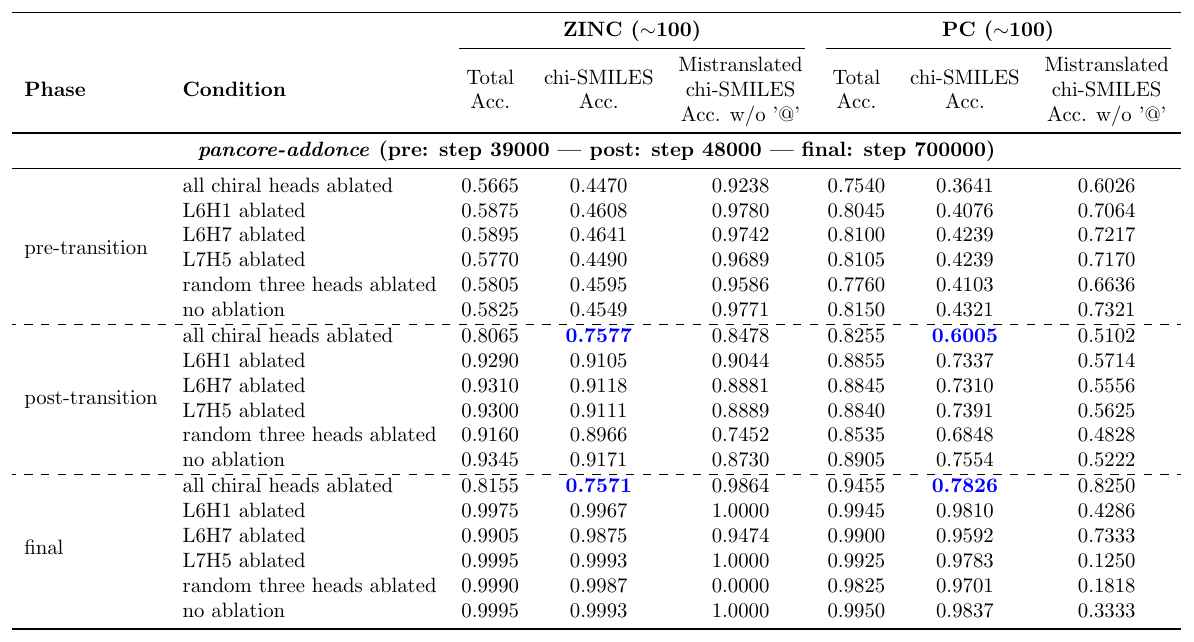}
    \label{headablation}
\end{table}

\newpage

\newpage

\newpage

\newpage

\newpage

\newpage

\newpage

\clearpage
\part*{Supplementary Information}
\setcounter{section}{0}
\renewcommand{\thesection}{\Alph{section}}
\setcounter{figure}{0}
\renewcommand{\thefigure}{S\arabic{figure}}
\setcounter{table}{0}
\renewcommand{\thetable}{S\arabic{table}}

\section{Dynamic Curriculum Learning via Bucket Sampling}
\label{sampling}
To efficiently train the Pan-CORE model on SMILES sequences of varying lengths, we implemented a dynamic curriculum learning strategy using a custom distributed sampler (\texttt{DynamicBucketDistributedSampler}). The training dataset is partitioned into distinct buckets based on sequence length. During training, the probability of sampling a mini-batch from a specific bucket changes dynamically over epochs. This allows the model to initially focus on shorter, simpler sequences before gradually transitioning to longer, more complex ones, thereby stabilizing the early stages of training.

\subsection*{Dynamic Probability Interpolation}
The sampling probabilities for each bucket transition from an initial distribution (e.g., heavily biased towards short sequences) to a final target distribution (e.g., uniform across all buckets) over a defined number of epochs. To achieve a smooth and staggered transition, we utilized a delayed sigmoid interpolation function. For a given bucket $i$ out of $N$ total buckets, a delay offset is defined to ensure that the probability ramp-up for longer sequence buckets occurs sequentially rather than simultaneously:
\begin{equation}
    \text{offset}_i = \text{delay\_factor} \times \frac{i}{N - 1}
\end{equation}
The unnormalized interpolation weight $y(t)$ at a given training progress $t \in [0, 1]$ is calculated using a steepness parameter $k$:
\begin{equation}
    y(t) = \frac{1}{1 + \exp(-k(t - \text{offset}_i))}
\end{equation}
This weight is then min-max normalized using $y(0)$ and $y(1)$ to bound it strictly between 0 and 1, yielding $t_{\text{delayed}, i}$. The dynamic probability for bucket $i$ is then computed as a linear combination of its initial and final probabilities weighted by $t_{\text{delayed}, i}$, followed by normalization across all buckets.

\begin{figure}[H]
    \centering
    \includegraphics[width=0.8\textwidth]{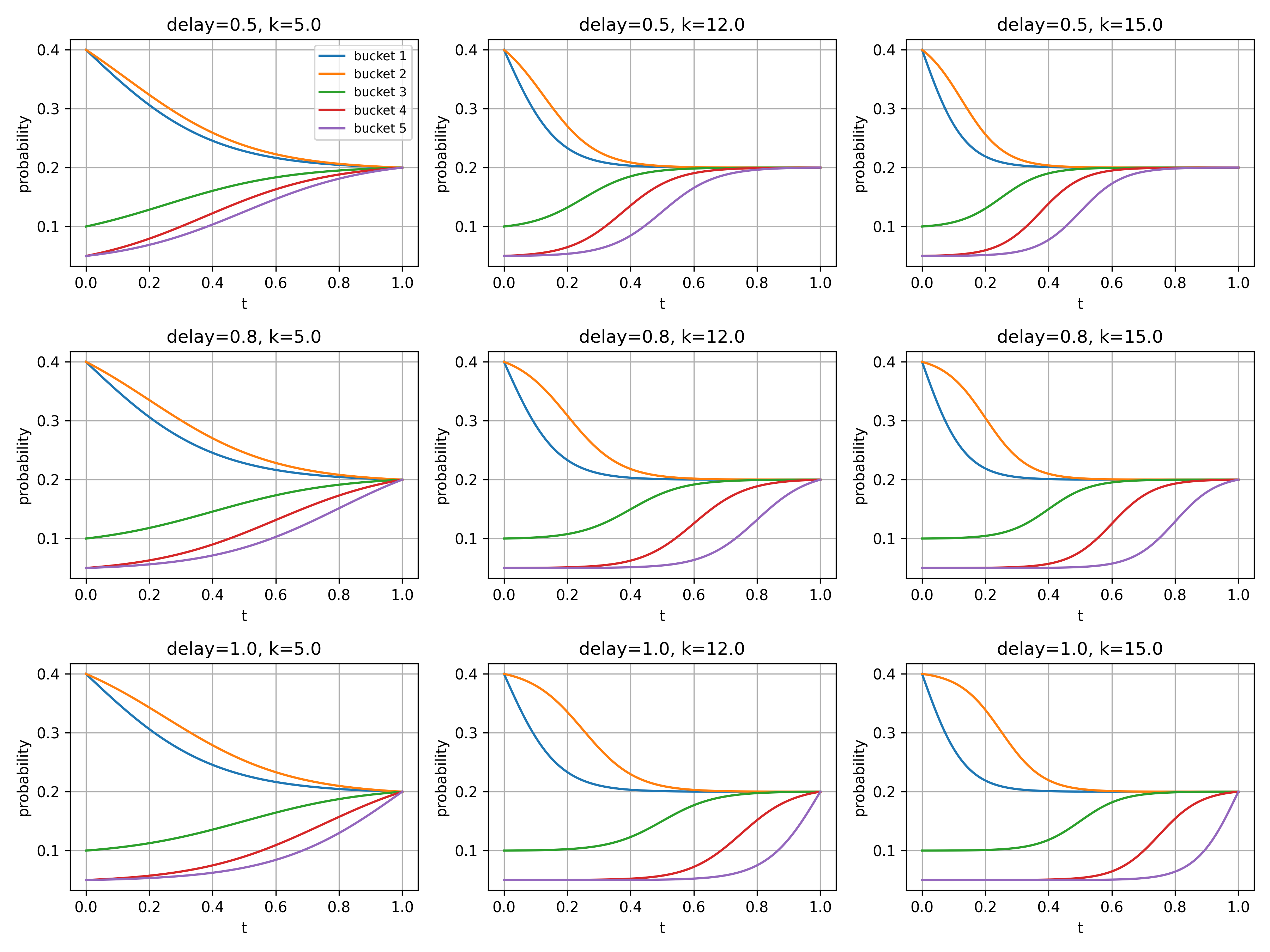} 
    \caption{Transition of sampling probabilities over training progress $t$ under various delayed-sigmoid parameter settings. The 5 lines represent the probabilities of the 5 sequence-length buckets. The center panel ($\text{delay} = 0.8$, $k = 12.0$) highlights the exact hyperparameter configuration adopted for training the Pan-CORE models.}
    \label{samplingprob}
\end{figure}

Figure \ref{samplingprob} illustrates the transition of sampling probabilities for the 5 sequence-length buckets under various combinations of the $\text{delay\_factor}$ and steepness parameter $k$. In this study, we adopted $\text{delay\_factor} = 0.8$ and $k = 12.0$. Under this specific configuration, the sampling probabilities for the buckets containing longer sequences (e.g., Bucket 4 and 5) remain low during the early stages of training and increase sequentially only in the later stages. This strict curriculum effectively prevents the model from being overwhelmed by complex, long-range dependencies before it has acquired the baseline chemical grammar from shorter sequences.

\subsection*{Persistent State and Gradient Accumulation}
To guarantee that all data points are uniformly sampled without being inadvertently skipped due to dynamic probability shifts, the sampler maintains a persistent state across epochs. It tracks the consumed indices for each bucket and only reshuffles a bucket when its specific data pool is fully exhausted. Furthermore, to optimize memory usage and stabilize gradient updates for long sequences, the sampler is designed to yield consecutive micro-batches from the identical bucket during gradient accumulation steps. This ensures that accumulated gradients are derived from sequences of similar lengths, minimizing padding inefficiencies.

\newpage
\section{Details of Attention-related Metrics}
\label{otherattnmetrics}
To analyze the internal dynamics of the Transformer attention mechanism regarding chiral awareness, we computed metrics focusing on the attention weights directed \textit{from} chiral tokens (@, @@) to other tokens in the sequence.

\subsection*{Attention Entropy}
To quantify the dispersion of attention directed from a chiral token to other tokens, we calculated the Shannon entropy of the attention weight distribution. Let $\mathcal{C}$ be the set of (sample, position) pairs $(i,t)$ where the token $t$ is a chiral token. For a given layer $l$ and head $h$, the attention entropy is defined as the average entropy across all chiral tokens:
\begin{equation}
\mathrm{AttnEntropy}^{(l,h)}
  = \frac{1}{|\mathcal{C}|}
    \sum_{(i,t)\in\mathcal{C}}
    \left(
      -\sum_{k=1}^{\ell_i} A^{(l,h)}_{i,t,k} \log\!\left(A^{(l,h)}_{i,t,k} + \epsilon\right)
    \right),
\end{equation}
where $A^{(l,h)}_{i,t,k}$ represents the post-softmax attention weight from the chiral query $t$ to key $k$ at layer $l$, head $h$, sample $i$, and $\ell_i$ is the valid sequence length of sample $i$. $\epsilon$ is a small constant ($1 \times 10^{-10}$) to prevent undefined logarithms. A lower entropy indicates that the attention head is sharply focused on a narrow subset of tokens, whereas a higher entropy indicates broader, more uniform dispersion.

\subsection*{Attention Graph Distance}
To understand whether the model's attention mechanism aligns with the actual 2D topology of the molecule, we computed the attention-weighted average graph distance from chiral tokens to the elements they attend to.

First, the theoretical shortest-path graph distance matrix between all valid atoms in the molecule was generated using RDKit (\texttt{Chem.GetDistanceMatrix}). We then mapped the 1D SMILES token sequence back to these 2D graph nodes. During parsing, structural tokens such as parentheses (indicating branching) were tracked using a stack-based algorithm to maintain the correct atom index. Non-atom tokens (e.g., bond symbols, brackets, and stereochemical marks) inherited the graph index of the most recently parsed relevant atom in their immediate local structure.

Let $D_{i,t,k}$ be the shortest path distance in the molecular graph between the atom corresponding to the chiral token $t$ and the atom corresponding to token $k$ in sample $i$. The Attention Graph Distance for layer $l$ and head $h$ is calculated as the weighted sum:
\begin{equation}
\mathrm{GraphDist}^{(l,h)}
  = \frac{1}{|\mathcal{C}|}
    \sum_{(i,t)\in\mathcal{C}}
    \sum_{k=1}^{\ell_i} A^{(l,h)}_{i,t,k} D_{i,t,k},
\end{equation}
This metric reveals whether specific attention heads have learned to focus on topologically adjacent substructures (resulting in a low average graph distance) despite those components potentially being separated by long string distances in the 1D SMILES sequence.

\newpage
\begin{table}[H]
    \centering
    \caption{Number of compounds in each sequence length bucket before randomized augmentation.}
    \includegraphics[width=0.5\textwidth]{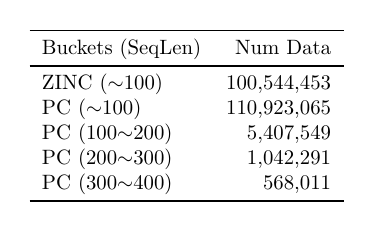}
    \label{numdataraw}
\end{table}
\newpage
\begin{table}[H]
    \centering
    \caption{Number of Random–Canonical SMILES pairs in each sequence length bucket after randomized augmentation.}
    \includegraphics[width=0.5\textwidth]{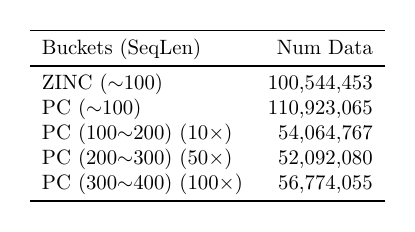}
    \label{numdataaug}
\end{table}

\newpage
\begin{table}[H]
    \centering
    \caption{Total number of trainable parameters for each model variant.}
    \label{modelsize}
    \includegraphics[width=0.5\textwidth]{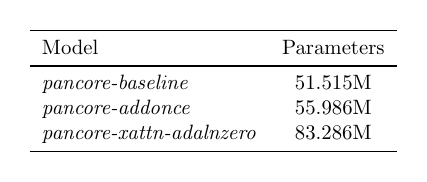}
\end{table}

\newpage
\begin{table}[H]
    \centering
    \caption{Cross-evaluation of translation accuracy using pre-transition and post-transition encoder–decoder configurations across all sequence length buckets and all three model variants.}
    \includegraphics[width=0.9\textwidth]{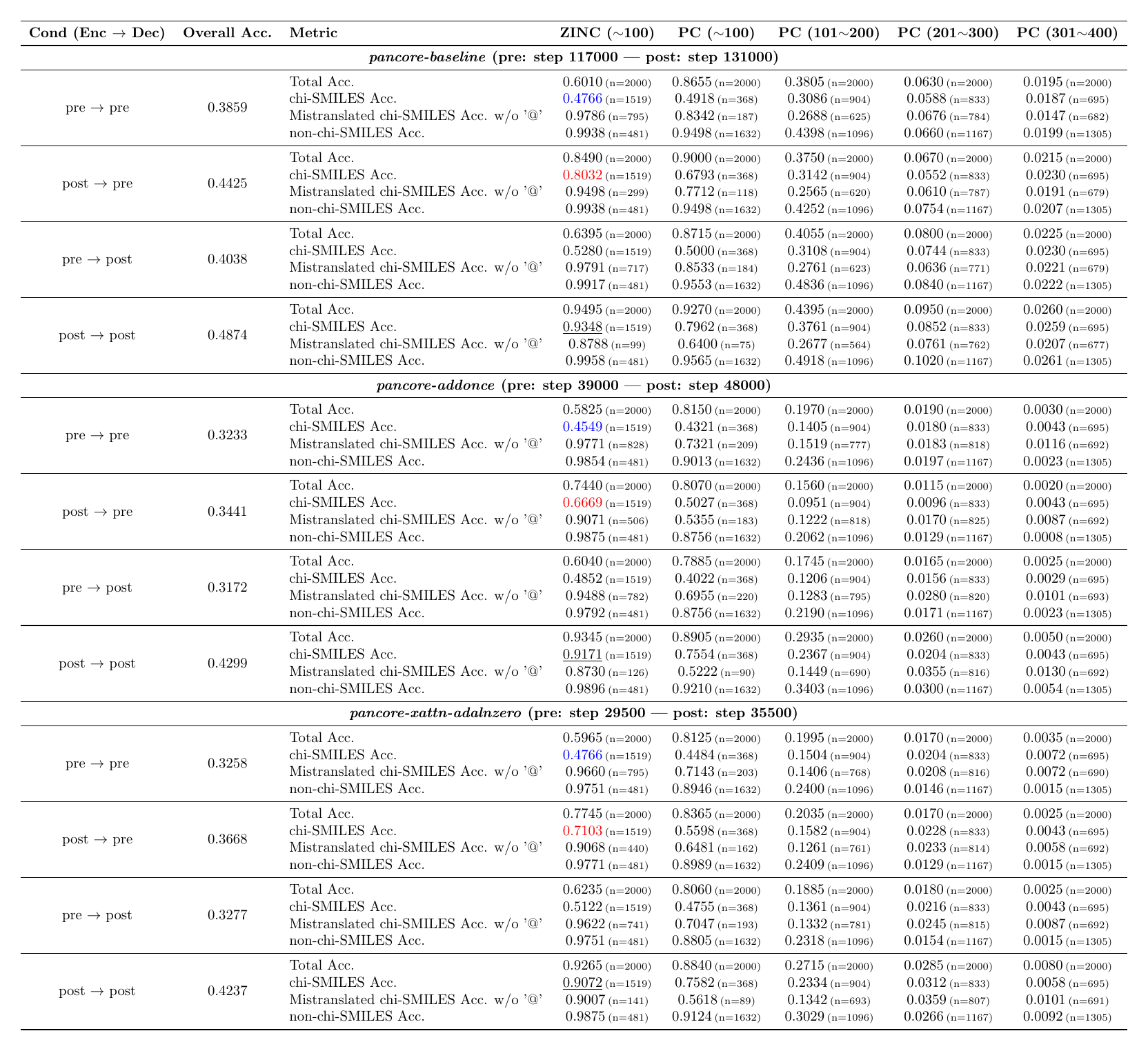}
    \label{encdeccrossaccuracyfull}
\end{table}

\begin{sidewaystable}
    \centering
    \caption{Robustness evaluation of chiral recognition via attention head ablation across pre-transition, post-transition, and final training stages for all three model variants and all sequence length buckets.}
    \includegraphics[width=0.8\textheight]{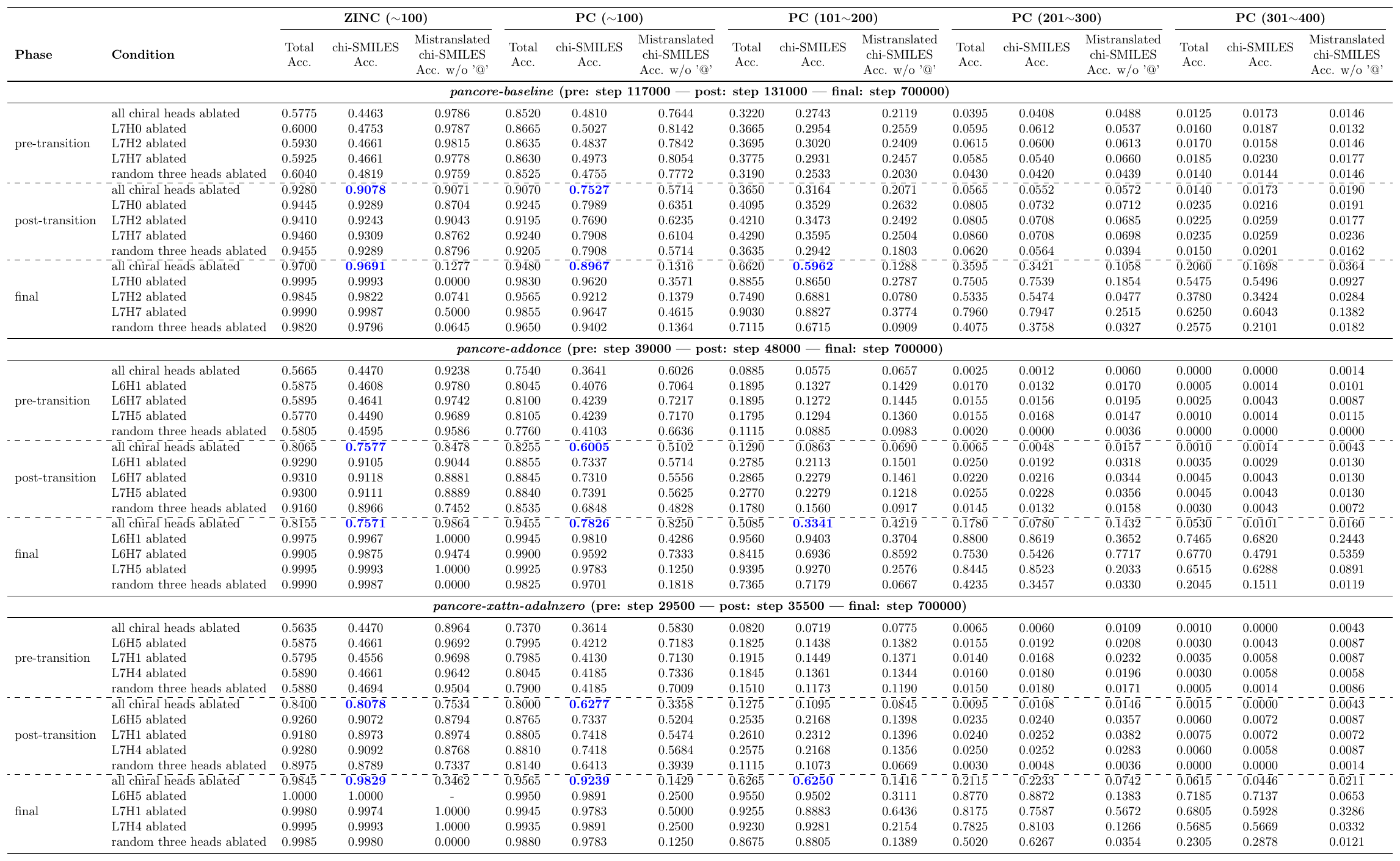}
    \label{headablationfull}
\end{sidewaystable}

\newpage

\newpage
\begin{figure}[H]
    \centering
    \begin{subfigure}{0.48\textwidth}
        \centering
        \includegraphics[width=\linewidth]{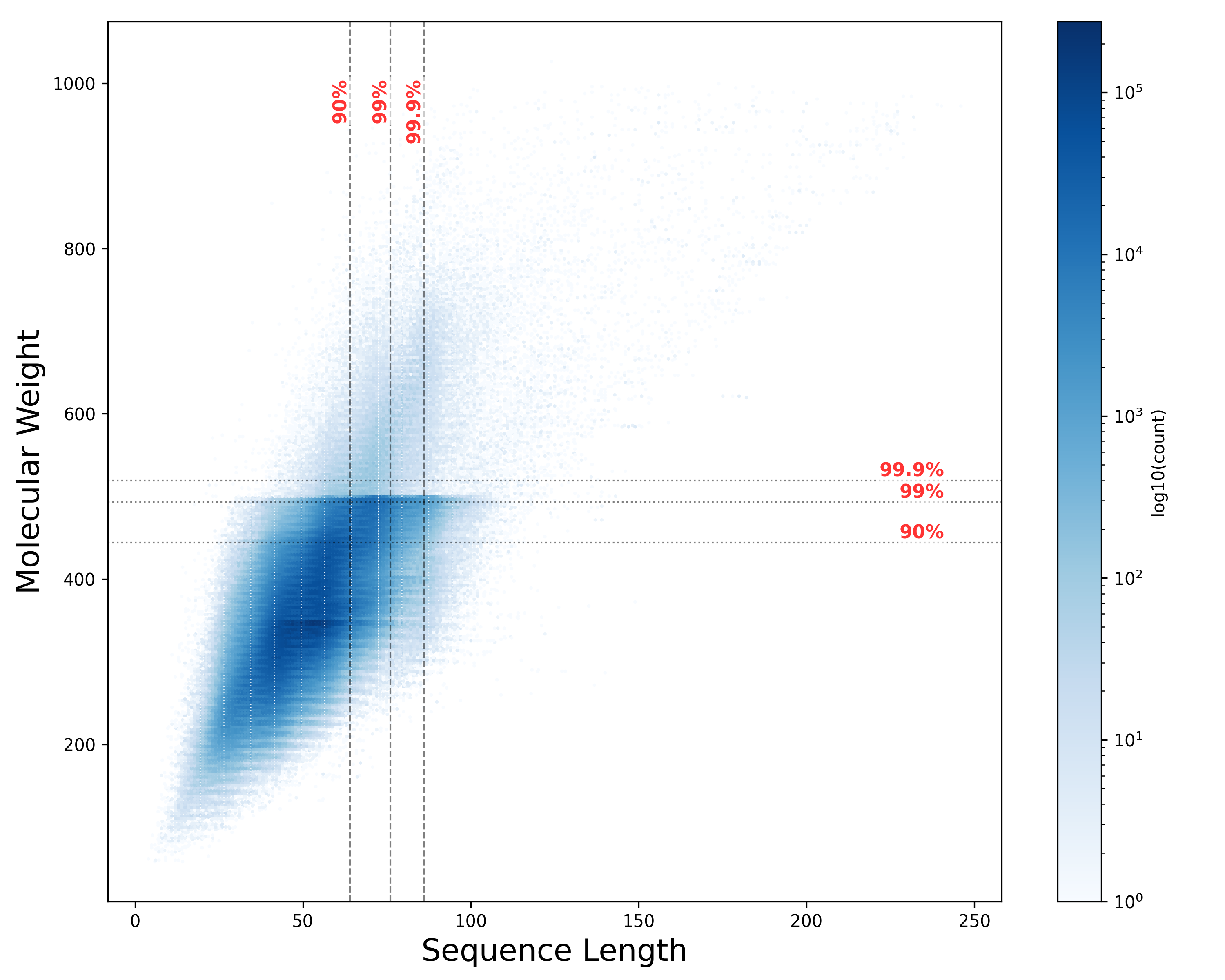}
        \caption{ZINC20}
        \label{zinc20mwseqlen}
    \end{subfigure}
    \hfill
    \begin{subfigure}{0.48\textwidth}
        \centering
        \includegraphics[width=\linewidth]{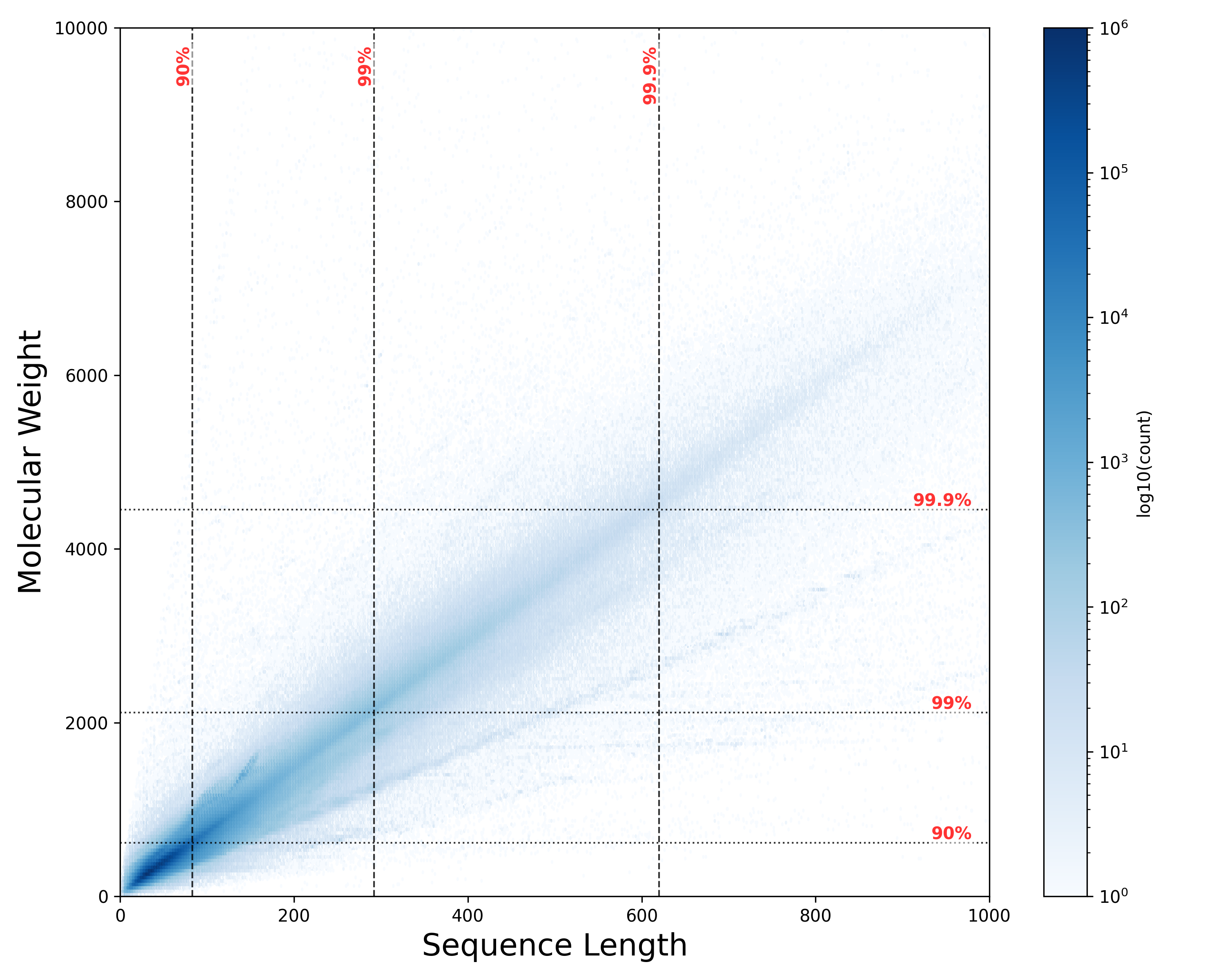}
        \caption{PubChem}
        \label{pubchemmwseqlen}
    \end{subfigure}
    \caption{Correlation between tokenized SMILES sequence length and molecular weight for (\subref{zinc20mwseqlen}) ZINC20 and (\subref{pubchemmwseqlen}) PubChem datasets, visualized as 2D histograms with logarithmic density scaling. Red dashed vertical lines indicate the sequence length cutoffs adopted for filtering, with annotated values representing the cumulative percentage of compounds retained at each threshold. Red dashed horizontal lines denote representative molecular weight boundaries, confirming that the applied sequence length criteria adequately encompass the target molecular weight range.}
    \label{mwseqlen}
\end{figure}

\newpage
\begin{figure}[H]
    \centering
    \begin{subfigure}{0.48\textwidth}
        \centering
        \includegraphics[width=\linewidth]{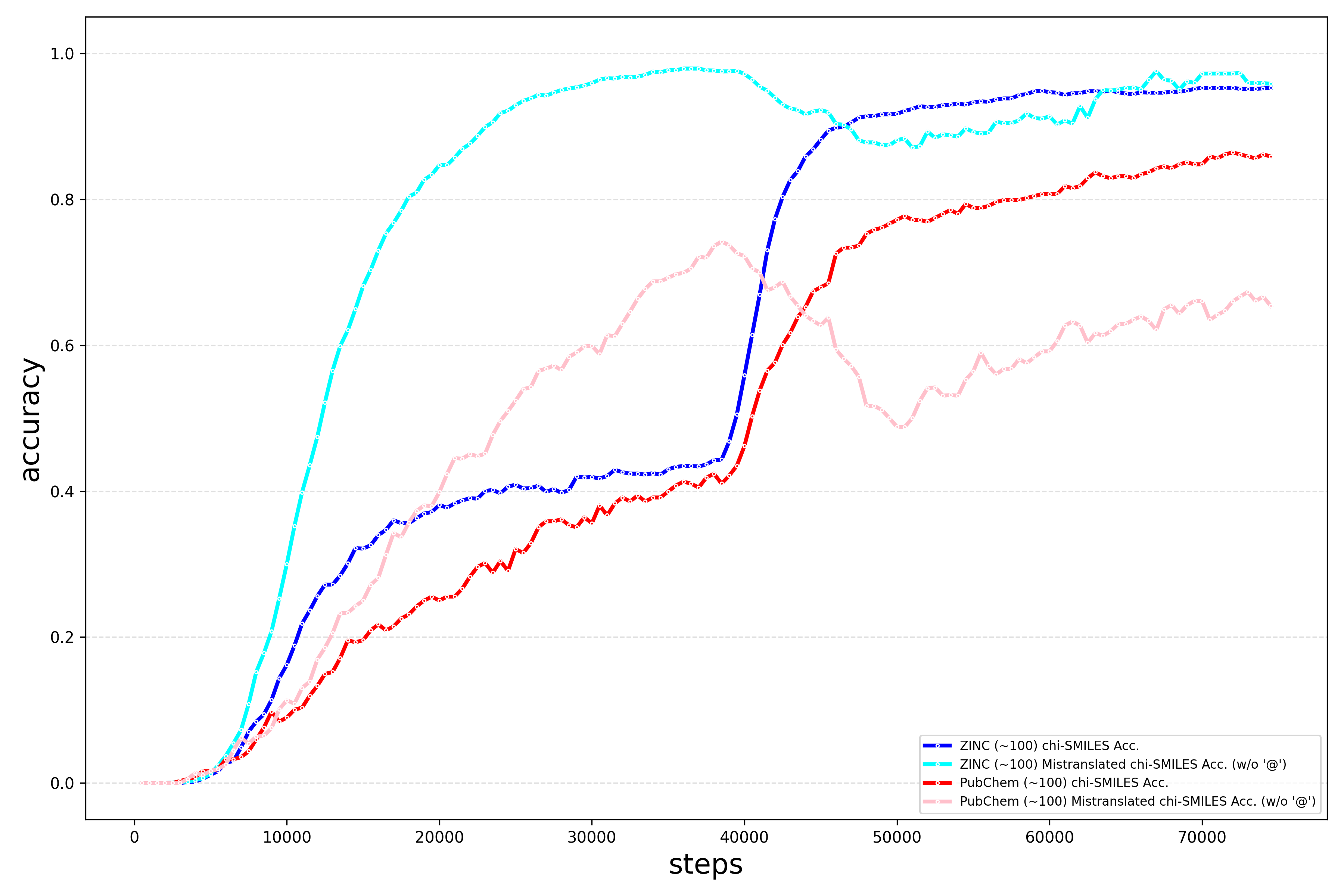}
        \caption{$\Delta$500 steps}
        \label{chiralaccdelta500}
    \end{subfigure}
    \hfill
    \begin{subfigure}{0.48\textwidth}
        \centering
        \includegraphics[width=\linewidth]{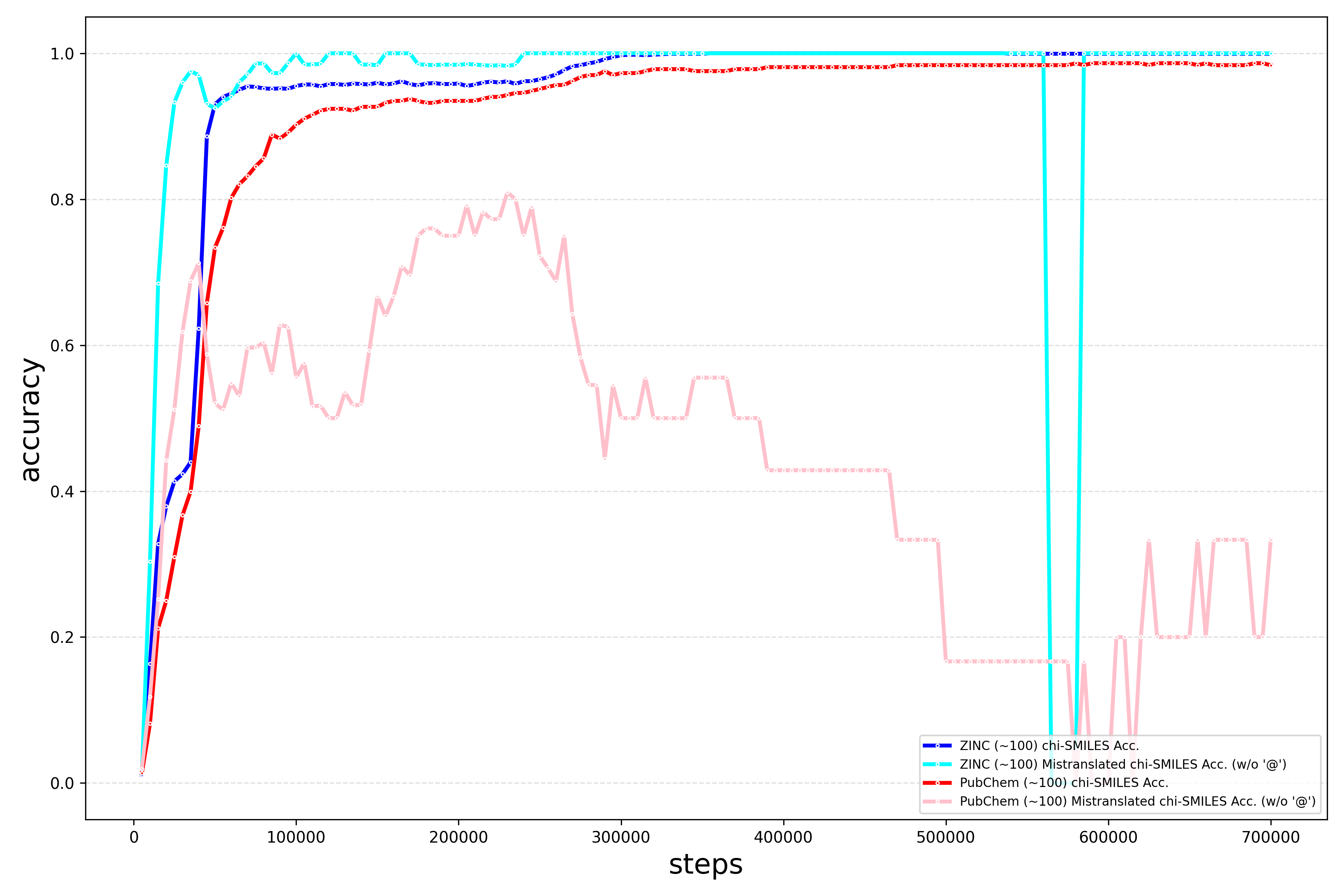}
        \caption{$\Delta$5000 steps}
        \label{chiralaccdelta5000}
    \end{subfigure}
    \caption{SMILES with chiral token (chi-SMILES) accuracy and mistranslated chi-SMILES accuracy without chiral marks for ZINC20 (\til 100) and PubChem (\til 100) buckets, evaluated using checkpoint intervals of (\subref{chiralaccdelta500}) $\Delta$ 500 steps and (\subref{chiralaccdelta5000}) $\Delta$ 5000 steps for \textit{pancore-addonce} Filtering strictly for chi-SMILES in the PubChem (\til 100) bucket reveals a similar abrupt transition to that observed in ZINC20, suggesting that the jump-up is partly obscured in overall PubChem accuracy by the lower frequency of chiral compounds as suggested by number of tokens in Fig.~\ref{tokenacc}.}
    \label{chiralacc}
\end{figure}

\newpage
\begin{figure}[H]
    \centering
    \begin{subfigure}{0.48\textwidth}
        \centering
        \includegraphics[width=\linewidth]{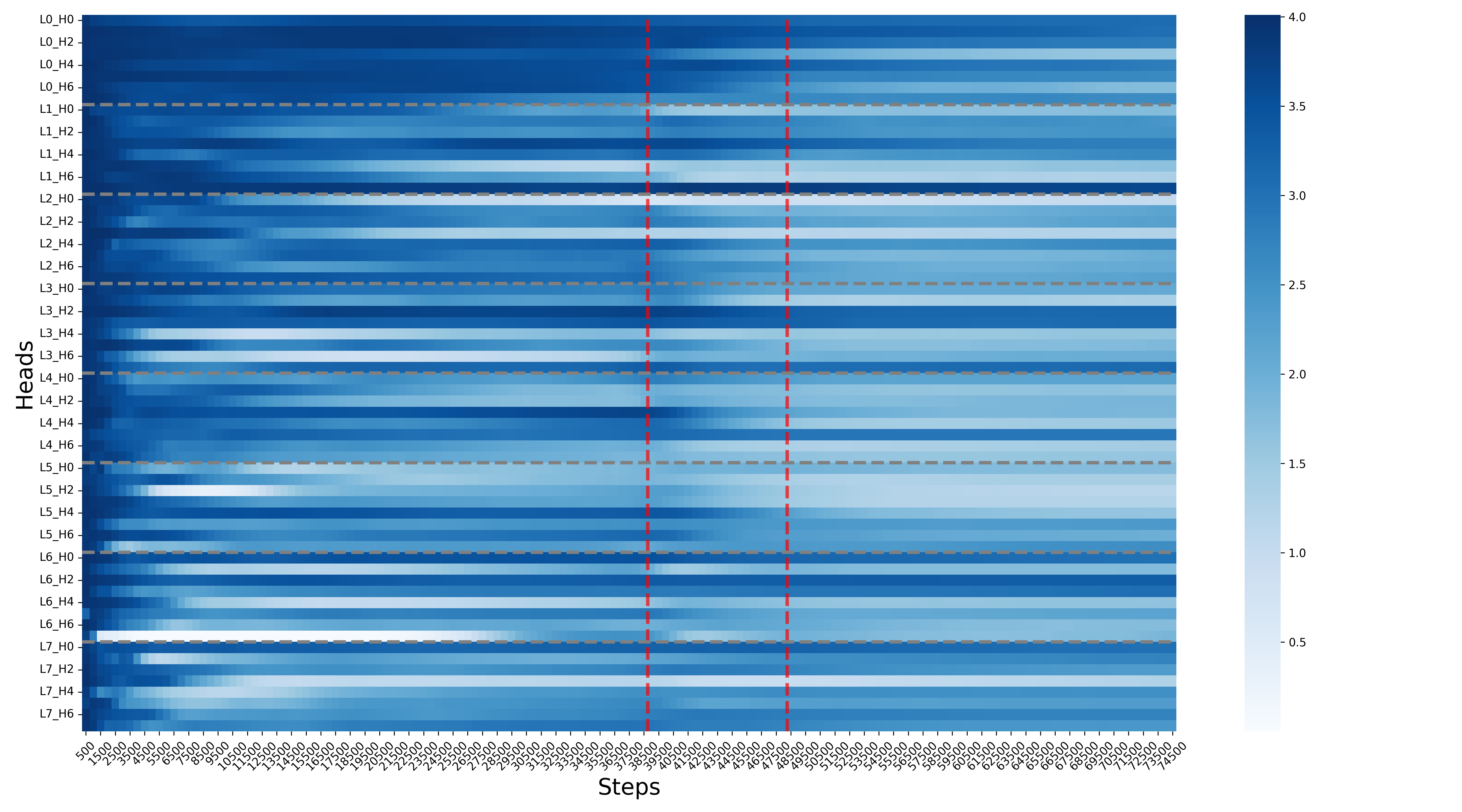}
        \caption{Encoder - all heads}
        \label{attnentropyencoderheatmap}
    \end{subfigure}
    \hfill
    \begin{subfigure}{0.48\textwidth}
        \centering
        \includegraphics[width=\linewidth]{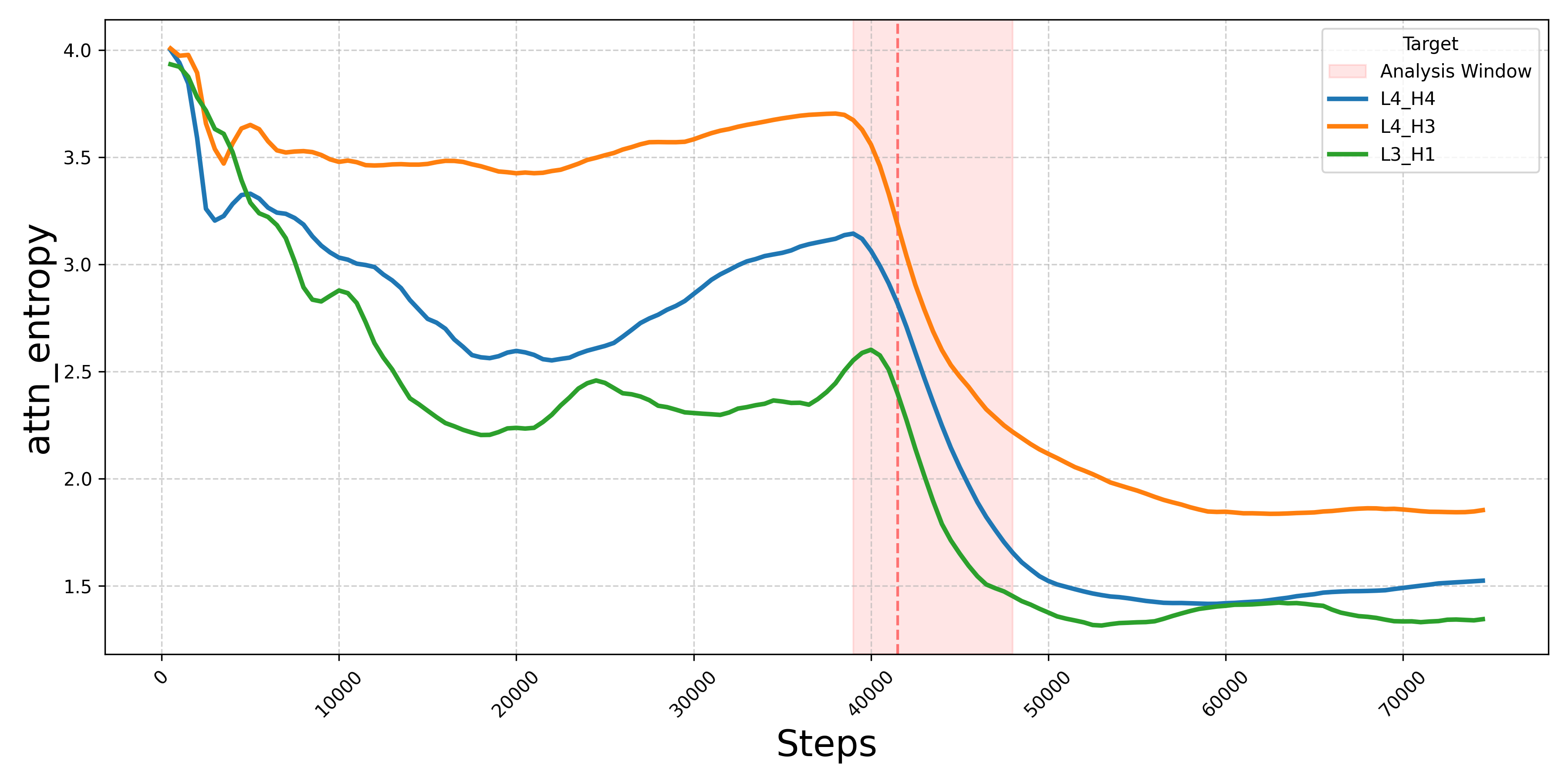}
        \caption{Encoder - Top3 heads}
        \label{attnentropyencodertop3}
    \end{subfigure}

    \vspace{1em}

    \begin{subfigure}{0.48\textwidth}
        \centering
        \includegraphics[width=\linewidth]{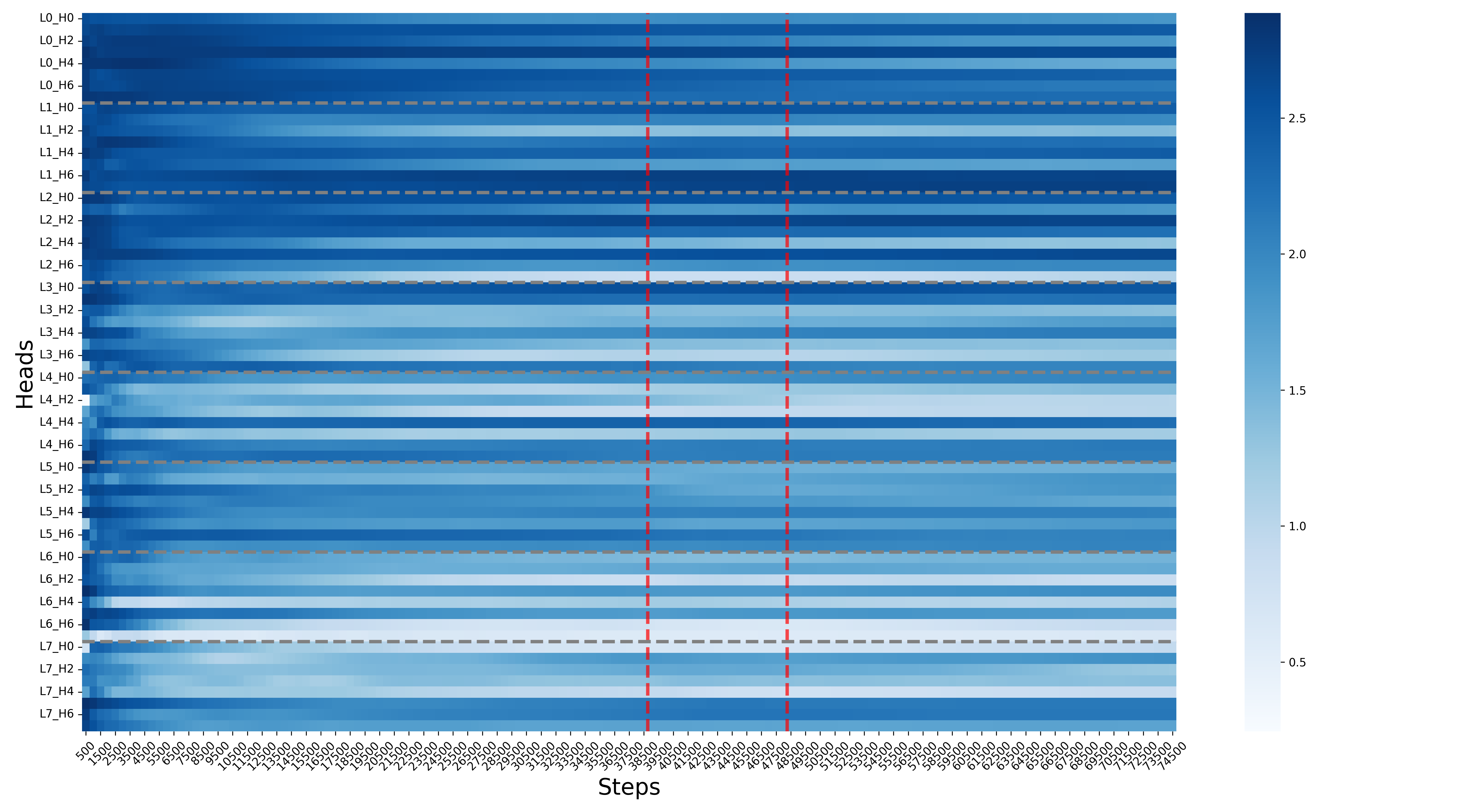}
        \caption{Decoder - all heads}
        \label{attnentropydecoderheatmap}
    \end{subfigure}
    \hfill
    \begin{subfigure}{0.48\textwidth}
        \centering
        \includegraphics[width=\linewidth]{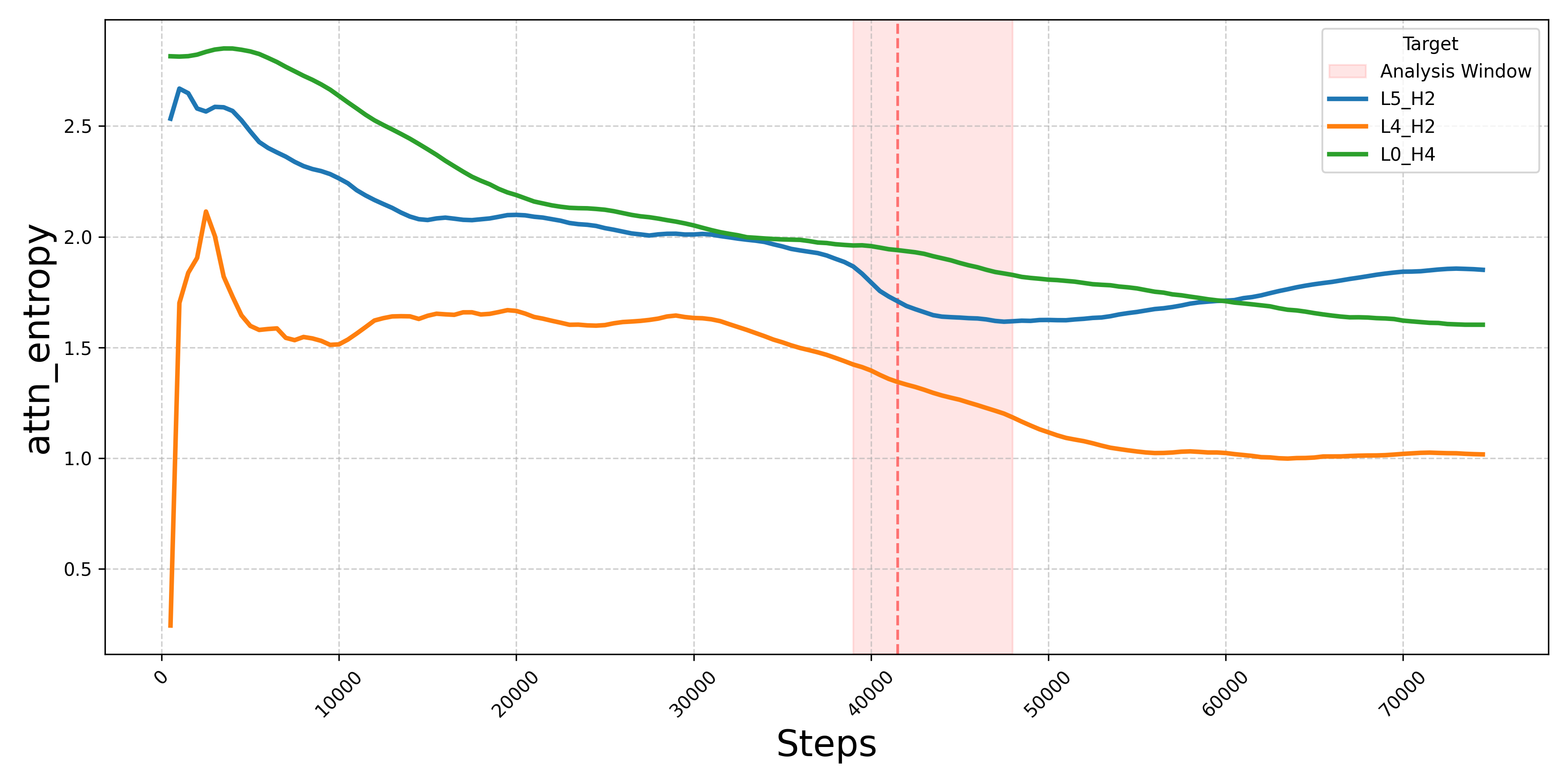}
        \caption{Decoder - Top3 heads}
        \label{attnentropydecodertop3}
    \end{subfigure}
    
    \caption{Trajectories of Attention Entropy directed from chiral tokens across all heads from L0H0 (layer 0, head 0) to L7H7, for \textit{pancore-addonce} on ZINC20 (\til 100)). (\subref{attnentropyencoderheatmap}, \subref{attnentropyencodertop3}) Encoder; (\subref{attnentropydecoderheatmap}, \subref{attnentropydecodertop3}) Decoder. (\subref{attnentropyencoderheatmap}, \subref{attnentropydecoderheatmap}) The region bounded by the two vertical dashed lines indicates the jump-up interval defined by perplexity. (\subref{attnentropyencodertop3}, \subref{attnentropydecodertop3}) Changes in the three heads with the largest entropy decrease within the jump-up interval. The red line and the light red shaded region indicate the perplexity-derived jump-up interval.}
    \label{attnentropy}
\end{figure}

\newpage
\begin{figure}[H]
    \centering
    \begin{subfigure}{0.48\textwidth}
        \centering
        \includegraphics[width=\linewidth]{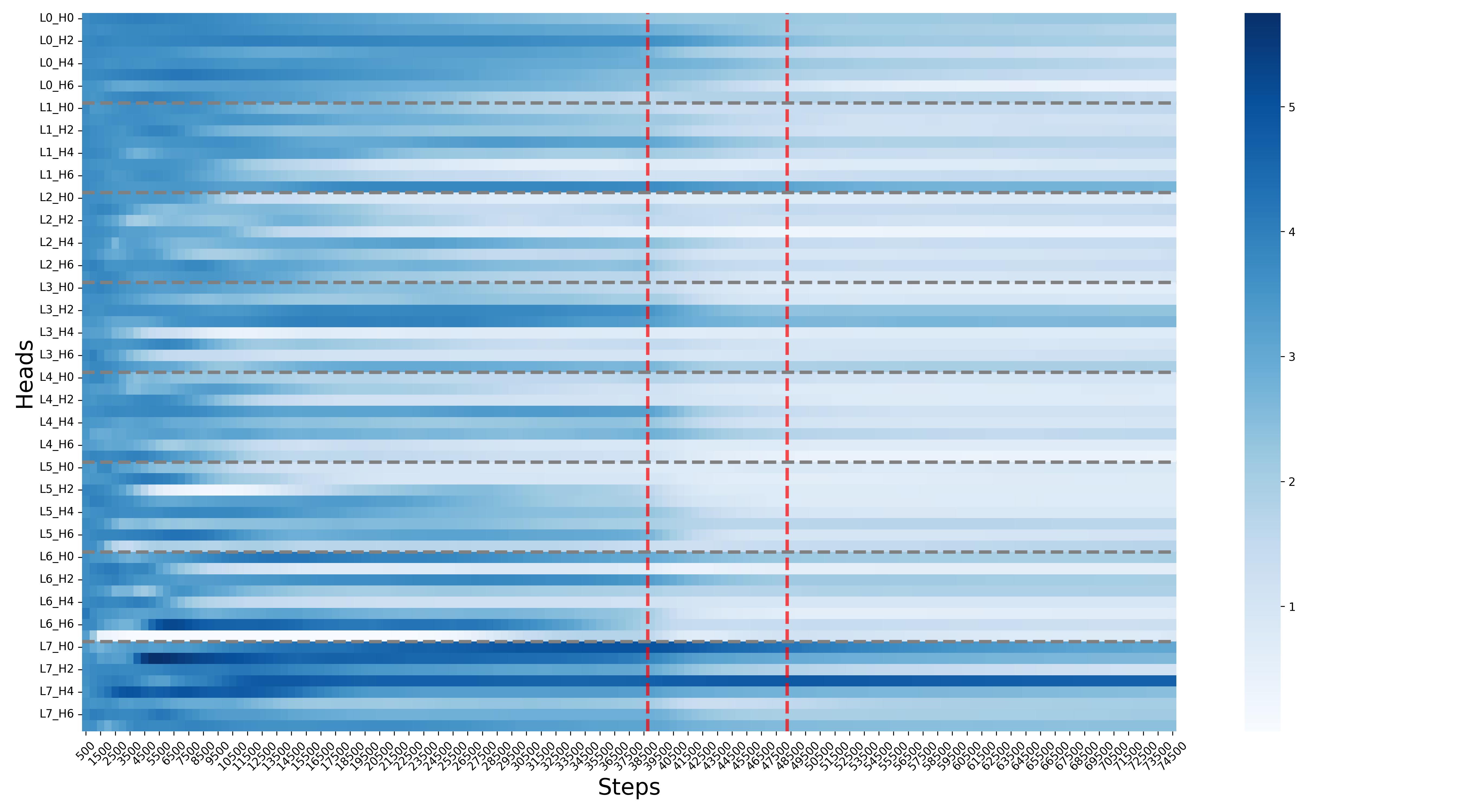}
        \caption{Encoder - all heads}
        \label{graphdistencoderheatmap}
    \end{subfigure}
    \hfill
    \begin{subfigure}{0.48\textwidth}
        \centering
        \includegraphics[width=\linewidth]{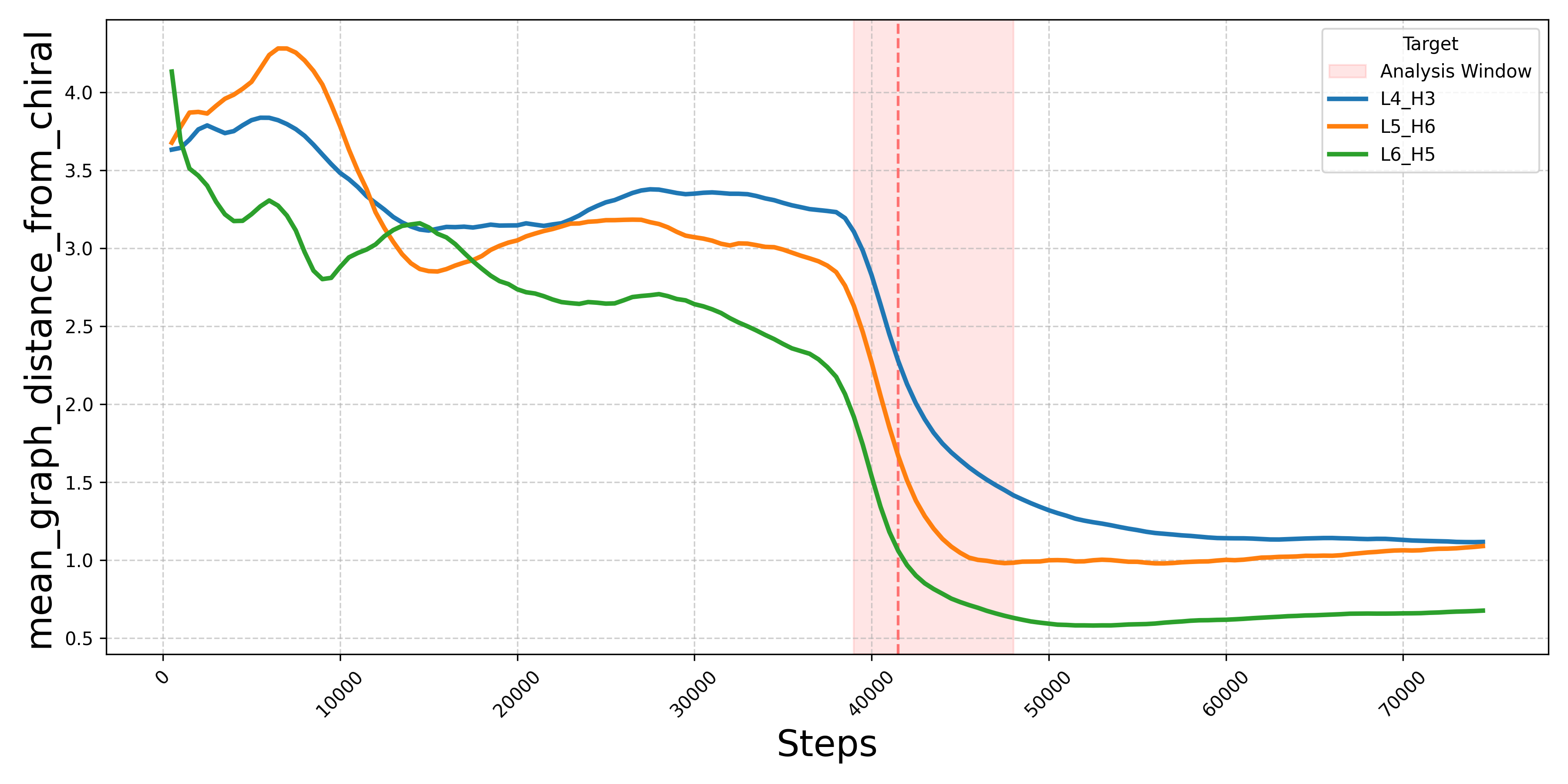}
        \caption{Encoder - Top3 heads}
        \label{graphdistencodertop3}
    \end{subfigure}

    \vspace{1em}

    \begin{subfigure}{0.48\textwidth}
        \centering
        \includegraphics[width=\linewidth]{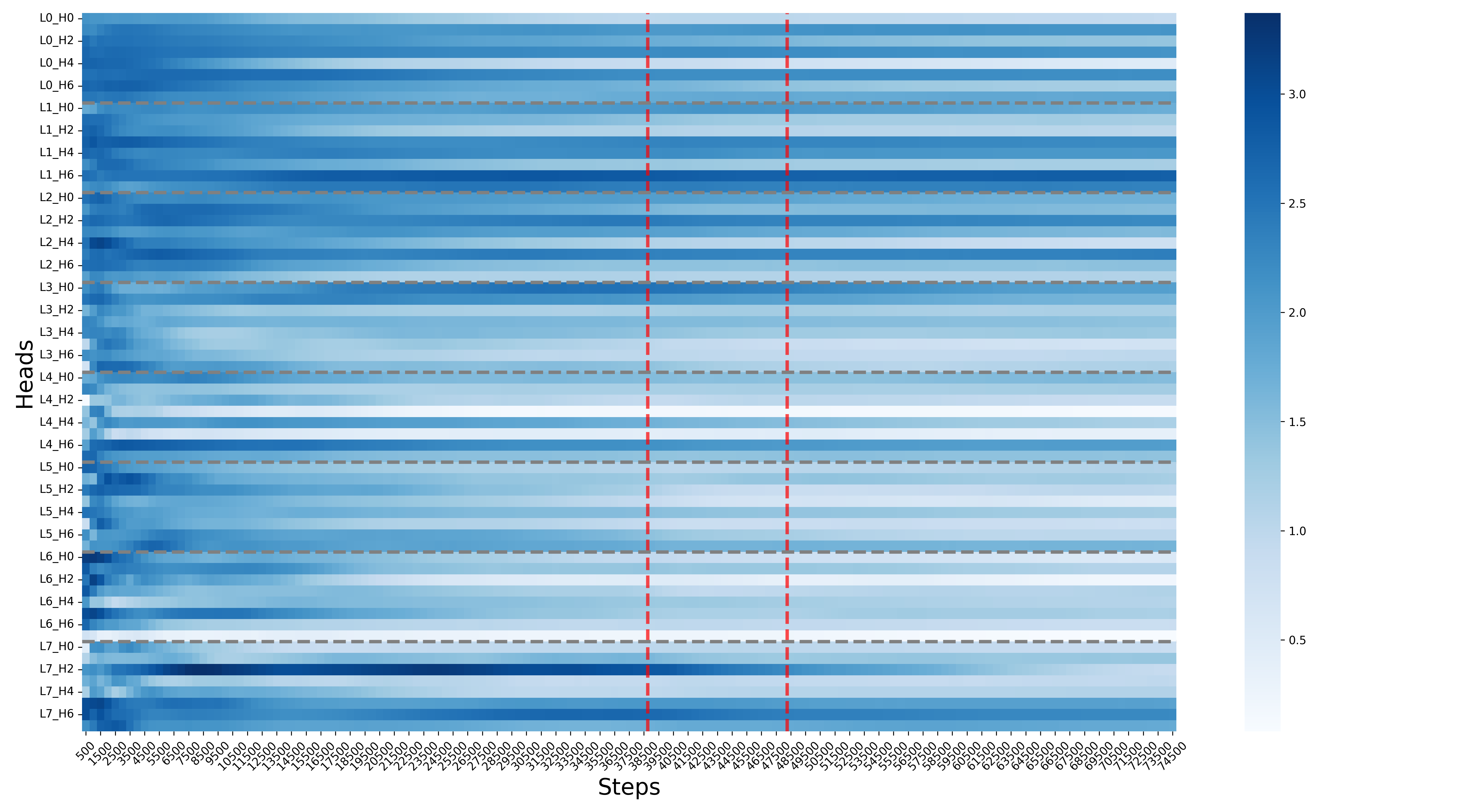}
        \caption{Decoder - all heads}
        \label{graphdistdecoderheatmap}
    \end{subfigure}
    \hfill
    \begin{subfigure}{0.48\textwidth}
        \centering
        \includegraphics[width=\linewidth]{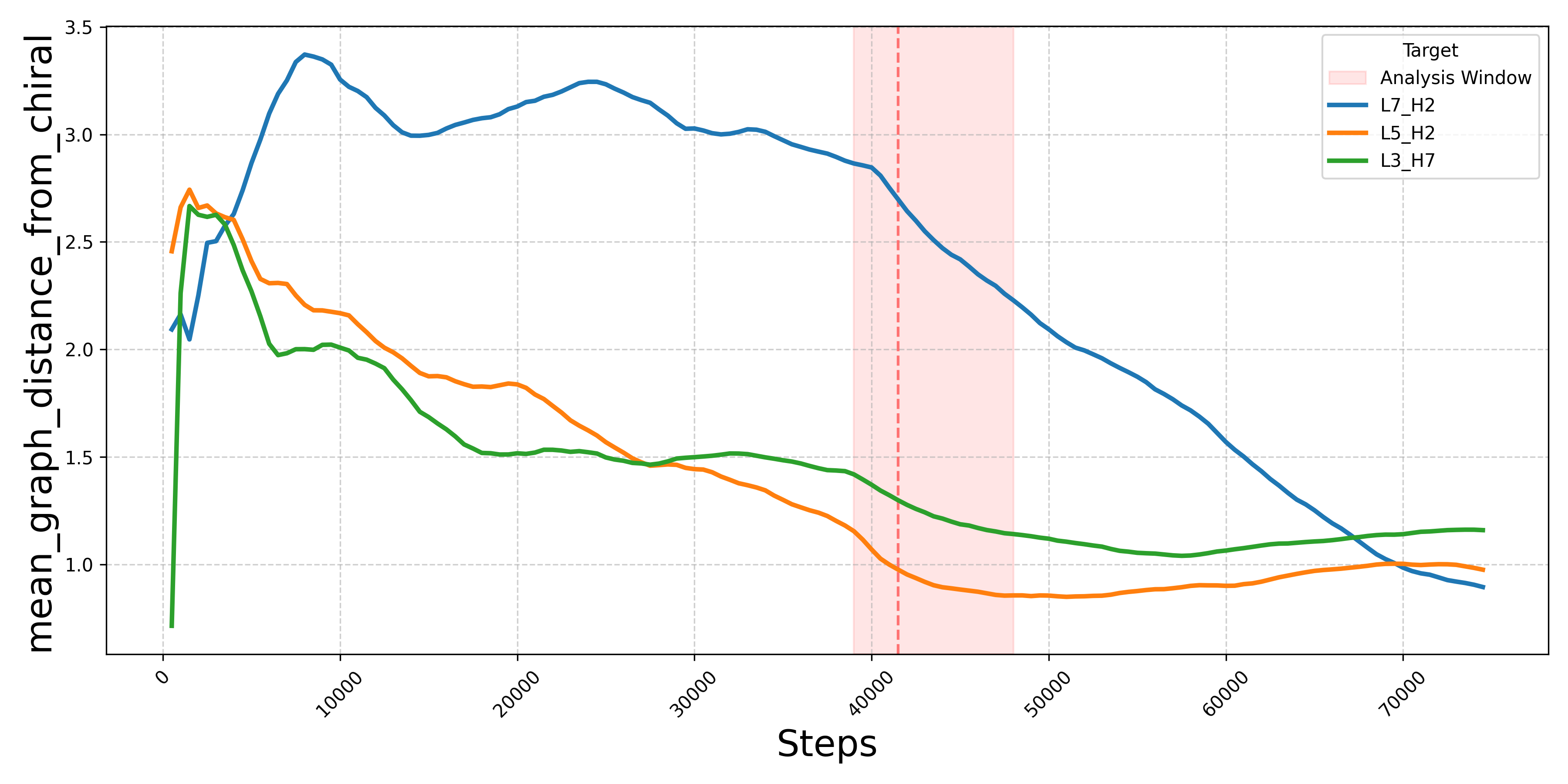}
        \caption{Decoder - Top3 heads}
        \label{graphdistdecodertop3}
    \end{subfigure}
    
    \caption{Trajectories of the attention-weighted average graph distance from chiral tokens to attended positions across all heads from L0H0 (layer 0, head 0) to L7H7, for \textit{pancore-addonce} on ZINC20 (\til 100). (\subref{graphdistencoderheatmap}, \subref{graphdistencodertop3}) Encoder; (\subref{graphdistdecoderheatmap}, \subref{graphdistdecodertop3}) Decoder. (\subref{graphdistencoderheatmap}, \subref{graphdistdecoderheatmap}) The region bounded by the two vertical dashed lines indicates the jump-up interval defined by perplexity. (\subref{graphdistencodertop3}, \subref{graphdistdecodertop3}) Changes in the three heads with the largest variation in graph distance within the jump-up interval. The red line and the light red shaded region indicate the perplexity-derived jump-up interval.}
    \label{graphdist}
\end{figure}

\newpage
\begin{figure}[H]
    \centering
    \begin{subfigure}{0.48\textwidth}
        \centering
        \includegraphics[width=\linewidth]{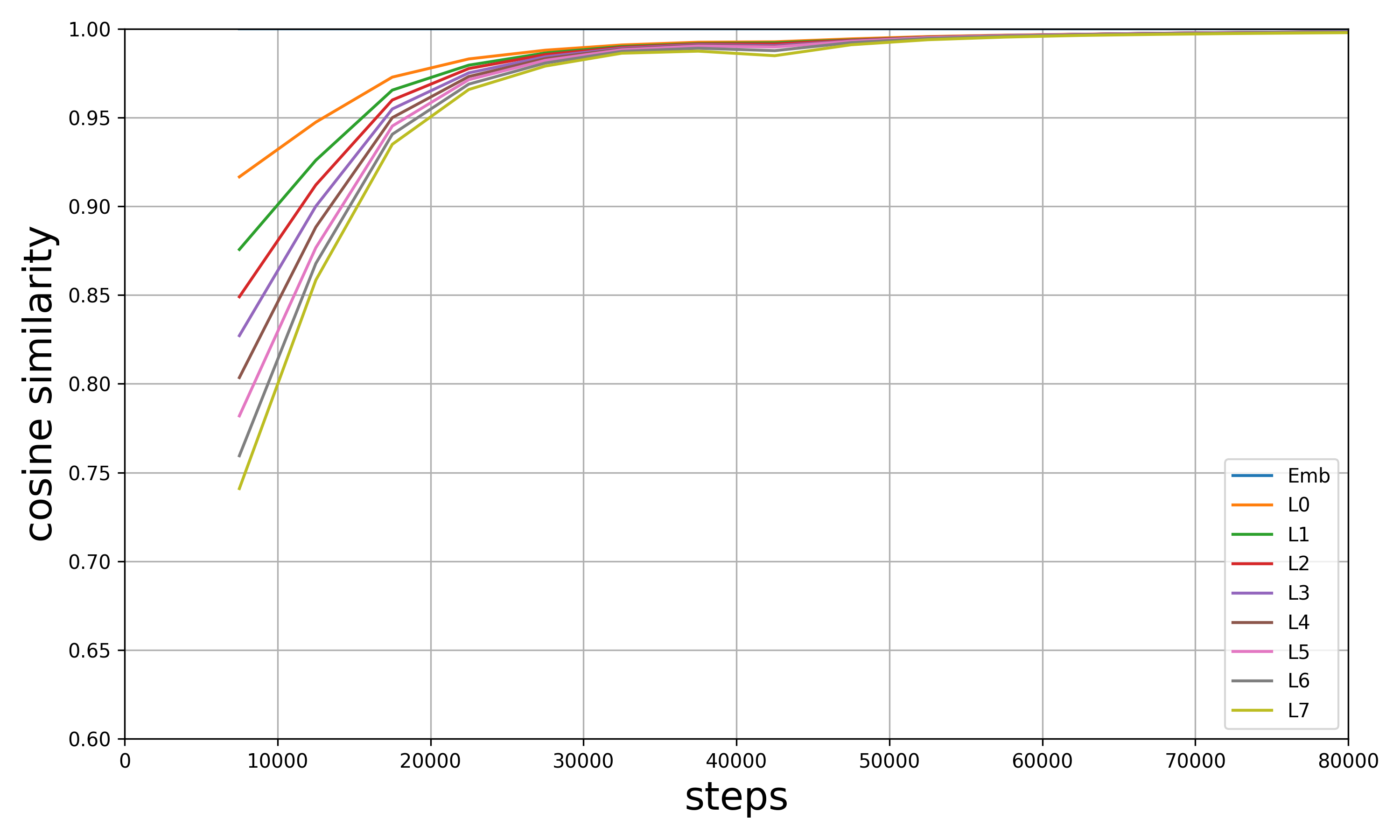}
        \caption{Background - cosine similarity}
        \label{backgroundcosdelta5000}
    \end{subfigure}
    \hfill
    \begin{subfigure}{0.48\textwidth}
        \centering
        \includegraphics[width=\linewidth]{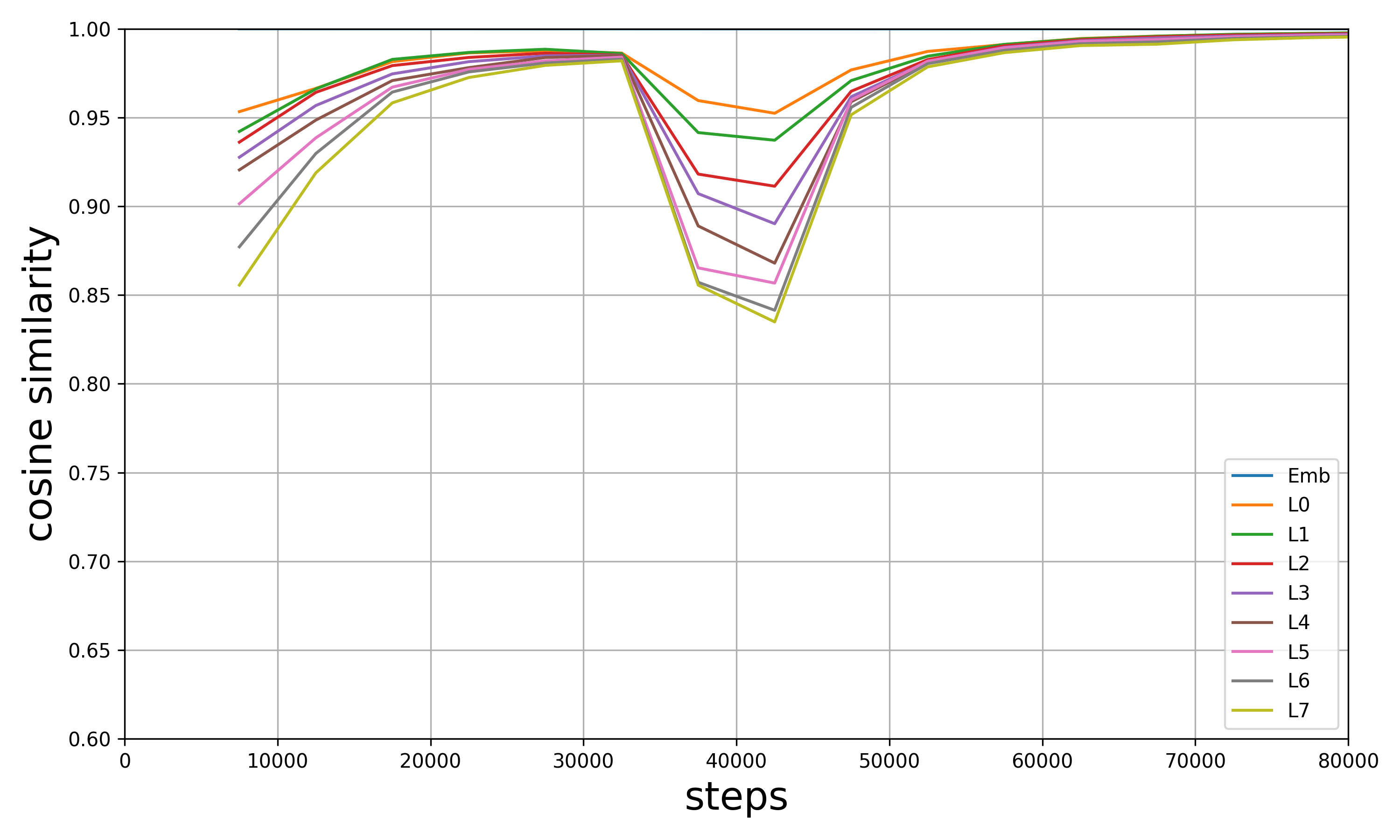}
        \caption{Chiral - cosine similarity}
        \label{chiralcosdelta5000}
    \end{subfigure}
    \caption{Cosine similarity of encoder residual stream vectors between consecutive checkpoints evaluated at a coarser 5,000-step interval ($\Delta5000$steps) for \textit{pancore-addonce}, ZINC20 (\til 100). (\subref{backgroundcosdelta5000}) Background tokens (excluding chiral tokens); (\subref{chiralcosdelta5000}) Chiral tokens (@ and @@). In contrast to the 500-step resolution shown in Fig.~\ref{residualmetrics}~\subref{backgroundcos},~\subref{chiralcos}, the 5,000-step resolution reveals a pronounced V-shaped drop in the directional stability of chiral token representations in absolute terms.}
    \label{residualcosdelta5000}
\end{figure}

\newpage
\begin{figure}[H]
    \centering
    \begin{subfigure}{0.48\textwidth}
        \centering
        \includegraphics[width=\linewidth]{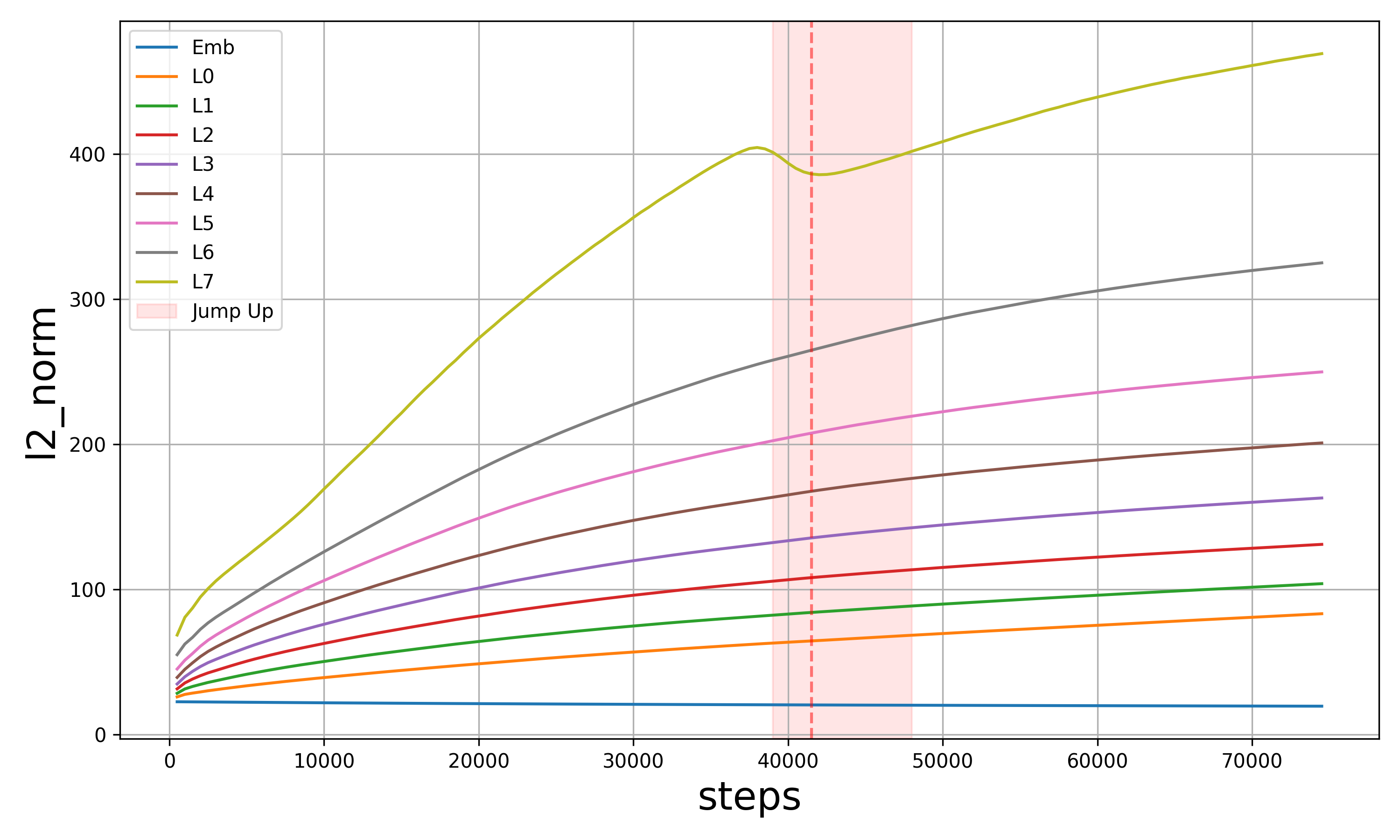}
        \caption{Background - L2 Norm}
        \label{backgroundl2normdecoder}
    \end{subfigure}
    \hfill
    \begin{subfigure}{0.48\textwidth}
        \centering
        \includegraphics[width=\linewidth]{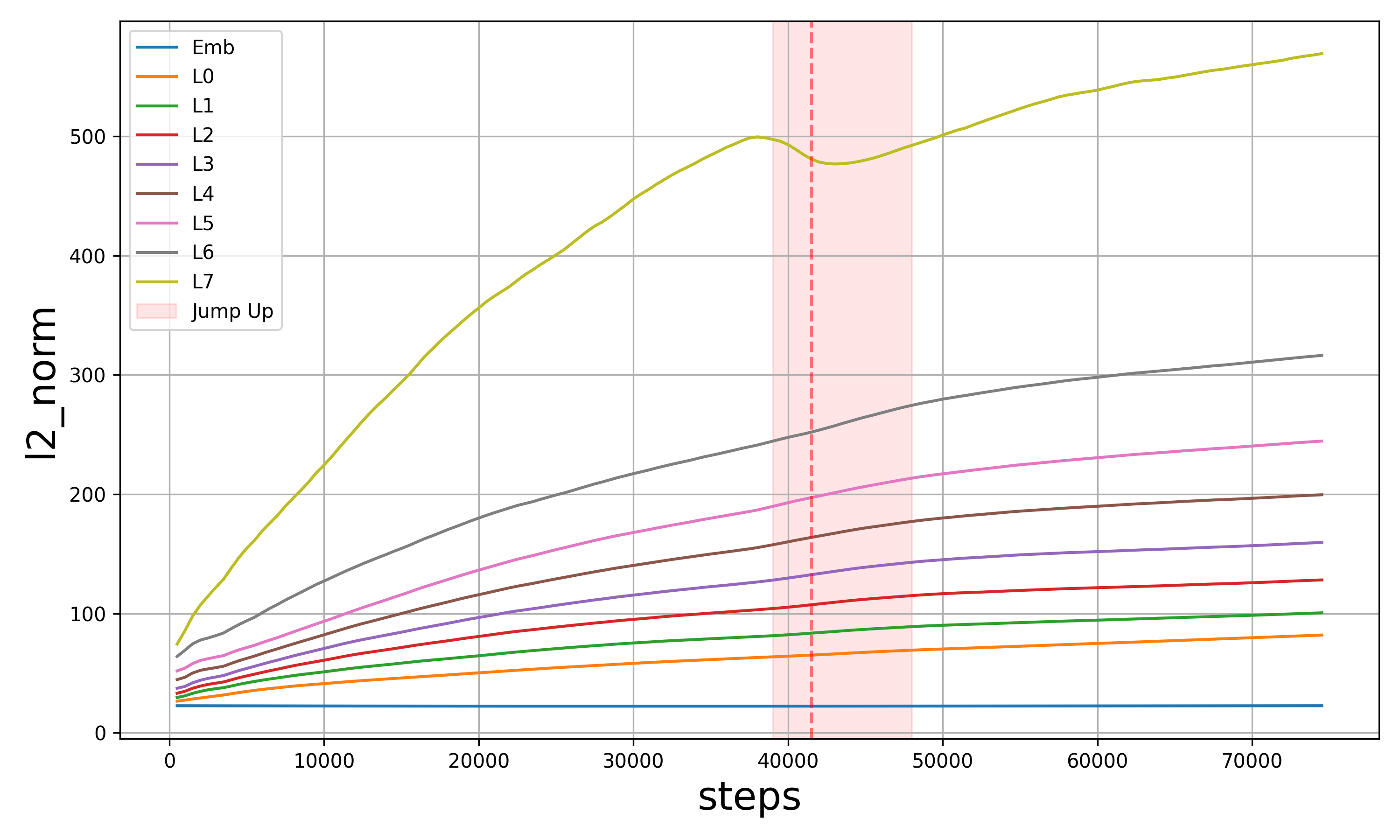}
        \caption{Chiral - L2 Norm}
        \label{chirall2normdecoder}
    \end{subfigure}

    \vspace{1em}

    \begin{subfigure}{0.48\textwidth}
        \centering
        \includegraphics[width=\linewidth]{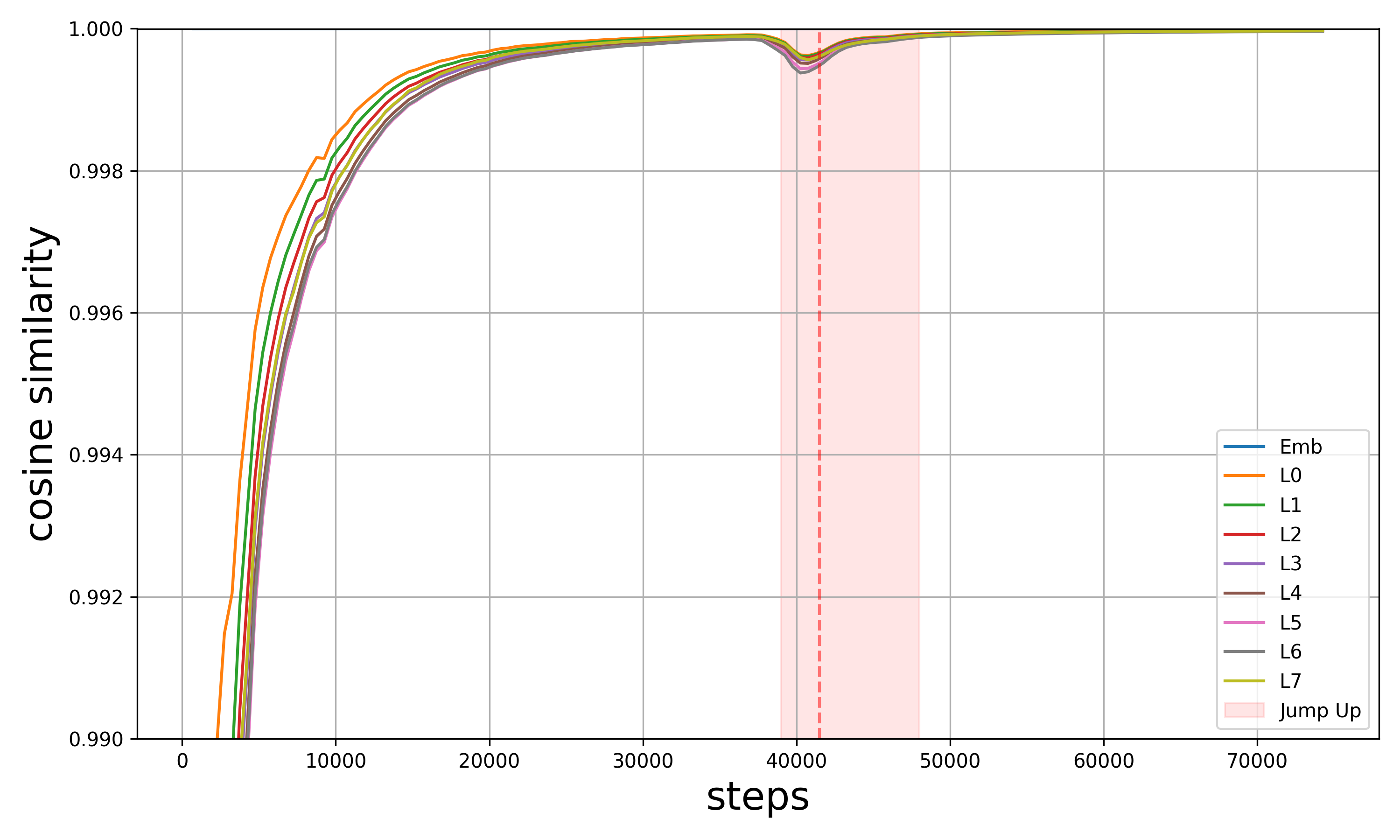}
        \caption{Background - cosine similarity}
        \label{backgroundcosdecoder}
    \end{subfigure}
    \hfill
    \begin{subfigure}{0.48\textwidth}
        \centering
        \includegraphics[width=\linewidth]{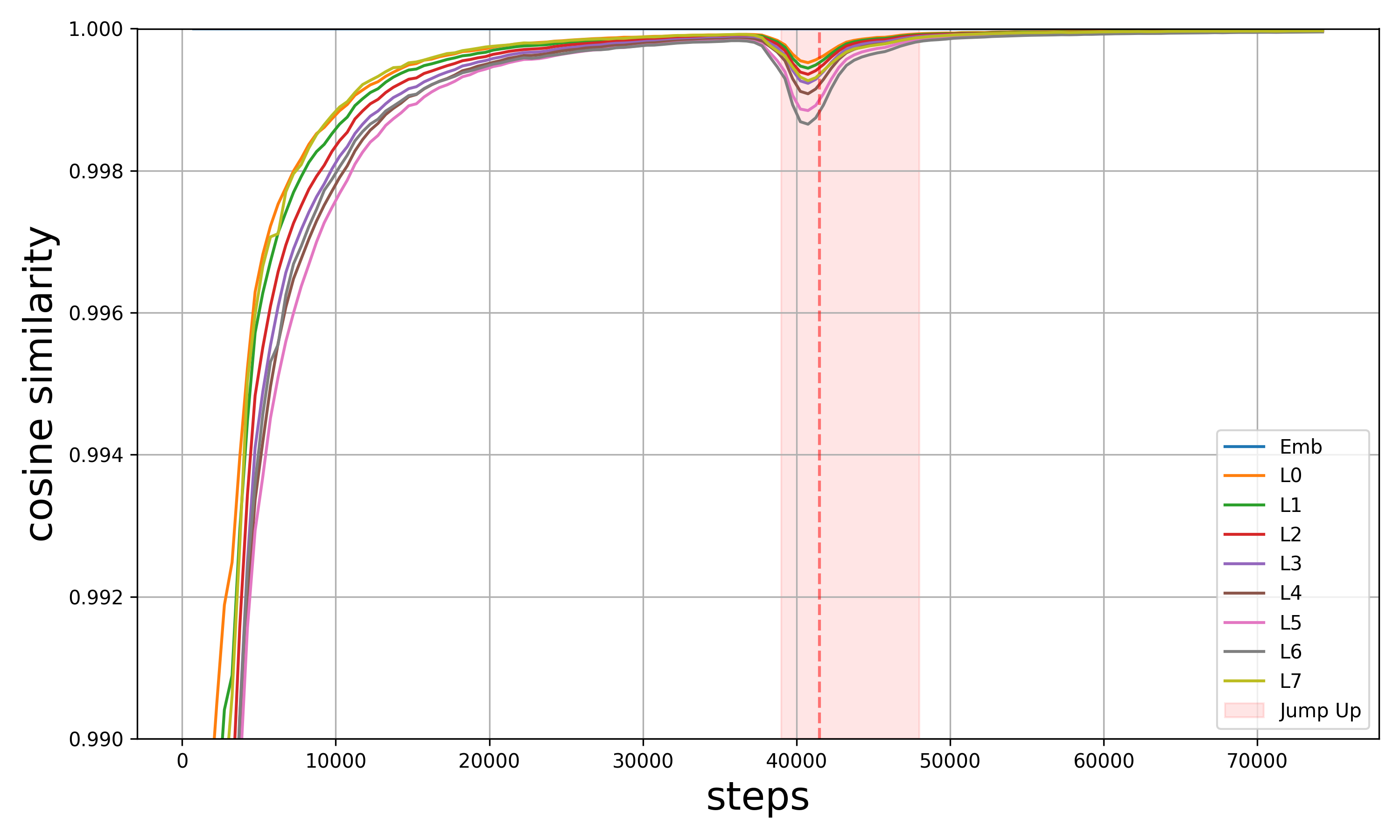}
        \caption{Chiral - cosine similarity}
        \label{chiralcosdecoder}
    \end{subfigure}
    
    \caption{Trajectories of residual stream–related metrics from L0 (layer 0) to L7 for the \textit{pancore-addonce} decoder on ZINC20 (\til 100). (\subref{backgroundl2normdecoder}, \subref{chirall2normdecoder}) L2 Norm; (\subref{backgroundcosdecoder}, \subref{chiralcosdecoder}) cosine similarity between consecutive checkpoints. (\subref{backgroundl2normdecoder}, \subref{backgroundcosdecoder}) Background tokens excluding chiral tokens. (\subref{chirall2normdecoder}, \subref{chiralcosdecoder}) Chiral tokens. The red vertical dashed line and the light red shaded region indicate the perplexity-derived jump-up interval.}
    \label{decoderresidualmetrics}
\end{figure}

\newpage
\begin{figure}[H]
    \centering
    \begin{subfigure}{0.48\textwidth}
        \centering
        \includegraphics[width=\linewidth]{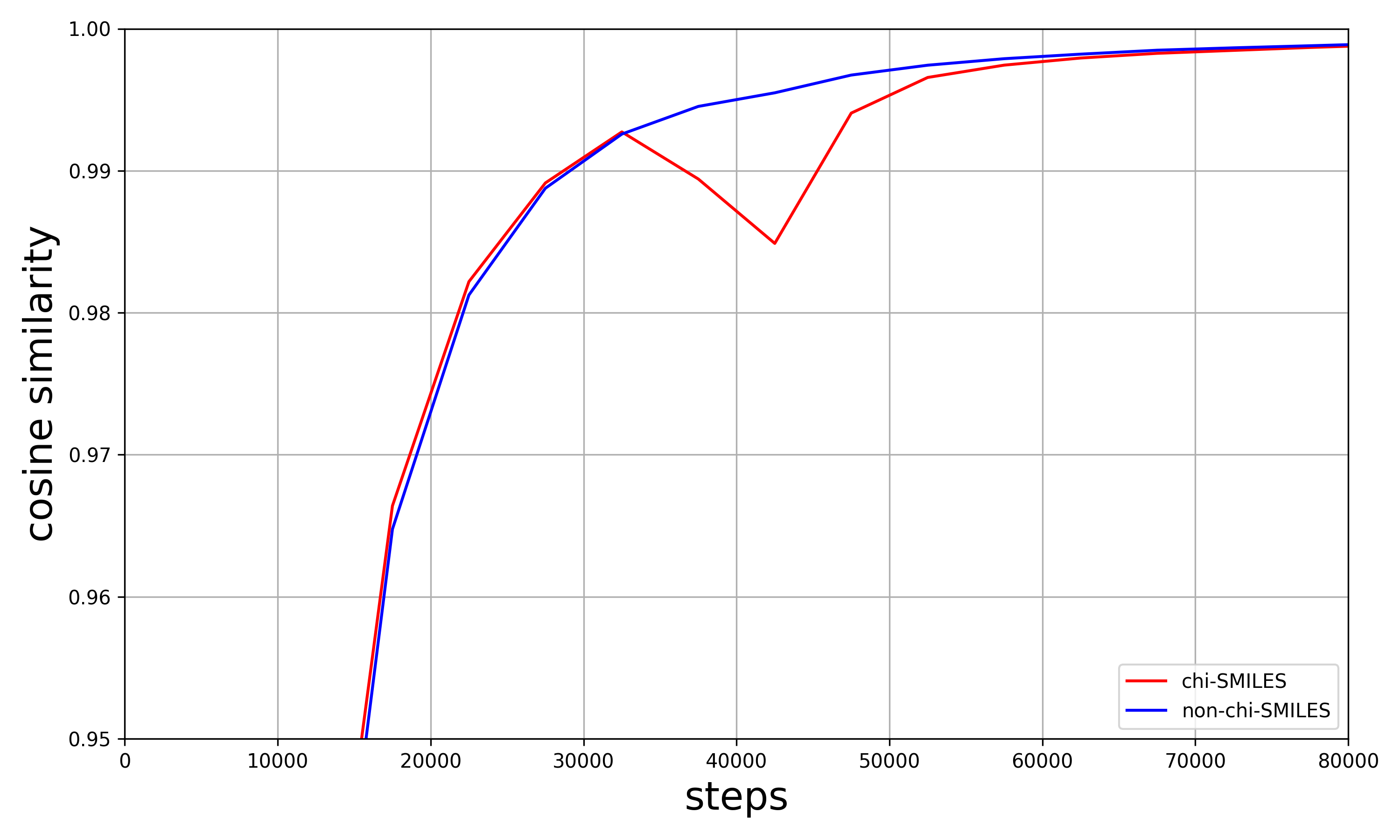}
        \caption{cosine similarity}
        \label{latentcosdelta5000}
    \end{subfigure}
    \hfill
    \begin{subfigure}{0.48\textwidth}
        \centering
        \includegraphics[width=\linewidth]{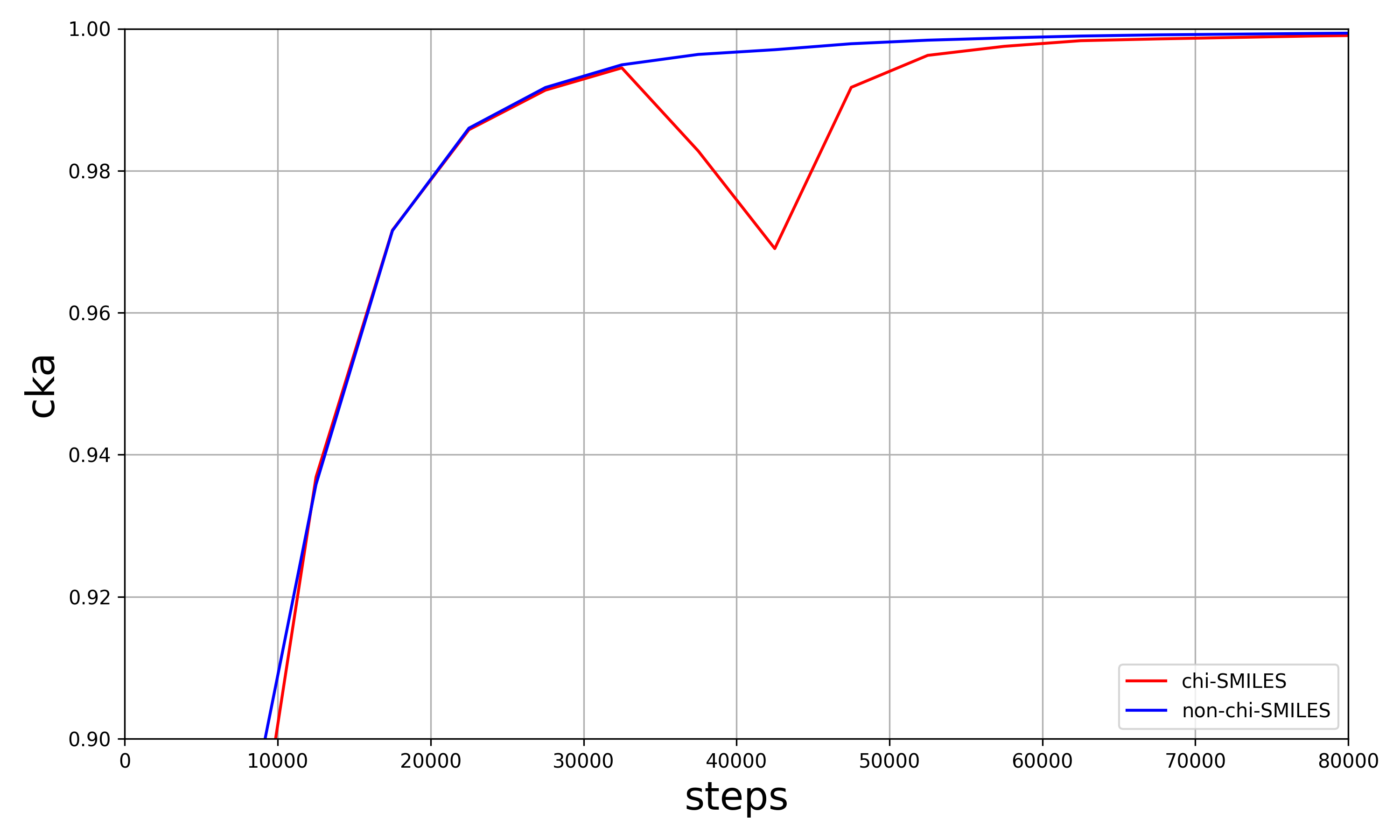}
        \caption{CKA}
        \label{latentckadelta5000}
    \end{subfigure}
    \caption{Latent vector metrics evaluated at 5,000-step resolution ($\Delta5000$steps) for \textit{pancore-addonce}, ZINC20 (\til 100)): (a) cosine similarity and (b) Linear CKA between consecutive checkpoints. Red lines: chi-SMILES; blue lines: non-chi-SMILES. In contrast to the 500-step resolution in Fig.~\ref{latentmetrics}~\subref{latentcos},~\subref{latentcka}, spike-like fluctuations exclusive to chiral SMILES are more pronounced in absolute terms.}
    \label{latentdelta5000}
\end{figure}

\end{document}